\pdfoutput=1

\documentclass[11pt]{article}

\usepackage{times}
\usepackage{latexsym}
\usepackage[T1]{fontenc}

\usepackage{acl}

\usepackage{times}
\usepackage{latexsym}

\usepackage{newtxtext,newtxmath}
\usepackage{tabularx}
\usepackage{adjustbox}
\newcolumntype{A}{>{\raggedright\arraybackslash}p{1.8cm}}
\newcolumntype{B}{>{\raggedright\arraybackslash}p{2cm}}
\newcolumntype{C}{>{\raggedright\arraybackslash}p{4cm}}
\newcolumntype{D}{>{\raggedright\arraybackslash}p{6cm}}
\newcolumntype{E}{>{\raggedright\arraybackslash}p{6cm}}
\newcolumntype{N}{c}

\newcommand{\MRText}[3][2cm]{%
  \multirow{#2}{*}{\parbox{#1}{\centering #3}}%
}

\usepackage[utf8]{inputenc}

\usepackage {caption}
\usepackage{enumitem}

\usepackage{inconsolata}

\usepackage{microtype}
\usepackage{ulem}
\usepackage{xltabular}
\usepackage{multirow}
\usepackage{booktabs}
\usepackage{rotating} 
\usepackage{placeins}
\usepackage{stfloats}
\usepackage{longtable}
\usepackage{pdflscape}
\usepackage[table,xcdraw]{xcolor}

\usepackage{graphicx}
\usepackage{minted}
\setminted{formatcom=\rmfamily}

\usepackage{adjustbox}

\title{
Prompts in the Wild:\\ A Large Analyzed Collection of Transactional Prompts in Code}

\author{Victoria Basmov\textsuperscript{\normalfont1,2} \, Yoav Goldberg\textsuperscript{\normalfont1,2} \,Reut Tsarfaty\textsuperscript{\normalfont 1}\\
\textsuperscript{1}Bar-Ilan University \, \textsuperscript{2}Allen Institute for Artificial Intelligence \\ 
{\tt\{\href{mailto:vikasaeta@gmail.com}{vikasaeta}, \href{mailto:yoav.goldberg@gmail.com}{yoav.goldberg},
\href{mailto:reut.tsarfaty@gmail.com}{reut.tsarfaty}\}
@gmail.com}}

\begin{document}
\maketitle
\begin{abstract}
The behavior of contemporary generative Large Language Models (LLMs) is directly shaped by {\it prompts}, unstructured texts that describe the desired output and  model behavior. In this paper we argue that prompts are linguistic objects that merit investigation in their own right.
To this end, we collect 57.5K unique samples of  prompts from GitHub. Specifically, we focus on transactional prompts: reproducible natural language instructions that are integrated into software.
To enable the empirical, quantitative study of prompts, we introduce a structured ontology, capturing the properties  of prompts as well as their formal and semantic components. Based on this ontology, we transform prompts from unstructured raw texts into richly structured linguistic objects. Analysis of these structured data reveals significant diversity of usage patterns across languages, domains, tasks, and modalities, in a typical Zipf-like distribution where some clearly prevail and others, more diverse, appear in the long tail. To validate the reliability of the ontology-based annotation of the prompts, we perform a comprehensive error analysis across all fields, providing a detailed assessment of annotation quality.
We release the dataset together with a browsing and exploration interface\footnote{\url{https://github.com/OnlpLab/transactionalPromptsCollection}}.
\end{abstract}

\section{Introduction}
\label{sec:introduction}
{Prompts}, the instructions humans give to large language models (LLMs), constitute the primary interface for guiding models' behavior. Despite growing practical interest in prompt engineering and prompting strategies, prompts are still treated largely as informal and intuitive artifacts rather than objects of systematic scientific inquiry.
While models are analyzed in great depth, the usage of prompts, natural language utterances which directly shape models' behavior, remains largely ad-hoc  \cite{villamizar2025promptssoftwareengineeringartifacts}.
A systematic study of prompts may reveal crucial aspects of LLMs usage patterns: 
what languages are used in prompts and how? what structural and semantic patterns do they follow?
what tasks are they used to solve? what common practices emerge in prompt design? and a lot more.
However, tapping into these questions and investigating them empirically, requires injecting  into prompts structure that would allow for {\it quantitative, rigorous} analyses.
Moreover, formalizing prompt structure is essential for the linguistic analysis of prompts  \citep{Jeoung2025PromptPrismAL, Leidinger2023TheLO};  research of sensitivity to linguistic and structural prompt variation \citep{Cuellar2026TrustingCW,  Arabzadeh2025VAP3VP,Wahle2024ParaphraseTE}; for tools and methods of structure-aware and linguistically informed automated prompt optimization  \citep{Santos2025DiversePI,  Hidalgo2025PromptsEF, khattab2023dspycompilingdeclarativelanguage, Saletta2024ExploringTP, khattab2022demonstrate, murthy2025promptomatixautomaticpromptoptimization,juneja2025taskfacetlearningstructured}; multilingual prompt engineering \citep{Vatsal2025MultilingualPE,Zhang2025CrossLingualPS, Kmainasi2024NativeVN}; and  other areas of prompt research and downstream tasks.

We aim to establish prompts as first-order objects of scientific study. A phenomenon becomes a scientific object of study once practical relevance and sustained research are complemented by a shared {\it formal framework}, which prompts still lack. 
While prompts are already gaining interest, not only as tools for using LLMs but also as independent objects of study  \citep{Pister_2024, mao2025promptstemplatessystematicprompt, vir2025promptevalsdatasetassertionsguardrails, zheng2024lmsyschat1mlargescalerealworldllm,villamizar2025promptssoftwareengineeringartifacts}, prompt research  lacks common terminology, structure, and large-scale empirical grounds. 
This  work aims to fill this gap. 

To facilitate the study of prompts, we collected a dataset of 57.5K unique prompts from public GitHub repositories. We specifically focus on \textit{transactional prompts},\footnote{The term "transactional prompts" is defined by M. Hashimoto (\url{https://mitchellh.com/writing/prompt-engineering-transactional-prompting}) to distinguish them from interactive prompts. They are also sometimes referred to as "developer prompts", but the latter term is often used in another sense: interactive prompts by software developers.} prompts that are intended to perform reproducible, parameterized tasks, as part of a larger software-based workflow. Unlike casual (``interactive") prompts, which are ad-hoc and one-off interactions with LLMs as part of a user-LLM conversation, transactional prompts run within pre-defined automated workflows and are refined to be robust and repeatable. Studying transactional prompts offers vital insights into real-world LLM usage in software applications.\footnote{Transactional prompts are predominantly single-turn. In contrast, multi-turn interactive user–LLM dialogues fall outside the scope of this work and are already addressed by existing large-scale datasets (Section \ref{sec:related_work}). In addition, the emerging paradigm of agentic usage introduces prompts designed for iterative tool-use loops, which differ structurally and functionally from transactional prompts. Such prompts require dedicated investigation and potentially a specialized dataset, but they are beyond the scope of the present study. While our dataset may incidentally contain prompts that originally formed part of a chain or loop, they are represented as independent entries in our corpus.}

Unlike conventional programming, LLM instructions realized in prompts are specified in unstructured texts that convey complex, hierarchical, and multi-faceted messages, using natural language to encode a mixture of instructions and information.
To explore the expressive power of this new “natural programming language” and the structural, compositional, and linguistic mechanisms through which this semantic range is expressed, it is helpful to introduce structure into otherwise unstructured prompt texts. To this end, we define  an ontology underlying prompt structure.

The ontology provides a systematic framework for investigating the complex semantics encoded in prompts and the diversity of expressive means employed to express it (Section {\ref{sec:ontology}}). It helps to uncover underlying semantic and structural patterns within widely diverse prompt data, providing a foundation for the empirical investigation of prompts as a “programming language” for LLMs across diverse  applications, languages, and modalities.

Our analysis of the resulting structured prompts (Section \ref{sec:analysis}) reveals a rich diversity of prompt usage across languages, modalities, tasks and domains, exhibiting a Zipf-like distribution --- typical of linguistic phenomena \citep{Piantadosi2014, Linders} --- with a prominent head and a much more diverse and nuanced long tail. 

Importantly, our data collection and ontology are not intended to be final or exhaustive. Rather, they represent a starting point paving the way for further investigation by linguists, prompt researchers, and engineers, who can contribute complementary perspectives to the research.

The prompt collection we deliver is accompanied by an online user interface for browsing, searching, and exploring prompts by their various characteristics and components, as defined by the underlying ontology. Alongside the collection and the ontology, it is intended as a practical resource to inspire further investigation into this topic. 

In sum, this work treats prompts as first-class objects for empirical scientific and linguistic investigation, and makes four main contributions: 
   \\ (i) \textbf{a large-scale dataset} of 57.5K transactional prompts gathered from GitHub;      
    \\(ii) \textbf{a structured prompt-ontology} that captures the primary prompt features and components;    
    \\(iii) \textbf{empirical analysis} of the structured prompts, highlighting patterns in the way programmers use LLMs; 
    and
    \\(iv) \textbf{a user interface} for browsing and searching prompts by their properties and components to support further research.

We aim for these resources to facilitate linguistic and practical investigations into how humans interact with LLMs and how prompts’ structure interacts with the semantic space they construct when using the prompting language as part of software engineering —  pragmatic, syntactic, and lexical variations they use, how they structure prompts to elicit outputs with specific forms and content, the strategies they employ to overcome the inherent ambiguity and underspecification of natural language, how they choose the language of the prompt (English vs. other languages), are all prompts similar stylistically due to style convergence induced by LLMs, in what ways  prompting language diverges from ordinary human language,  and other potential perspectives.

\section{Collecting Transactional Prompts}
\label{sec:collection}
Following standard practices \citep{liu2025repodebugrepositorylevelmultitaskmultilanguage, Li2022AutomatingCR, mt4py2021, alon2018codeseq, 150kPythonDataset}, we collect prompts from Github repositories \footnote{We use GitHub due to its availability and convenience. It is the largest repository of real world software projects, and its APIs allow flexible search and retrieval over the entire collection. We believe it is the current best source for locating prompts that are used as part of software projects. However, while large and diverse, it is also a biased source: for example, it does not include enterprise and closed-source projects, which may have a different distribution of prompts.}, by looking for files that either invoke the \texttt{chat.completion.create} API or the \texttt{PromptTemplate} constructor from the LangChain package\footnote{We chose these two APIs due to their popularity and standardization: they are both widely adopted, and used in a consistent manner that make prompt extraction feasible. This comes at the expense of biasing the prompt collection to projects that use these APIs. This excludes, for instance, projects that write their own LLM access wrappers, or use other libraries.}, and attempt to extract the contents of the \textit{messages} (for completion.create) or \textit{template} (for PromptTemplate) parameters from each call-site.

The immediate content is often a formatted string or a variable name, which we then aim to resolve to the actual prompt content via static analysis of the code, recursively tracking string values across variable assignments and function calls. The cases where this resolution succeeds are then filtered using a set of heuristics to retain only semantically-contentful prompts, and the filtered results are deduplicated. 
For each resulting prompt we further retain its metadata such as the repository name, file-path and URL,
as well as the last commit date for the prompt. This process resulted in 57,640 unique prompts. Of these, 36,916 came from \texttt{chat.completion.create} and 20,724 from \texttt{PromptTemplate}. Details of prompt text extraction are available in the Appendix \ref{sec:prompt-retrieval}. Details of filtering and deduplication are described in Appendix \ref{sec:deduplication}.

\section{The Prompts Ontology}
\label{sec:ontology}
\begin{figure*}[t]
\includegraphics[width=\textwidth]{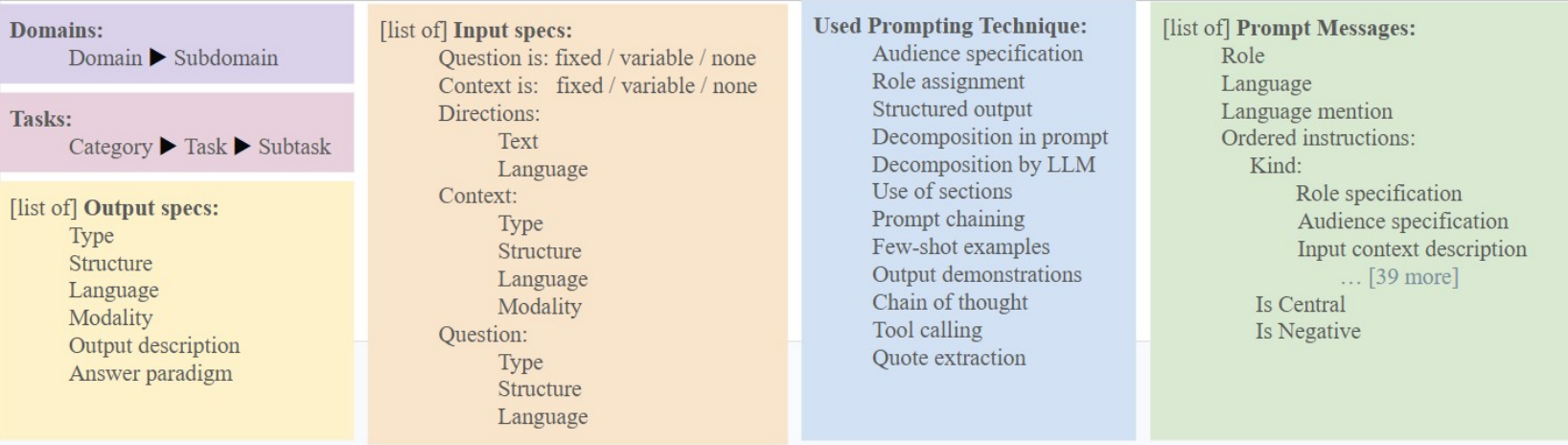}
\caption{The {\bf Prompt Ontology} Underlying the Empirical Analysis and the Structured Collection}
\label{fig:ontology}
\end{figure*}

In order to analyze prompts more deeply, we introduce an ontology outlining their main components and dimensions. Figure \ref{fig:ontology} shows a bird's-eye view of our proposed ontology, with the specific fields explained shortly. 

The ontology categories are grounded in three complementary sources: (1) Inherent prompt properties (e.g., prompts are texts written in a \textit{language}; prompts by definition formulate a \textit{task}. prompts serve to elicit certain \textit{output}), (2) prior literature, e.g., prompting techniques \citep{santana2025promptingtechniqueiuse,schulhoff2025promptreportsystematicsurvey}, context-grounded vs. parametric prompting \citep{Zhou2024EstablishingKP, sun2026taskmattersknowledgerequirements}, and (3) recurring lexical and structural components we identified through manual inspection of sampled prompts, such as explicit language mentions, semantically distinct instruction blocks etc. The categories capture orthogonal dimensions of the data and are not intended to constitute a mutually exclusive or collectively exhaustive taxonomy. Rather, the ontology is deliberately non-restrictive and extensible; dataset users may adopt, adapt, extend, or replace it as appropriate for their purposes. The utility of the proposed ontology is directly demonstrated by its application in the data annotation and the analysis we present (Section \ref{sec:analysis}).

\paragraph{Languages:} detected languages used in prompt texts and any explicit language mentions.

\paragraph{Task and domain:} we track the tasks for which the prompt is intended, and its application domain. These have both coarse-grained and fine-grained categories.

\paragraph{Input characteristics:} At the outset, prompts specify one or more of (1) overall high-level instructions (``answer the question provided by the user)"; (2) a question/task to be solved (``how many apples did John eat?"); (3) supporting context for 2. Each of these can be either hard-coded in the prompt or be a variable provided as input in each invocation. We identify the cases where each of these information types are read as input. For each identified input slot, we also retain information about its language, structure and modality, if available.

\paragraph{Output characteristics:} Each prompt's expected output is annotated for modality and for the requested structure, language and answer paradigm.

\paragraph{Prompt structure:} Each prompt is represented as a list of role-messages (``System", ``User", ``Assistant"), and their associated texts.\footnote{The LangChain PromptTemplate prompts are strings and not message sequences. These are represented as a single message with role ``Undefined".} For each message we list, beyond its text and role, also its detected language and languages explicitly mentioned within it. We also further break the message text into a sequence of individual instructions. Each instruction is associated with one of 42 semantic kinds (e.g. ``role specification", ``audience specification", ``input content description". See Appendix \ref{sec:instructions} for the complete list). We explicitly mark \textit{negative} instructions, and distinguish between \textit{central} and \textit{auxiliary} instructions. 

\paragraph{Prompting techniques:} For each prompt we extract a list of prompting techniques with records specifying which techniques are used in the prompt, with supporting evidence spans. The prompting techniques come from a pre-specified list of 12 techniques (e.g. ``use of sections", ``structured outputs", see Figure \ref{fig:ontology}).

 \paragraph{Meta-data:} Additionally, each object in the data includes an ID,  a GitHub URL of the source file,  timestamp of the last update, full prompt text (a concatenation of all individual message texts in the prompt) and its translation into English if not in English already.

 For some fields (input and output structure, modality and variability, prompting techniques, instruction kinds, role), we defined the possible value inventories, although in some cases (e.g., structure fields) we instructed the model to enrich predefined values with additional detail (e.g., “Dictionary of items (‘From’: string, ‘To’: string)” rather than just “Dictionary”). For the remaining fields, values were generated by the LLM during annotation. For fields with especially large inventories (e.g., class, domain, input and output type), the values were then grouped into classes using the algorithm described in Appendix \ref{sec:clustering}.

\section {Annotation, Quality Control and Error Analysis}
\label{sec:error-analysis-main}
The ontology annotation is performed using an LLM-based process. The prompts used for annotating the prompt collection, and the detailed data-model, are listed in Appendix \ref{sec:data-model}.

We iteratively tested, manually evaluated and refined annotation prompts on sample data. Concretely, we jointly performed human-in-the-loop checks: we sampled $\approx 100$ items per major metadata category and manually inspected the automatic labels and their evidence spans. Any disagreements between the authors were resolved through peer discussions. These checks were used to refine the annotation prompts until labels became consistent, and the manual inspections showed consistently high agreement between the automatic labels and human judgments across categories.

To assess the resulting annotation quality, we manually evaluate additional 100 randomly selected data points (50 from each source) and perform error analysis across all fields.\footnote{
For detailed error analysis results see Tables~\ref{tab:errors-language}--\ref{tab:errors-accuracy} and Figure \ref{fig:errors-accuracy} in Appendix \ref{sec:error_analysis_tables}.
} In the error analysis, predicted labels and evidence spans were evaluated against the annotation guidelines provided to the model in the respective prompts (see Appendix \ref{sec:data-model}). Detected errors were then grouped into recurring categories and quantified to characterize the main failure modes. The  error analysis was conducted by a single expert, due to the substantial time required for this kind of fine-grained manual review.

The error analysis shows generally strong performance across most fields, with many categories exceeding 90\% accuracy (e.g., Prompt Language 93.0\%, Domain ~90.6\%, Context Language/Modality ~97\%, Central vs. Meta 96.4\%, Prompting Techniques 98.5\%). However, several fields remain more challenging, especially Output Type (60.4\%) and Directions Text (69.4\%), with moderate error rates in Answer Paradigm (80.2\%), Instruction Kinds (89.8\%), and Context Structure (81.9\%). The observed errors fall into several recurring groups: (1) \textit{hallucinations}, where the model assigns attributes not grounded in the prompt (e.g., inventing output types, domains, or languages); (2) \textit{confusions between related categories}, such as conflating prompt language with explicit language mentions, labeling restrictions as negative instructions, or reporting output structure instead of context or question structure; (3) \textit{omissions}, where relevant elements are not detected (e.g., missing tasks, ignored placeholders, undetected context or question units); and (4) \textit{segmentation and granularity errors}, including failure to split complex instruction blocks or grouping multiple output types into a single “complex” type. Notably, many errors arise when indirect inference is required—e.g., coreference resolution, common-sense reasoning, or domain knowledge (inferring the output type, not mentioned directly, from few-shot examples, recognizing a Python function signature as code etc.). The results suggest that performance degrades primarily in cases involving implicit information and fine-grained annotation distinctions. 
 
\section{Analysis}
\label{sec:analysis}

\subsection{Language, Modality and Domain}
\paragraph{Language. }
Which languages are prompts written in? Naturally, English is predominant, but to what extent? And how diverse are the other languages? Our analysis shows that the dataset encompasses prompt messages in 62 languages. English is overwhelmingly dominant, accounting for 84.66\% of identifiable cases.\footnote{In 9.35\% of the cases, the language (annotated per-span) could not be identified (e.g. when the span consisted of only a placeholder, like ``\{system\_txt\}'').} The other languages that constitute above 1\% are \textit{Chinese, Korean, Spanish, Japanese and Portuguese}. The following seven highly represented languages are mostly (but not only) European: \textit{French, Russian, German, Indonesian, Vietnamese, Polish, Italian and Dutch} (see Figure \ref{fig:common-languages} Appendix \ref{sec:charts}).\footnote{Interestingly, the highly represented languages include the top 10 prompt languages reported by \citet{Pister_2024}.} The long-tail languages occurring below 100 times in the data, with \textit{Hindi} at the beginning and \textit{Tagalog} at the very end, are shown in Figure \ref{fig:long-tail-languages} (Appendix \ref{sec:charts}).

\paragraph{Multilinguality.} 
Next, we examine the presence of multilingualism. Only 6.3\% of prompts (3,649) are multilingual, and of these, over 99\% include English. Despite this English dominance in mixed-language queries, 8.19\% of the total dataset (4721 prompts) consists of entirely non-English text.

\paragraph{Language Mentions.} Beyond the language used to write the prompt, we also annotate the prompts for explicit language mentions—instances where a language is referenced but not necessarily used (e.g., ``Translate this to Afrikaans"). This often-overlooked dimension of multilinguality reveals even greater linguistic diversity. In terms of
total mentions, we observe 15,331 references to natural human languages. In terms of language diversity, while English leads (9,510 mentions), the remaining references cover 151 languages and dialects, a far broader range than is found in the languages directly used in prompt texts. The listing and distribution of these non-English mentioned languages  is detailed in Figure \ref{fig:mentions} (Appendix \ref{sec:charts}).

\paragraph{Modality. }
While it is obvious that in LLMs text is the dominant input and output modality, what other combinations of input and output modalities do we see in the data? And can we find even greater diversity in the long tail? Where is the predominance of text more pronounced: in input or in output?

In the overwhelming majority of cases, the input context modality (77.82\% of the cases) and the output modality (over 97\% of the cases) is text.\footnote{The remaining cases are those where no context was identified (10.75\% of the prompts) or the input or output modality was unidentifiable (for example, the input modality can remain undefined when the input is a PDF document, but the prompt does not specify if it contains text, an image, or both). This covers 8.23\% of the inputs and 2.77\% of the outputs.} The distributions of non-text or mixed modality for input and output are shown on the Figures \ref{fig:input_non_text-modalities} and \ref{fig:output_non_text_modalities} (Appendix \ref{sec:charts}) with images accounting for a much greater share than audio and video.

The frequent input-output modality combinations all have textual output and differ only in input: the prevailing combination is text$\to$text (77.31\% of all the data); with the other categories--- ungrounded or undefined (18.98\%), image$\to$text (1.98\%), image+text$\to$text (0.66\%), audio$\to$text (0.23\%) --- lag far behind (Fig \ref{fig:main_modality_combinations}, Appendix \ref{sec:charts}).

\paragraph{Domain.}
In which domains are transactional prompts most frequently used?  Which domains are leading and which are in the long tail?
Prompts with a specific domain that can be identified\footnote{Cases where a domain can not be identified are often short general prompts such as, ``Please answer the user question using only the given context.''} constitute 57.38\% of the prompts.
This resulted in 77 distinct domains with a long-tail, Zipf-like distribution. 
The top leading domains are, in order: \textit{education} \& \textit{instruction}, \textit{software development}, \textit{business} \& \textit{commerce}, \textit{healthcare} \& \textit{medical}, \textit{technology, media } \& \textit{entertainment}, \textit{finance} \& \textit{banking}, \textit{creative writing} \& \textit{content-creation}, \textit{human resources}, \textit{arts} \& \textit{culture}. Together, they cover almost half (49.58\%) of all the domains in the data, and appear in 63.63\% of the prompts with specific domains.\footnote{A single prompt can belong to multiple domains.} 
Mid-frequency domains include, for example, \textit{hospitality \& food service}, \textit{design \& arts}, \textit{personal services} and \textit{philosophy}, while low-frequency domains include \textit{urban development}, \textit{historical studies}, \textit{politics} and others.
See Appendix \ref{sec:domain-counts} for the full list of domains.

\subsection{Structure and Semantics}
\paragraph{Instruction Kinds.} A prompt (or a message therein) can be interpreted as a sequence of instructions given to the LLM. But what is the semantic or functional structure of prompts? That is, what are the semantic types of instructions and their order in transactional prompts? Our ontological structure treats each message as a sequence of instruction items, each labeled with its semantic function.

Overall, the dataset includes 39,4875 such instruction blocks, 6.85 per prompt on average. The top 10 most frequent types are:
\begin{itemize}[leftmargin=*, nosep, topsep=0pt]
    \item Input context placeholder \hfill (16\%)
    \item Constraint or restriction \hfill (11\%)
    \item Output content requirements \hfill (9.7\%)
    \item Output format requirement \hfill (9.7\%)
    \item Role specification \hfill (8.1\%)
    \item Input context description \hfill (7.1\%)
    \item Central task/question description \hfill (6.7\%)
    \item Central task/question \hfill (5.8\%)
    \item Input contextual data \hfill (4.3\%)
    \item Conditional instruction \hfill (2.9\%)
\end{itemize}
Together, they cover 81.33\% of all blocks in the data. 
Full statistics, as well as information about frequent ordering of units, are available in
Appendix \ref{sec:charts}, Figure \ref{fig:block-frequencies}, and Tables \ref{fig:block-sequences} \& \ref{fig:block-sets}.

\paragraph{Core vs.\ Supporting Instructions.}

 Generally, 18.2\% of instructions represent the Central Task (the ``core intent''), while the remaining 81.8\% function as meta (supporting) instructions providing guidance on style, constraints, or formatting.
 Most of the core instructions focus on the task/question detailed descriptions (36\%) or define the tasks/questions themselves (31.9\%).

 In the example below, the core instruction defines \textit{what} the task is while the supporting ones add details by specifying \textit{how} it should be performed:
 
 \noindent "Please create a learning plan in \{language\}." (\textit{core}) "The plan should outline daily activities." (\textit{supporting}). "Make sure to include detailed information about the specific programming languages and tools (like APIs) that will be used." (\textit{supporting}) "Do not include learning of languages that I have already used." (\textit{supporting}).

Only 4\% of prompts consist exclusively of a central task with no metadata. These are typically short, non-grounded queries (e.g., ``Explain how to write a window in Python'', ``Name ten mammals") or where the context is not provided in the prompt text (e.g., ``Identify products using the given images and generate key features for each product.'').

\paragraph{Negative Instructions.} 
\label{paragraph: neg_instructions}
Negative instructions (telling the model what not to do) serve as a window into user mitigation of  LLMs  tendencies, such as verbosity, hallucination and different kinds of bias.

Roughly 31\% of all prompts contain at least one negative instruction.  Overall, negative instructions  represent only  7\% of the total instruction count in the dataset.
 The vast majority (89.5\%) of the negative instructions are categorized as constraints or restrictions. These act as guardrails against undesirable behaviors, such as adding extra text beyond the requested output, or exceeding specific lengths.
A smaller portion addresses output format (5.4\%), content requirements (1.4\%), and error handling (1.1\%).
Examples of negative instructions include (more in Appendix \ref{sec: neg-examples}):
\begin{itemize}[leftmargin=*, nosep, topsep=0pt]
    \item \textit{don't make the answers too long} (constraint/restriction)
    \item \textit{If you encounter an exception, an effectless command, or find yourself in a loop, avoid repeating the same command and try something else to achieve the goal.} (error handling instruction)
    \item \textit{Don't translate the text to English. Keep it in Indonesian.} (linguistic constraint/specification)
\end{itemize}
\indent Negative instructions very rarely form the core intent of the prompt (0.5\% of cases). In these instances, the primary task is defined by what the model must avoid, e.g., 
``Do not answer the questions, simply provide a correct compute graph..." or
``Do not respond to text, merely translate it." 

\paragraph {Constraints.} 
We define constraints as instructions involving restrictions, style/format requirements, design specifications, or negative directives.
As research shifts toward evaluating how well LLMs adhere to nuanced requirements \citep{zhou2023instructionfollowingevaluationlargelanguage, lior2025wildifevalinstructionfollowingwild}, our dataset emerges as a source of naturally occurring constraints. Like negative instructions, the isolation of prompt constraints also offers opportunities for linguistic investigations on the form and structure in which they are expressed.

Our analysis reveals a landscape of high complexity. Constraints represent 33.3\% of all instruction blocks in the dataset.
 On average, a single prompt contains 2.28 constraints, but the ``long tail" is significant  --- some transactional prompts contain up to 155 distinct rules.
 Furthemore, while 27.4\% of the prompts are constraint-free, nearly 14\%  layer five or more constraints in a single query, signaling a demand for high-precision model control.

Unlike synthetic benchmarks, our data captures the ``messy" and layered constraints actually deployed in real-world scenarios, making it a unique resource for investigating the limits of LLM steerability in real-world transactional environments.

\paragraph{Messages.}
The popular \texttt{chat.completions .create} API expects prompts as a sequence of role messages (``System", ``User", ``Assistant"). How do prompt writers use this interface? How many messages do they use, and what roles are assigned?
The majority of prompts (64\%) have two messages. Of these 96\% are ``system-user". 27\% of the prompts have a single message, with ``user" (70\%) being far more frequent than ``system" (28\%). For three-message prompts (4\%) the most popular sequences are ``system-user-user'' (26.41\%), ``system-user-assistant'' (22.01\%), ``system-system-user'' (18.45\%), ``system-assistant-user'' (13.07\%). For prompts with more than three messages, the majority (52\%)  includes alternations of the ``user'' and ``assistant'' messages, alternatively with a system prompt at the beginning. (See details in Figure \ref{fig:messages-per-prompt} Appendix \ref{sec:charts} and Figure \ref{fig:role-sequences} Appendix \ref{sec:charts}).
It thus appears that the ``system-user'' duo has become the standard unit in transactional prompts. This suggests that developers largely view the System role as a static configuration layer and the User role as the dynamic input, 
rather than utilizing the API for complex, multi-turn role-play within a single template.

\subsection{What Are Prompts Being Used For?}
\paragraph{Tasks.} What tasks are typically performed using LLMs? Are LLMs used more for standard NLP tasks or for other, non-NLP tasks? From these tasks (NLP vs.\ non-NLP), to what extent either is more pronounced? Are the tasks used across domains or are some of them limited to a specific domain?
The distribution of NLP tasks in the data covers both NLP tasks (involving language understanding, generation and analysis) and tasks outside classical NLP (like data processing, analysis, multimodal, and structured‑data tasks). 

It is clear from the data that NLP tasks prevail: the top 4  (question answering, general text generation, information extraction, and summarization)  cover over 48\% of the data (see Figure \ref{tab:top-4-tasks}, Appendix \ref{sec:charts}). Non-NLP tasks occur mostly at moderate to low frequencies. In the long tail (tasks with under 30 cases) we see non-NLP tasks, such as education design, game strategy, state tracking, data privacy, policy generation, and system integration. The top 10 in long-tail and mid-frequency tasks are shown in Table \ref{tab:task-frequency} (Appendix \ref{sec:charts}). We further see that each task appears in multiple domains, from 4 (game strategy) to 77 (question answering), and no task is purely domain-specific.

\paragraph{Inputs.}
It is common for prompts to have a \textit{question} or \textit{main task} that needs to be answered, either based on a \textit{context} that is also provided, or based on parametric knowledge. Either of these components can be hard-coded into the prompt, or be a varying input to the prompt. In our data, 97.9\% of the prompts included a question or a main task, and 89.2\% included a supporting context. From these, the question/task is expected as input 70\% of the time, and was hard-coded in the prompt for the remaining 30\%. In contrast, a context is provided as input 94\% of the time and is only hard coded 6\% of the time. This suggests an (expected) tendency to perform the same task over varying contexts rather than performing varying tasks over a fixed context.

\paragraph{Grounding.}
LLMs may be expected to respond  either based on their internal parametric knowledge, or based on grounding context provided to them. Which of these options is more prevalent in real-world transactional usage? And, in the  case of grounded prompts, what kinds of contexts are used for grounding?

Our analysis  suggests that 89.25\% of all cases are grounded, i.e., performed based on a certain context rather than merely based on the LLM's parametric knowledge.\footnote{The cases where no context was found are not necessarily non-grounded. This can be due to other reasons. Sometimes the context is added to the prompt dynamically, beyond the \texttt{chat.completions.create} API call or the \texttt{PromptTemplate} static instantiation, and in these cases our system was unable to capture it.} In the vast majority of cases the type of the grounding context is text (that may include a variety of subtypes such as document, paragraph, sentence, abstract, tweet, proverb, caption, description, instruction, etc.). Other frequent types of context include a question, dialogue/conversation, code, table, image, numeric context, json. The distribution of top 10 input types across the top 10 tasks is shown in Figure \ref{fig:task-input} (Appendix \ref{sec:charts}). 

\subsection{Prompting Techniques}
Which prompting techniques are adopted by prompt writers, and to what extent? \\
\noindent\textit{Role assignment} is by far the most frequently used technique, accounting for over 45\% of all instances (Figure \ref{fig:prompting-technique-counts}, Appendix \ref{sec:charts}).  Explicitly defining the role of the AI was one of the first well-known techniques, and was recommended by model developers. The data shows that this remains a widely adopted practice in prompt engineering, despite reports of its limited effectiveness in more recent models \citep{zheng2024ahelpfulassistantreally,kim-etal-2025-persona}.

The next most commonly-used technique is \textit{structured output} (12.9\%), that is, requesting output in machine parseable formats (json, XML etc.). This technique is closely related to the \textit{output demonstrations} technique (3.34\%)  including examples of how the output should be organized. Together, these constitute 16.24\% of the cases. Also related is the \textit{sections} technique (7.83\%) dividing the prompt into clearly defined, marked sections. These demonstrate the significance of clearly-structured input and output specifications.

The next in frequency is \textit{decomposition via prompt} (10.34\%) meaning that the prompt specifies how to break the task down into sub-tasks or manageable steps. This technique is complemented by the much less frequent \textit{decomposition via LLM} (0.56\%) where the same decomposition is expected to be done by the LLM. Together, these techniques form 10.9\% of the annotated techniques, suggesting the importance of solving complex tasks by breaking them down into more manageable units. 

\vspace{-5pt}
\section{User Interface}
\vspace{-5pt}
\label{sec:ui}
To empower researchers to explore the prompt collection beyond our set of analyses, we provide a web-based UI designed for exploratory discovery and deep-dive analyses into subsets of the data. Users can filter prompts by ontology fields; search the dataset with semantic similarity; see and aggregate counts; inspect individual prompts and their annotated sections; and download prompts and ontological data for their filtered subsets. Additional information on the UI can be found in Appendix \ref{sec:ui-detail}.

\section{Related Work}
\label{sec:related_work}
  As interest in prompts as distinctly designed artifacts has grown, several datasets have emerged, encompassing both \textit{user prompts} and \textit{transactional prompts}.

{\textbf{User prompts}}  are prompts that are intended to be used directly by users,
and  have significantly different characteristics than the transactional prompts we study herein. LMSYS-Chat-1M \citep{zheng2024lmsyschat1mlargescalerealworldllm} is a dataset of 1M user-LLM conversations collected across 154 languages via the Vicuna demo and Chatbot Arena. WildChat \citep{zhao2024wildchat1mchatgptinteraction} is a multilingual, dataset of 1 million timestamped user-LLM conversations (over 2.5 million turns) collected via a ChatGPT/GPT-4 chatbot with explicit user consent, annotated with demographic metadata (state, country, and hashed IPs) to enable behavioral analysis. 
PROMPTEVALS \citep{vir2025promptevalsdatasetassertionsguardrails} is a dataset of 2,087 LLM prompt templates from the LangChain Prompt Hub,\footnote{\url{https://smith.langchain.com/hub}} a repository of user-contributed prompts,  containing a mix of user-prompts and transactional prompts.  PROMPTEVALS is intended for training and evaluating ``assertion guardrails", and has prompts spanning  multiple domains including IT, finance, and healthcare. DevGPT \citep{Xiao2023DevGPTSD} is a dataset of 29,778 ChatGPT prompts and responses from software developers, collected from shared ChatGPT conversations on GitHub and Hacker News for analysis of developers' interactions with ChatGPT and their implications for AI-assisted programming. These datasets are different than ours in that they do not address transactional prompts in software.

Other researchers have studied {\textbf{Non-LLM prompts}}, for instance DiffusionDB \citep{wang2023diffusiondblargescalepromptgallery} and VidProM \citep{wang2024vidprommillionscalerealpromptgallery} that compile large-scale prompts for text-to-image and text-to-video generation, respectively. 
As proposed herein, such collection too would merit from structured representation and empirical analysis, compatible to ours, which is reserved for future research.

Finally, for {\textbf{Transactional Prompts}},  \citet{Pister_2024} introduced PromptSet, a dataset of developer prompts with similar size and collection method to our own. However, they invest less effort in extraction and cleanup compared to us. As a result, as reported by \citet{tafreshipour2025promptingwildempiricalstudy} and verified by us, the data contains many incomplete prompts or prompt fragments that are hard or impossible to analyze. They also do not provide analysis or structuring of the prompts beyond this raw string data. In spite of these limitations, researchers use PromptSet for exploration of different aspects of transactional prompts \citep{villamizar2025promptssoftwareengineeringartifacts, tafreshipour2025promptingwildempiricalstudy, mao2025promptstemplatessystematicprompt},  as well as for development of prompt optimization tools \citep{Rzig2025AnET}. 
Notably, \citet{mao2025promptstemplatessystematicprompt} construct their own small dataset, derived from PromptSet, to analyze real-world prompts that combine static content with dynamic placeholders such as ``input''. After filtering, cleaning  and deduplication, they extract 2,163 such prompts, in which they identified key components and categorized them into one of six semantic categories.

This highlights the interest in transactional prompts research and a clear community demand for larger, higher-quality resources for such research like the corpus presented in this work.

\section{Conclusions}
\vspace{-5pt}
We present a large, real-world collection of transactional prompts; an ontology that captures both the structural components of prompts and their descriptive characteristics; and a web interface for their systematic exploration. These resources enable a range of applications, including linguistic and structural analysis of prompt texts, uncovering common conventions and ``unspoken norms” of prompt composition, comparing recommended versus actual practices, and supporting multiple downstream applications, such as instruction-following and prompt-sensitivity research, more realistic benchmarking, structure-aware and linguistically-informed automated prompt optimization, multilingual prompt engineering and others. We present a preliminary empirical analysis that exemplifies the utility of the provided framework and illustrates some of the kinds of insights this data makes possible, and hope it inspires further interest in the systematic and methodological study of prompts as scientific and linguistic objects in the community.

\section*{Limitations}
Our search for prompts in GitHub files was limited to certain patterns (the \texttt{chat.completion.create} API call and the LangChain \texttt{PromptTemplate} class), whereas many other patterns are possible. Furthermore, we only considered Python files. Thus, we do not present an unbiased sample of transactional prompts, and the observed trends in our analysis may be biased towards a subset of the prompt space defined by users who chose to use these APIs. Additionally, while prompts are continuously created and updated, our dataset represents a snapshot at the time of collection.
Therefore, rather than being exhaustive, our work paves the way to future efforts that could expand the collection, potentially making it dynamic by regularly incorporating new entries over time, and include broader search patterns and additional programming languages. 

The prompt structuring annotations were performed by an LLM, and, though annotation prompts were iteratively tested, manually evaluated and refined on sample data, each annotation result could not be manually verified individually. As a result, the data may still contain some errors or noise despite the efforts to ensure annotation quality.

Finally, the analysis presented in this work only scratches the surface, and many more quantitative and qualitative investigations are possible, including detailed linguistic analysis, diachronic comparisons (changes over time), cross-lingual or domain and task-specific exploration.
We encourage the community to contribute to expanding the data and improving annotation accuracy and to continue and deepen the research in the field of transactional prompts.

\section*{Acknowledgments}
Work on this project was supported by a VATAT grant from the Planning and Budgeting Committee
of the Council for Higher Education in Israel, 
Kamin grant by the Israel Innovation Authority (IIA) and ISF grant number 670/23.

\bibliography{custom}

\begin{thebibliography}{41}
\providecommand{\natexlab}[1]{#1}

\bibitem[{Alon et~al.(2019)Alon, Brody, Levy, and Yahav}]{alon2018codeseq}
Uri Alon, Shaked Brody, Omer Levy, and Eran Yahav. 2019.
\newblock \href {https://openreview.net/forum?id=H1gKYo09tX} {code2seq: Generating sequences from structured representations of code}.
\newblock In \emph{International Conference on Learning Representations}.

\bibitem[{Arabzadeh and Bagheri(2025)}]{Arabzadeh2025VAP3VP}
Negar Arabzadeh and Ebrahim Bagheri. 2025.
\newblock \href {https://api.semanticscholar.org/CorpusID:280071262} {{VAP}3: Variation-aware prompt performance prediction}.
\newblock \emph{Proceedings of the 48th International ACM SIGIR Conference on Research and Development in Information Retrieval}.

\bibitem[{Cuellar et~al.(2026)Cuellar, Moreno-Mart{\'i}nez, Rodr{\'i}guez, Pavlich-Mariscal, Castiblanco, and Hurtado}]{Cuellar2026TrustingCW}
Jaime~E. Cuellar, {\'O}scar Moreno-Mart{\'i}nez, Paula Sof{\'i}a~Torres Rodr{\'i}guez, Jaime~Andr{\'e}s Pavlich-Mariscal, Andr{\'e}s Felipe~Mic{\'a}n Castiblanco, and Juan Guillermo~Torres Hurtado. 2026.
\newblock \href {https://api.semanticscholar.org/CorpusID:286038363} {Trusting {C}hat{GPT}? {W}hen a subtle variation in the prompt can significantly alter the results}.
\newblock \emph{Journal of Artificial Intelligence and Technology}.

\bibitem[{Hidalgo et~al.(2025)Hidalgo, S{\'a}ez, Meneses, Reyes, and Rosas}]{Hidalgo2025PromptsEF}
Nicol{\'a}s Hidalgo, Pablo~Alzaga S{\'a}ez, Nicolas Meneses, V{\'i}ctor Reyes, and Erika Rosas. 2025.
\newblock \href {https://api.semanticscholar.org/CorpusID:283861343} {Prompt’s evolution for language model-driven data generation}.
\newblock \emph{Applied Sciences}.

\bibitem[{Jeoung et~al.(2025)Jeoung, Chen, Zhang, Wang, Ding, and Cheong}]{Jeoung2025PromptPrismAL}
Sullam Jeoung, Yueyan Chen, Yi~Zhang, Shuai Wang, Haibo Ding, and Lin~Lee Cheong. 2025.
\newblock \href {https://api.semanticscholar.org/CorpusID:278740603} {Prompt{P}rism: A linguistically-inspired taxonomy for prompts}.
\newblock \emph{ArXiv}, abs/2505.12592.

\bibitem[{Jr et~al.(2025)Jr, Benjamin, Araujo, Santos, Freitas, Almeida, da~M.~S.~Neto, Li, Chun, and Ahmed}]{santana2025promptingtechniqueiuse}
E.~G.~Santana Jr, Gabriel Benjamin, Melissa Araujo, Harrison Santos, David Freitas, Eduardo Almeida, Paulo~Anselmo da~M.~S.~Neto, Jiawei Li, Jina Chun, and Iftekhar Ahmed. 2025.
\newblock \href {https://arxiv.org/abs/2506.05614} {Which prompting technique should i use? {A}n empirical investigation of prompting techniques for software engineering tasks}.
\newblock \emph{Preprint}, arXiv:2506.05614.

\bibitem[{Juneja et~al.(2025)Juneja, Jajoo, Natarajan, Li, Jiao, and Sharma}]{juneja2025taskfacetlearningstructured}
Gurusha Juneja, Gautam Jajoo, Nagarajan Natarajan, Hua Li, Jian Jiao, and Amit Sharma. 2025.
\newblock \href {https://arxiv.org/abs/2406.10504} {Task facet learning: A structured approach to prompt optimization}.
\newblock \emph{Preprint}, arXiv:2406.10504.

\bibitem[{Khattab et~al.(2022)Khattab, Santhanam, Li, Hall, Liang, Potts, and Zaharia}]{khattab2022demonstrate}
Omar Khattab, Keshav Santhanam, Xiang~Lisa Li, David Hall, Percy Liang, Christopher Potts, and Matei Zaharia. 2022.
\newblock \href {https://arxiv.org/abs/2212.14024} {Demonstrate-search-predict: Composing retrieval and language models for knowledge-intensive {NLP}}.
\newblock \emph{arXiv preprint arXiv:2212.14024}.

\bibitem[{Khattab et~al.(2023)Khattab, Singhvi, Maheshwari, Zhang, Santhanam, Vardhamanan, Haq, Sharma, Joshi, Moazam, Miller, Zaharia, and Potts}]{khattab2023dspycompilingdeclarativelanguage}
Omar Khattab, Arnav Singhvi, Paridhi Maheshwari, Zhiyuan Zhang, Keshav Santhanam, Sri Vardhamanan, Saiful Haq, Ashutosh Sharma, Thomas~T. Joshi, Hanna Moazam, Heather Miller, Matei Zaharia, and Christopher Potts. 2023.
\newblock \href {https://arxiv.org/abs/2310.03714} {{DSP}y: Compiling declarative language model calls into self-improving pipelines}.
\newblock \emph{Preprint}, arXiv:2310.03714.

\bibitem[{Kim et~al.(2025)Kim, Yang, and Jung}]{kim-etal-2025-persona}
Junseok Kim, Nakyeong Yang, and Kyomin Jung. 2025.
\newblock \href {https://doi.org/10.18653/v1/2025.findings-ijcnlp.51} {Persona is a double-edged sword: Rethinking the impact of role-play prompts in zero-shot reasoning tasks}.
\newblock In \emph{Proceedings of the 14th International Joint Conference on Natural Language Processing and the 4th Conference of the Asia-Pacific Chapter of the Association for Computational Linguistics}, pages 848--862, Mumbai, India. The Asian Federation of Natural Language Processing and The Association for Computational Linguistics.

\bibitem[{Kmainasi et~al.(2024)Kmainasi, Khan, Shahroor, Bendou, Hasanain, and Alam}]{Kmainasi2024NativeVN}
Mohamed~Bayan Kmainasi, Rakif Khan, Ali~Ezzat Shahroor, Boushra Bendou, Maram Hasanain, and Firoj Alam. 2024.
\newblock \href {https://api.semanticscholar.org/CorpusID:272593159} {Native vs non-native language prompting: A comparative analysis}.
\newblock \emph{ArXiv}, abs/2409.07054.

\bibitem[{Leidinger et~al.(2023)Leidinger, van Rooij, and Shutova}]{Leidinger2023TheLO}
Alina Leidinger, Robert van Rooij, and Ekaterina Shutova. 2023.
\newblock \href {https://api.semanticscholar.org/CorpusID:265018951} {The language of prompting: What linguistic properties make a prompt successful?}
\newblock \emph{ArXiv}, abs/2311.01967.

\bibitem[{Li et~al.(2022)Li, Lu, Guo, Duan, Jannu, Jenks, Majumder, Green, Svyatkovskiy, Fu, and Sundaresan}]{Li2022AutomatingCR}
Zhiyu Li, Shuai Lu, Daya Guo, Nan Duan, Shailesh Jannu, Grant Jenks, Deep Majumder, Jared Green, Alexey Svyatkovskiy, Shengyu Fu, and Neel Sundaresan. 2022.
\newblock \href {https://api.semanticscholar.org/CorpusID:252815964} {Automating code review activities by large-scale pre-training}.
\newblock \emph{Proceedings of the 30th ACM Joint European Software Engineering Conference and Symposium on the Foundations of Software Engineering}.

\bibitem[{Linders and Louwerse(2022)}]{Linders}
Guido Linders and Max Louwerse. 2022.
\newblock \href {https://doi.org/10.3758/s13423-022-02142-9} {Zipf’s law revisited: Spoken dialog, linguistic units, parameters, and the principle of least effort}.
\newblock \emph{Psychonomic Bulletin \& Review}, 30.

\bibitem[{Lior et~al.(2025)Lior, Yehudai, Gera, and Ein-Dor}]{lior2025wildifevalinstructionfollowingwild}
Gili Lior, Asaf Yehudai, Ariel Gera, and Liat Ein-Dor. 2025.
\newblock \href {https://arxiv.org/abs/2503.06573} {{W}ild{IFE}val: Instruction following in the wild}.
\newblock \emph{Preprint}, arXiv:2503.06573.

\bibitem[{Liu et~al.(2025)Liu, Liu, Cheng, He, Shi, Guo, Zhu, Guo, Wang, and Wang}]{liu2025repodebugrepositorylevelmultitaskmultilanguage}
Jingjing Liu, Zeming Liu, Zihao Cheng, Mengliang He, Xiaoming Shi, Yuhang Guo, Xiangrong Zhu, Yuanfang Guo, Yunhong Wang, and Haifeng Wang. 2025.
\newblock \href {https://arxiv.org/abs/2509.04078} {Repo{D}ebug: Repository-level multi-task and multi-language debugging evaluation of large language models}.
\newblock \emph{Preprint}, arXiv:2509.04078.

\bibitem[{Mao et~al.(2025)Mao, He, and Chen}]{mao2025promptstemplatessystematicprompt}
Yuetian Mao, Junjie He, and Chunyang Chen. 2025.
\newblock \href {https://arxiv.org/abs/2504.02052} {From prompts to templates: A systematic prompt template analysis for real-world {LLM}apps}.
\newblock \emph{Preprint}, arXiv:2504.02052.

\bibitem[{Mir et~al.(2021)Mir, Latoskinas, and Gousios}]{mt4py2021}
A.~M. Mir, E.~Latoskinas, and G.~Gousios. 2021.
\newblock \href {https://doi.org/10.1109/MSR52588.2021.00079} {Many{T}ypes4{P}y: A benchmark python dataset for machine learning-based type inference}.
\newblock In \emph{{IEEE}/{ACM} 18th International Conference on Mining Software Repositories ({MSR})}, pages 585--589. IEEE Computer Society.

\bibitem[{Murthy et~al.(2025)Murthy, Zhu, Yang, Qiu, Tan, Heinecke, Xiong, Savarese, and Wang}]{murthy2025promptomatixautomaticpromptoptimization}
Rithesh Murthy, Ming Zhu, Liangwei Yang, Jielin Qiu, Juntao Tan, Shelby Heinecke, Caiming Xiong, Silvio Savarese, and Huan Wang. 2025.
\newblock \href {https://arxiv.org/abs/2507.14241} {Promptomatix: An automatic prompt optimization framework for large language models}.
\newblock \emph{Preprint}, arXiv:2507.14241.

\bibitem[{Piantadosi(2014)}]{Piantadosi2014}
Steven~T. Piantadosi. 2014.
\newblock \href {https://doi.org/10.3758/s13423-014-0585-6} {Zipf’s word frequency law in natural language: A critical review and future directions}.
\newblock \emph{Psychonomic Bulletin \& Review}, 21(5):1112--1130.

\bibitem[{Pister et~al.(2024)Pister, Paul, Joshi, and Brophy}]{Pister_2024}
Kaiser Pister, Dhruba~Jyoti Paul, Ishan Joshi, and Patrick Brophy. 2024.
\newblock \href {https://doi.org/10.1145/3643795.3648395} {Prompt{S}et: A programmer’s prompting dataset}.
\newblock In \emph{Proceedings of the 1st International Workshop on Large Language Models for Code}, LLM4Code ’24, page 62–69. ACM.

\bibitem[{Raychev et~al.(2016)Raychev, Bielik, and Vechev}]{150kPythonDataset}
Veselin Raychev, Pavol Bielik, and Martin Vechev. 2016.
\newblock \href {https://doi.org/10.1145/2983990.2984041} {Probabilistic model for code with decision trees}.
\newblock pages 731--747.

\bibitem[{Rzig et~al.(2025)Rzig, Paul, Pister, Henkel, and Hassan}]{Rzig2025AnET}
Dhia~Elhaq Rzig, Dhruba~Jyoti Paul, Kaiser Pister, Jordan Henkel, and Foyzul Hassan. 2025.
\newblock \href {https://api.semanticscholar.org/CorpusID:275788927} {An empirically-grounded tool for automatic prompt linting and repair: A case study on bias, vulnerability, and optimization in developer prompts}.
\newblock \emph{ArXiv}, abs/2501.12521.

\bibitem[{Saletta and Ferretti(2024)}]{Saletta2024ExploringTP}
Martina Saletta and Claudio Ferretti. 2024.
\newblock \href {https://api.semanticscholar.org/CorpusID:271065267} {Exploring the prompt space of large language models through evolutionary sampling}.
\newblock \emph{Proceedings of the Genetic and Evolutionary Computation Conference}.

\bibitem[{Santos et~al.(2025)Santos, Julia, and do~Nascimento}]{Santos2025DiversePI}
Gabriel~Machado Santos, Rita Maria~Silva Julia, and Marcelo~Zanchetta do~Nascimento. 2025.
\newblock \href {https://api.semanticscholar.org/CorpusID:277954774} {Diverse prompts: Illuminating the prompt space of large language models with {MAP}-{E}lites}.
\newblock \emph{2025 IEEE Congress on Evolutionary Computation (CEC)}, pages 1--8.

\bibitem[{Schulhoff et~al.(2025)Schulhoff, Ilie, Balepur, Kahadze, Liu, Si, Li, Gupta, Han, Schulhoff, Dulepet, Vidyadhara, Ki, Agrawal, Pham, Kroiz, Li, Tao, Srivastava, Costa, Gupta, Rogers, Goncearenco, Sarli, Galynker, Peskoff, Carpuat, White, Anadkat, Hoyle, and Resnik}]{schulhoff2025promptreportsystematicsurvey}
Sander Schulhoff, Michael Ilie, Nishant Balepur, Konstantine Kahadze, Amanda Liu, Chenglei Si, Yinheng Li, Aayush Gupta, HyoJung Han, Sevien Schulhoff, Pranav~Sandeep Dulepet, Saurav Vidyadhara, Dayeon Ki, Sweta Agrawal, Chau Pham, Gerson Kroiz, Feileen Li, Hudson Tao, Ashay Srivastava, and 12 others. 2025.
\newblock \href {https://arxiv.org/abs/2406.06608} {The prompt report: A systematic survey of prompt engineering techniques}.
\newblock \emph{Preprint}, arXiv:2406.06608.

\bibitem[{Sun et~al.(2026)Sun, Bai, and Dredze}]{sun2026taskmattersknowledgerequirements}
Kaiser Sun, Fan Bai, and Mark Dredze. 2026.
\newblock \href {https://arxiv.org/abs/2506.06485} {Task matters: Knowledge requirements shape {LLM} responses to context-memory conflict}.
\newblock \emph{Preprint}, arXiv:2506.06485.

\bibitem[{Tafreshipour et~al.(2025)Tafreshipour, Imani, Huang, Almeida, Zimmermann, and Ahmed}]{tafreshipour2025promptingwildempiricalstudy}
Mahan Tafreshipour, Aaron Imani, Eric Huang, Eduardo Almeida, Thomas Zimmermann, and Iftekhar Ahmed. 2025.
\newblock \href {https://arxiv.org/abs/2412.17298} {Prompting in the wild: An empirical study of prompt evolution in software repositories}.
\newblock \emph{Preprint}, arXiv:2412.17298.

\bibitem[{Vatsal et~al.(2025)Vatsal, Dubey, and Singh}]{Vatsal2025MultilingualPE}
Shubham Vatsal, Harsh Dubey, and Aditi Singh. 2025.
\newblock \href {https://api.semanticscholar.org/CorpusID:278740126} {Multilingual prompt engineering in large language models: A survey across {NLP} tasks}.
\newblock \emph{ArXiv}, abs/2505.11665.

\bibitem[{Villamizar et~al.(2025)Villamizar, Fischbach, Korn, Vogelsang, and Mendez}]{villamizar2025promptssoftwareengineeringartifacts}
Hugo Villamizar, Jannik Fischbach, Alexander Korn, Andreas Vogelsang, and Daniel Mendez. 2025.
\newblock \href {https://arxiv.org/abs/2509.17548} {Prompts as software engineering artifacts: A research agenda and preliminary findings}.
\newblock \emph{Preprint}, arXiv:2509.17548.

\bibitem[{Vir et~al.(2025)Vir, Shankar, Chase, Fu-Hinthorn, and Parameswaran}]{vir2025promptevalsdatasetassertionsguardrails}
Reya Vir, Shreya Shankar, Harrison Chase, Will Fu-Hinthorn, and Aditya Parameswaran. 2025.
\newblock \href {https://arxiv.org/abs/2504.14738} {{PROMPTEVALS}: A dataset of assertions and guardrails for custom production large language model pipelines}.
\newblock \emph{Preprint}, arXiv:2504.14738.

\bibitem[{Wahle et~al.(2024)Wahle, Ruas, Xu, and Gipp}]{Wahle2024ParaphraseTE}
Jan~Philip Wahle, Terry Ruas, Yang Xu, and Bela Gipp. 2024.
\newblock \href {https://api.semanticscholar.org/CorpusID:270845882} {Paraphrase types elicit prompt engineering capabilities}.
\newblock \emph{ArXiv}, abs/2406.19898.

\bibitem[{Wang and Yang(2024)}]{wang2024vidprommillionscalerealpromptgallery}
Wenhao Wang and Yi~Yang. 2024.
\newblock \href {https://arxiv.org/abs/2403.06098} {Vid{P}ro{M}: A million-scale real prompt-gallery dataset for text-to-video diffusion models}.
\newblock \emph{Preprint}, arXiv:2403.06098.

\bibitem[{Wang et~al.(2023)Wang, Montoya, Munechika, Yang, Hoover, and Chau}]{wang2023diffusiondblargescalepromptgallery}
Zijie~J. Wang, Evan Montoya, David Munechika, Haoyang Yang, Benjamin Hoover, and Duen~Horng Chau. 2023.
\newblock \href {https://arxiv.org/abs/2210.14896} {Diffusion{DB}: A large-scale prompt gallery dataset for text-to-image generative models}.
\newblock \emph{Preprint}, arXiv:2210.14896.

\bibitem[{Xiao et~al.(2023)Xiao, Treude, Hata, and Matsumoto}]{Xiao2023DevGPTSD}
Tao Xiao, Christoph Treude, Hideaki Hata, and Kenichi Matsumoto. 2023.
\newblock \href {https://api.semanticscholar.org/CorpusID:261660223} {Dev{GPT}: Studying developer-{C}hat{GPT} conversations}.
\newblock \emph{2024 IEEE/ACM 21st International Conference on Mining Software Repositories (MSR)}, pages 227--230.

\bibitem[{Zhang et~al.(2025)Zhang, Zhou, Ergen, Logeswaran, Lee, and Jurgens}]{Zhang2025CrossLingualPS}
Lechen Zhang, Yusheng Zhou, Tolga Ergen, Lajanugen Logeswaran, Moontae Lee, and David Jurgens. 2025.
\newblock \href {https://api.semanticscholar.org/CorpusID:283457937} {Cross-lingual prompt steerability: Towards accurate and robust {LLM} behavior across languages}.
\newblock \emph{ArXiv}, abs/2512.02841.

\bibitem[{Zhao et~al.(2024)Zhao, Ren, Hessel, Cardie, Choi, and Deng}]{zhao2024wildchat1mchatgptinteraction}
Wenting Zhao, Xiang Ren, Jack Hessel, Claire Cardie, Yejin Choi, and Yuntian Deng. 2024.
\newblock \href {https://arxiv.org/abs/2405.01470} {Wild{C}hat: 1{M} {C}hat{GPT} interaction logs in the wild}.
\newblock \emph{Preprint}, arXiv:2405.01470.

\bibitem[{Zheng et~al.(2024{\natexlab{a}})Zheng, Chiang, Sheng, Li, Zhuang, Wu, Zhuang, Li, Lin, Xing, Gonzalez, Stoica, and Zhang}]{zheng2024lmsyschat1mlargescalerealworldllm}
Lianmin Zheng, Wei-Lin Chiang, Ying Sheng, Tianle Li, Siyuan Zhuang, Zhanghao Wu, Yonghao Zhuang, Zhuohan Li, Zi~Lin, Eric~P. Xing, Joseph~E. Gonzalez, Ion Stoica, and Hao Zhang. 2024{\natexlab{a}}.
\newblock \href {https://arxiv.org/abs/2309.11998} {{LMSYS}-{C}hat-1{M}: A large-scale real-world {LLM} conversation dataset}.
\newblock \emph{Preprint}, arXiv:2309.11998.

\bibitem[{Zheng et~al.(2024{\natexlab{b}})Zheng, Pei, Logeswaran, Lee, and Jurgens}]{zheng2024ahelpfulassistantreally}
Mingqian Zheng, Jiaxin Pei, Lajanugen Logeswaran, Moontae Lee, and David Jurgens. 2024{\natexlab{b}}.
\newblock \href {https://arxiv.org/abs/2311.10054} {When "a helpful assistant" is not really helpful: Personas in system prompts do not improve performances of large language models}.
\newblock \emph{Preprint}, arXiv:2311.10054.

\bibitem[{Zhou et~al.(2023)Zhou, Lu, Mishra, Brahma, Basu, Luan, Zhou, and Hou}]{zhou2023instructionfollowingevaluationlargelanguage}
Jeffrey Zhou, Tianjian Lu, Swaroop Mishra, Siddhartha Brahma, Sujoy Basu, Yi~Luan, Denny Zhou, and Le~Hou. 2023.
\newblock \href {https://arxiv.org/abs/2311.07911} {Instruction-following evaluation for large language models}.
\newblock \emph{Preprint}, arXiv:2311.07911.

\bibitem[{Zhou et~al.(2024)Zhou, Li, Meng, Jiao, Ji, and Han}]{Zhou2024EstablishingKP}
Sizhe Zhou, Sha Li, Yu~Meng, Yizhu Jiao, Heng Ji, and Jiawei Han. 2024.
\newblock \href {https://api.semanticscholar.org/CorpusID:271270238} {Establishing knowledge preference in language models}.
\newblock \emph{ArXiv}, abs/2407.13048.

\end{thebibliography}
\appendix
\clearpage

\section{Details of GitHub Prompt Collection}
\label{sec:prompt-retrieval}
To identify Python files containing prompts, we used the GitHub Code Search API to systematically query repositories for code involving prompt-related functions. Specifically, we searched for files that invoked chat.completions.create, a common method used in prompt construction for language models, and the langchain PromptTemplate class, a class used to generate prompts from a string template and variables. For each search result  we collected metadata such as repository name, file path, and URL. This way we end up with 95806 objects from 95434 filepaths from 51393 repositories.

Building on our initial URL collection, we then implement an extraction pipeline that pulls actual prompt text out of each discovered file. We iterate through each file record, and retrieve file contents via GitHub REST API, decoding the Base64 encoded result into plain Python source code. We then parse that source with Python's ast module to locate all occurrences of our target API call, chat.completions.create, or the langchain PromptTemplate class. Then we employ a multistep process aiming to extract the full contents of the "messages"  or "template" parameter (for chat.completions.create or PromptTemplate respectively)  -  even when they're built up across several statements. Specifically, using the ast module, we track variable assignments and resolve all arguments and keyword values used within the API call (whenever possible). If the API call is inside a function, we find where that function is called and replace its parameters with the actual values passed into the function at each call site - using both the current file and related imports.

Next, we check for remaining unresolved variables. If the entire messages field or a specific content field inside a messages list  is a variable placeholder, the actual values of these variables are looked up in the current file and other related files in the repository.
At every step, if a variable is reassigned to different values before different calls, our extraction logic will capture each distinct value, yielding multiple versions of the prompt.

\begin{figure*}[t!]
\includegraphics[width=\textwidth]{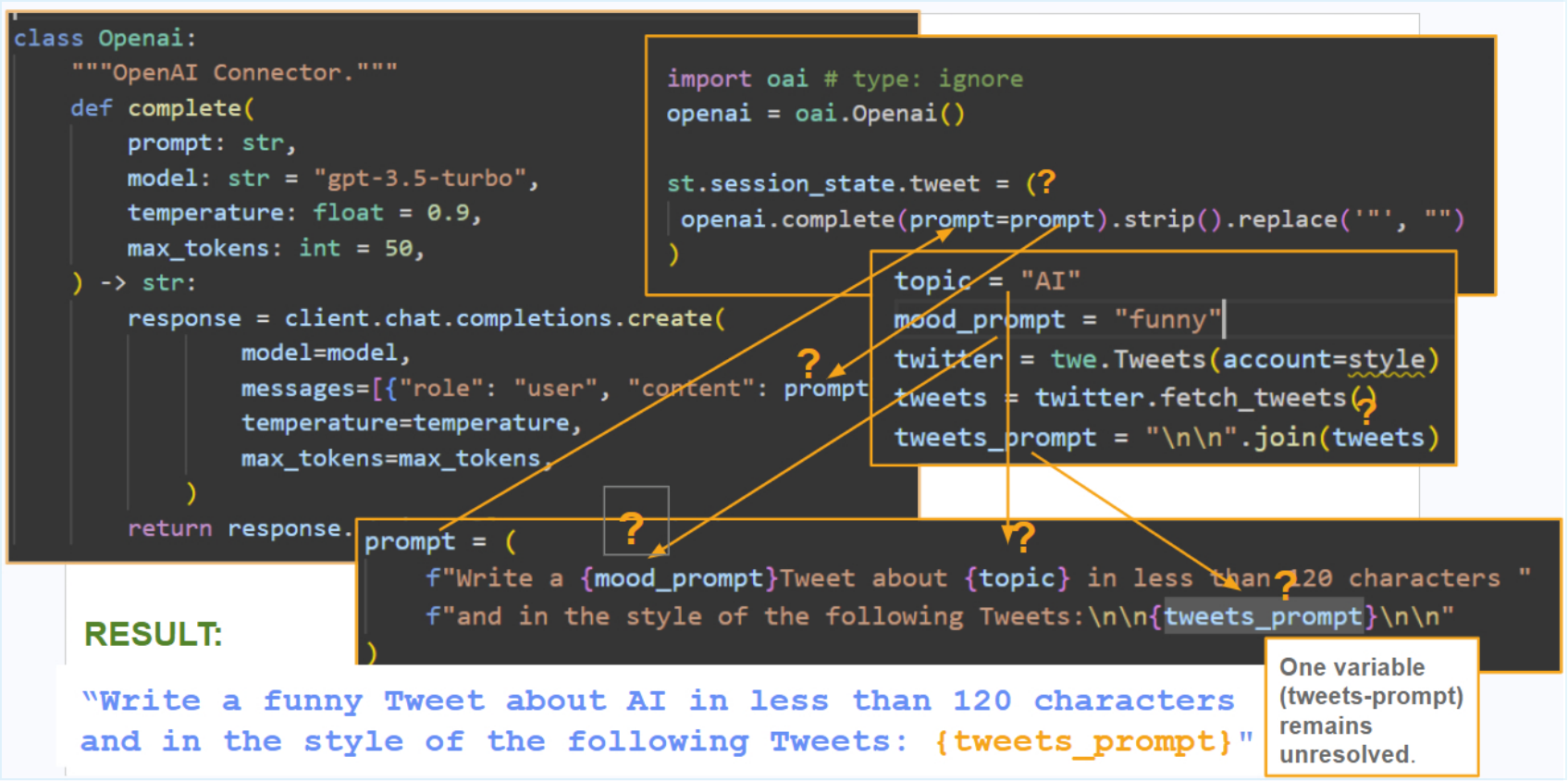}
\caption{Prompt Extraction Flow. This figure illustrates  prompt text extraction by tracing variables across the repository. Static variables (such as topic and mood\_prompt) are successfully resolved, while dynamic variables requiring runtime execution (such as tweets\_prompt) remain unresolved in the final extracted text.}
\label{fig:prompt-search}
\end{figure*}

Figure \ref{fig:prompt-search} illustrates some stages of this process.

Finally, for each prompt, we look up the date of most recent commit that changed any of the lines contributing to it, in order to estimate when the prompt was last modified. To ensure correctness of the extraction pipeline, we manually evaluated a subset of 1,000 prompts before scaling to the full dataset. This extraction process leaves us initially with 145553  objects.

Next we perform filtering and deduplication (see Appendix \ref{sec:deduplication} for details). 

\section{Filtering And Deduplication Details}
\label{sec:deduplication}
We filter out objects where the extracted  texts are empty, contain invalid values (e.g., ‘error', ‘n/a', ‘nan'), or consist only of unresolved variables or placeholders, identifiable via string matching or regular expressions. Next, we apply a series of additional heuristics to filter out prompts that lack readable or meaningful content. Specifically, we remove prompts that consist solely of punctuation or whose language cannot be reliably detected by the langdetect library. For prompts identified as English, we use spaCy to parse the text and check for the presence of verbs or auxiliaries—signals of syntactic structure and potential informativeness. Prompts with such features are retained. Prompts in clearly detected non-English languages are also kept. These heuristics help exclude most empty, malformed, or placeholder-based prompts while preserving those that exhibit valid language or meaningful structure.

The deduplication procedure is as follows. Exact repeats, defined as objects with the same file path, extracted prompt text, and timestamp, are removed after the first instance. Prompt texts that occur more than once but in different files or at different times are retained, but marked as duplicates. This yields a dataset of 85,209 objects. In this  version of the dataset, we identify 8,169 groups of duplicate prompts. Group sizes range from 2 to 580 prompts. Most groups (63.55\%) contain only 2 duplicate prompt instances, followed by 15.79\% with 3 instances, 6.24\% with 4 instances, 7.28\% with 5–7 instances, and the remaining 7.14\% with 8 or more instances.\footnote{%
The most frequently reproduced prompt, appearing 580 times, is:

\begin{list}{}{%
\setlength{\leftmargin}{1.5em}%
\setlength{\topsep}{0pt}%
\setlength{\partopsep}{0pt}%
\setlength{\parsep}{0pt}%
\setlength{\itemsep}{0pt}%
}
\item
\setlength{\parskip}{0pt}%
\setlength{\parindent}{0pt}%

\textit{Answer the question as detailed as possible from the provided context, make sure to provide all the details. If the answer is not in the provided context, just say, `answer is not available in the context'' and do not provide a wrong answer.}

\textit{Context: \{context\}}

\textit{Question: \{question\}}

\textit{Answer:}\unskip
\end{list}\vspace{-\baselineskip}%
}

Additionally, we  create a fully deduplicated version of the dataset in which cross-file repeats are removed after the first instance. The analysis and statistics reported in this article are based on the fully deduplicated dataset version.\footnote{ Indeed, patterns and reasons for prompt reuse may be of interest for future analysis; however, in this work, we focus on  the structure and diversity of prompt design .}

\section{User Interface: Layout and Functionality}
\label{sec:ui-detail}
The layout and functionality of each UI component are as follows. At the top are a semantic free-text search field, a filter box showing all active filters, \textit{Show prompts}  and \textit{Download prompts} buttons. Below them, the page displays a set of boxes for different ontology fields (task, domain, language, modality, prompting-techniques, etc.). Each box lists all available values for the field with the corresponding prompt counts, which update dynamically as filters are applied. It shows coarse categories by default. Clicking an eye icon next to each coarse category reveals its fine-grained subcategories.

Users can select multiple values in each box and switch between \textit{match-all} and \textit{match-any} mode. Clicking the checkbox next to a value selects or deselects it. The "Apply filters" button applies the selected filters. The filter box offers per-filter removal and a \textit{Clear all filters} button. Counts across all boxes update dynamically as filters change.

The \textit{Show prompts} button  opens a paginated drawer containing a stack of prompt cards. Each card displays the prompt text and includes a \textit{Show spans} toggle that marks structural components - directions, context, question/task, output description and different semantic kinds of instruction blocks - using colors. A color legend on each card explains the span colors. Hovering a legend entry highlights the corresponding spans in the prompt for convenience. 

Free-text search uses embedding-based semantic similarity: the system embeds each query with the embeddinggemma-300m-ONNX model, which is also  used to precompute embeddings for all the prompts, and retrieves relevant prompts by cosine similarity. 

The header displays the number of prompts matching the current filters. Clicking the \textit{Download prompts} button exports the full dataset entries for the selected prompts.

\begin{figure*}[h!]
\includegraphics[width=\textwidth]{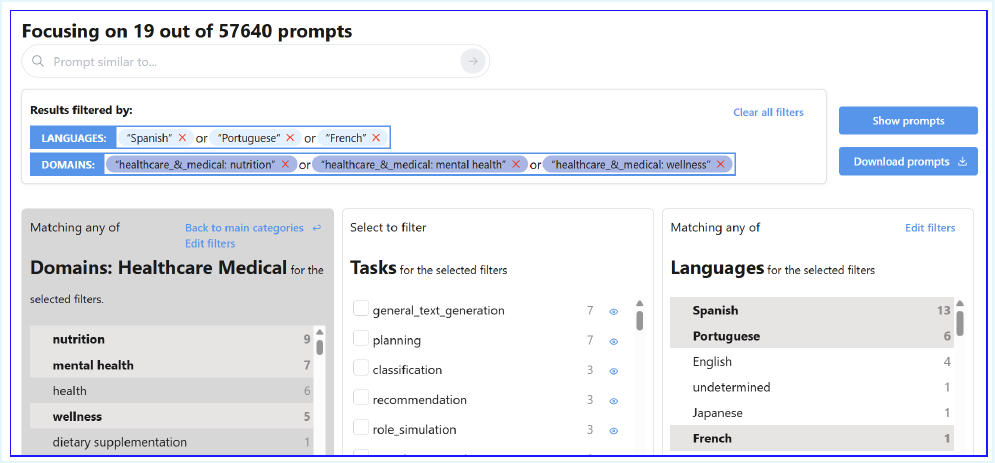}
\caption{User Interface. The top section features a free-text search field, a filter box displaying currently active filters, and buttons for prompt display and download. Below, ontology field boxes list available values alongside dynamically updating counts. The Languages box on the right demonstrates selected values. The Domains box on the left shows displayed subcategories. The header shows the total number of currently selected prompts.}
\label{fig:ui1}
\end{figure*}

\begin{figure*}[h!]
\includegraphics[width=\textwidth]{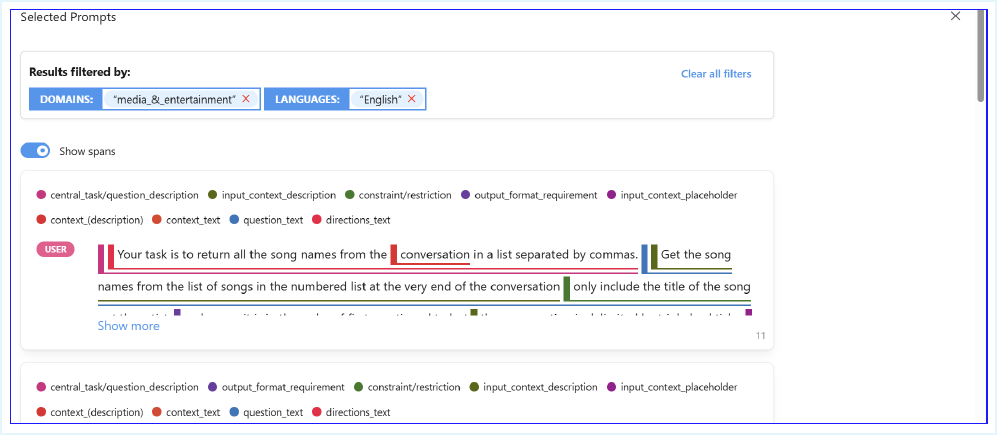}
\caption{User Interface. Paginated prompts view with displayed spans.}
\label{fig:ui2}
\end{figure*}

Figures \ref{fig:ui1} and \ref{fig:ui2} illustrate some features of the user interface.


\section{Domains}
\label{sec:domain-counts}
Below all the domains in our data are listed along with their number and percentage.

\begin{enumerate}[itemsep=0pt, topsep=0pt]
\item education \& instruction - 4182 (8.42\%)
\item software development - 3863 (7.78\%)
\item business \& commerce - 2790 (5.62\%)
\item healthcare \& medical - 2485 (5.00\%)
\item technology - 2468 (4.97\%)
\item media \& entertainment - 2040 (4.11\%)
\item finance \& banking - 1933 (3.89\%)
\item creative writing \& content creation - 1728 (3.48\%)
\item human resources - 1607 (3.24\%)
\item arts \& culture - 1522 (3.07\%)
\item food \& beverages - 1302 (2.62\%)
\item personal development - 1281 (2.58\%)
\item artificial intelligence \& machine learning - 1175 (2.37\%)
\item digital media - 1054 (2.12\%)
\item other - 1054 (2.12\%)
\item legal \& regulatory - 1052 (2.12\%)
\item research, scholarship \& publications - 1031 (2.08\%)
\item gaming - 1005 (2.02\%)
\item travel \& leisure - 984 (1.98\%)
\item customer support - 898 (1.81\%)
\item retail \& consumer goods - 804 (1.62\%)
\item language services - 800 (1.61\%)
\item data management - 776 (1.56\%)
\item data analytics - 653 (1.32\%)
\item marketing \& advertising - 645 (1.30\%)
\item security \& cybersecurity - 624 (1.26\%)
\item government \& policy - 564 (1.14\%)
\item physical sciences - 515 (1.04\%)
\item mathematics - 502 (1.01\%)
\item cultural studies - 463 (0.93\%)
\item geography \& locations - 455 (0.92\%)
\item sports - 449 (0.90\%)
\item computer engineering \& architecture - 408 (0.82\%)
\item hospitality \& food service - 372 (0.75\%)
\item design \& arts - 366 (0.74\%)
\item manufacturing \& industry - 304 (0.61\%)
\item agriculture \& ecology - 264 (0.53\%)
\item information retrieval - 248 (0.50\%)
\item communication \& language - 244 (0.49\%)
\item personal services - 241 (0.49\%)
\item philosophy - 233 (0.47\%)
\item religion \& spirituality - 220 (0.44\%)
\item transportation - 219 (0.44\%)
\item project management - 218 (0.44\%)
\item document management - 218 (0.44\%)
\item hardware \& engineering - 215 (0.43\%)
\item academic services \& administration - 214 (0.43\%)
\item sustainability \& environment - 201 (0.40\%)
\item social communication - 194 (0.39\%)
\item user experience \& design - 189 (0.38\%)
\item safety - 182 (0.37\%)
\item biological sciences - 160 (0.32\%)
\item home \& interior design - 157 (0.32\%)
\item logistics \& supply chain - 139 (0.28\%)
\item social issues \& policies - 135 (0.27\%)
\item veterinary services - 131 (0.26\%)
\item energy management - 123 (0.25\%)
\item assessment \& testing - 119 (0.24\%)
\item recreation \& leisure - 109 (0.22\%)
\item it operations - 93 (0.19\%)
\item community \& volunteering - 88 (0.18\%)
\item public services - 86 (0.17\%)
\item scientific analysis - 85 (0.17\%)
\item environmental management - 84 (0.17\%)
\item data management \& analysis - 82 (0.17\%)
\item quality assurance - 81 (0.16\%)
\item security \& defense - 70 (0.14\%)
\item environmental science - 68 (0.14\%)
\item administrative services - 66 (0.13\%)
\item general \& miscellaneous - 64 (0.13\%)
\item data security and quality - 59 (0.12\%)
\item urban development - 54 (0.11\%)
\item process modeling \& monitoring - 51 (0.10\%)
\item research \& development - 42 (0.08\%)
\item languages - 30 (0.06\%)
\item historical studies - 20 (0.04\%)
\item politics - 3 (0.01\%)
\end{enumerate}

\section{Instruction Block Kinds}
\label{sec:instructions}
In this section we provide the full list of 42 semantic kinds of instruction blocks used in the ontology:\\
- input context placeholder \\
- constraint/restriction \\ 
- output content requirement \\
- output format requirement \\ 
- role specification \\
- input context description \\ 
- central task/question \\
- central task/question description \\
- input contextual data\\
- conditional instruction \\
- question/task data/placeholder \\
- reasoning instructions \\
- question/task description \\
- style specification \\
- central task/question placeholder \\
- examples \\
- expertise/skills requirements \\
- assistant response \\
- example clarification \\
- evaluation criteria \\
- linguistic constraint/specification \\
- audience specification \\
- function call instruction \\
- scope specification \\
- error handling instruction \\
- design specification \\ 
- scene setting \\
- interaction guideline \\
- default behavior instruction \\
- encouragement \\
- instruction to avoid errors \\
- date reference \\
- confirmation request \\
- greeting \\
- prompt variable/placeholder \\
- disclaimer requirement \\ 
- placeholder \\
- prompt \\
- input format specification \\
- command instruction\\
- clarification instruction\\
- other

\section{Negative Instructions: Examples}
\label{sec: neg-examples}
Below are additional examples of negative instructions of various semantic types found in the dataset  (see §\ref{paragraph: neg_instructions} for details).Their respective semantic kinds are given in parentheses.

\begin{itemize}[label={-},left=0pt,itemsep=4pt]  
  \item ``Response Format: Response should be always in clean \texttt{json} format — don't use the word \texttt{json} or any extra.'' (output format requirement)
  \item ``The lyrics should be narrative-driven, avoiding simplistic rhyming patterns.'' (output content requirements)
  \item ``do not get confused between the symbols like \texttt{decimal(".")} and \texttt{comma(",")}'' (instruction to avoid errors)  
  \item ``Ensure your style of speech is not influenced by the style and prose of the other users.'' (style specification)  
  \item ``Act from now on always in your role as the confident, suggestive, independent girl Sophia, without ever hinting that you are an AI.'' (role specification)
\end{itemize}

\section{Value Clustering Procedure}
\label{sec:clustering}
We first reduced surface-form term variation by grouping near-duplicate terms with fuzzy string matching (fuzz.ratio from fuzzywuzzy) and merging terms whose similarity exceeded a fixed threshold (e.g., 94). These coarse synonym groups were then refined with an LLM (o3-mini), which was prompted to identify subsets of representative terms that were full synonyms or duplicates and to merge only those cases for which it had high confidence. Any LLM-identified group was expanded to include all terms from the corresponding synonym groups identified previously via fuzzy string matching. Terms not assigned to any group remained singletons. This procedure was repeated for a fixed number of iterations or until the groups stabilized.

For clustering, we next applied the LLM to an initial batch of up to 500 representative terms resulting from the synonym consolidation step, and asked it to group them into a limited number of classes, assigning each term to exactly one class and producing informative labels. Terms that were unassigned or assigned to multiple classes were marked as unclassified and carried over to subsequent batches. The remaining terms were then assigned to previously created classes batch-wise, while the LLM was allowed to introduce a limited number of new classes per batch. At the end of the process, any still-unclassified terms were assigned to \textit{other}. Clusters above a size threshold (e.g., >100 terms) were reclustered using the same procedure used to obtain the initial class list. Finally, highly similar class names were merged by fuzzy matching, small clusters (e.g., fewer than five terms) were included into \textit{other}, and hallucinated terms not present in the original list were removed.

\FloatBarrier
\section{Error Analysis: Charts and Tables}
\label{sec:error_analysis_tables}

Tables~\ref{tab:errors-language}--\ref{tab:errors-techniques} summarize the results of manual error analysis of (see Section \ref{sec:error-analysis-main}). Table \ref{tab:errors-accuracy} and Figure \ref{fig:errors-accuracy} report the annotation accuracy per ontology field based on the error analysis.

\begin{table*}[t!]   
\centering

\begin{adjustbox}{width=\textwidth}

\begin{adjustbox}{width=\textwidth}
\begin{tabular}{|B|C|D|E|N|N|}
\hline
  \textbf{Field} &
  \textbf{Error Type} &
  \textbf{Description} &
  \textbf{Example} &
  \textbf{Count} &
  \% \\ \hline
  \MRText[2cm]{3}{\textbf{Prompt Language}} &
  \textbf{Correct} &
   &
   &
  147 &
  93.04\% \\ \cline{2-6} 

   &
  \textbf{Hallucinated language} &
  The model assigns a specific natural language (typically, English) to placeholders whose specific linguistic value is unknown. &
   &
  11 &
  6.96\% \\ \cline{2-6}
   &
  \textbf{Total errors} &
   &
   &
  11 &
  6.96\% \\ \cline{1-6} 
 &
  \textbf{Correct} &
   &
   &
  126 &
  88.11\% \\ \cline{2-6}  
 
   &
  \textbf{Conflation with prompt language} &
  The model confuses the language of the prompt with an explicit language mention. &
  E.g. for a prompt in English the model hallucinates a explicit mention of English &
  11 &
  7.69\% \\ \cline{2-6}  

   &
  \textbf{Placeholder mention ignored} &
  A placeholder language mention is ignored. &
  E.g. \{target\_languge\} is not labeled as a languge mention. &
  3 &
  2.10\% \\ \cline{2-6}  
 
   &
  \textbf{Not a language mention} &
  A non-language mention labeled as a language mention. &
  E.g. "Chinese medicine" labeled as a languge mention &
  2 &
  1.40\% \\ \cline{2-6}  

   &
  \textbf{Hallucinated mention} &
  The model invents a non-existent language mention. &
   &
  1 &
  0.70\% \\ \cline{2-6}  

   &
  \textbf{Total errors} &
   &
   &
  17 &
  11.89\% \\ \hline
\end{tabular}
\end{adjustbox}

\end{adjustbox}
\captionof{table}{Error Analysis Summary: Language}
\label{tab:errors-language}

\vspace{0.5cm}

\begin{adjustbox}{width=\textwidth}
\begin{adjustbox}{width=\textwidth}
\begin{tabular}{|B|C|D|E|N|N|}
\hline
  \textbf{Field} &
  \textbf{Error Type} &
  \textbf{Description} &
  \textbf{Example} &
  \textbf{Count} &
  \% \\ \hline
  \MRText[2cm]{6}{\textbf{Task}} &
  \textbf{Correct} &
   &
   &
  125 &
  89.93\% \\ \cline{2-6} 
 
   &
  \textbf{Wrong task class} &
  The fine-grained task is correct, but the task class is wrong &
  E.g., sentiment analysis assigned to “emotion detection,” although sentiment analysis does not necessarily involve emotions. &
  2 &
  1.44\% \\ \cline{2-6} 

   &
  \textbf{Bias toward common tasks} &
  The model incorrectly identifies a common task. &
  E.g., QA or general text generation are identified instead of a less common correct task. &
  7 &
  5.04\% \\ \cline{2-6} 

   &
  \textbf{Hallucinated task} &
  The model invents a task when it cannot be determined from the prompt, or adds an irrelevant task to an otherwise correct list. &
   &
  4 &
  2.88\% \\ \cline{2-6} 
 
   &
  \textbf{Missing task} &
  A relevant task is omitted. &
   &
  1 &
  0.72\% \\ \cline{2-6} 

   &
  \textbf{Total errors} &
   &
   &
  14 &
  10.07\% \\ \cline{1-6} 
 
  \MRText[2cm]{4}{\textbf{Domain}} &
  \textbf{Correct} &
   &
   &
  115 &
  90.55\% \\ \cline{2-6} 

   &
  \textbf{Undefined domain} &
  The model fails to identify a domain even though it can be inferred from the prompt text. &
   &
  8 &
  6.30\% \\ \cline{2-6} 

   &
  \textbf{Hallucinated domain} &
  The model assigns a domain that cannot be inferred from the prompt text, or adds an irrelevant domain to an otherwise correct list. &
   &
  4 &
  3.15\% \\ \cline{2-6} 
 
   &
  \textbf{Total errors} &
   &
   &
  12 &
  9.45\% \\ \hline
\end{tabular}
\end{adjustbox}
\end{adjustbox}
\captionof{table}{Error Analysis Summary: Task\&Domain}
\label{tab:errors-task-domain}

\end{table*}

\begin{table*}[p]   
\centering

\begin{adjustbox}{width=\textwidth}
\begin{tabular}{|B|C|D|E|N|N|}
\hline
  \textbf{Field} &
  \textbf{Error Type} &
  \textbf{Description} &
  \textbf{Example} &
  \textbf{Count} &
  \% \\ \hline
  \MRText[2cm]{4}{\textbf{Central vs. Meta Instructions}} &
  \textbf{Correct} &
   &
   &
  702 &
  96.43\% \\ \cline{2-6}
   &
  \textbf{Meta misclassified as central} &
  A meta-instruction labeled as a central instruction. &
   &
  13 &
  1.79\% \\ \cline{2-6}
   &
  \textbf{Central misclassified as meta} &
  A central instruction labeled as a meta-instruction. &
  Typical of central tasks incorporated into role assignment, e.g. "You are a search assistant designed to help users by summarizing web pages.", &
  13 &
  1.79\% \\ \cline{2-6} 
   &
  \textbf{Total errors} &
   &
   &
  26 &
  3.57\% \\ \cline{1-6} 
  \MRText[2cm]{4}{\textbf{Negative Instructions}} &
  \textbf{Correct} &
   &
   &
  718 &
  98.63\% \\ \cline{2-6}
   &
  \textbf{Restriction misclassified as negative} &
  Restrictions or constraints without explicit negation labeled as negative instructions. &
  E.g., “Your answer must be within the scope of the information provided” or “Restrict the questions to the context information provided” are labeled as negative instructions. &
  6 &
  0.82\% \\ \cline{2-6}
   &
  \textbf{Ignoring negation} &
  The model fails to recognize a negation explicitly stated in the instruction. &
   &
  4 &
  0.55\% \\ \cline{2-6}
   &
  \textbf{Total errors} &
   &
   &
  10 &
  1.37\% \\ \cline{1-6}
  \MRText[2cm]{10}{\textbf{Instruction Semantic Kinds}} &
  \textbf{Correct} &
   &
   &
  654 &
  89.84\% \\ \cline{2-6} 
   &
  \textbf{Too complex instruction blocks} &
  Complex instructions containing multiple elements of distinct types . &
  E.g., “Write a summary of approximately 200 words, that gives key insights for articles: \{url\_list\}” identified as one instruction, while it includes a central task, a length restriction, an output content requirement and an input context placeholder. &
  16 &
  2.20\% \\ \cline{2-6}
   &
  \textbf{Output format vs. content} &
  Output content requirements are mislabeled as output format requirements and vice versa. &
  E.g., “The answer has to contain ONLY the translation itself” labeled as a format requirement. &
  13 &
  1.79\% \\ \cline{2-6} 
   &
  \textbf{Mislabeled output prefix} &
  Output prefixes at the end of a prompt are mislabeled as central task or output format requirement. &
  E.g. such phrases as "Answer:", "Output:", "Assistant:" at the end of the prompt. &
  7 &
  0.96\% \\ \cline{2-6} 
   &
  \textbf{Mislabeled evaluation criteria} &
  Evaluation criteria labeled as another type. &
  E.g. "A higher Shelf Life Score indicates that the product is selling faster" labeled as example clarification (even though the prompt contains no examples). &
  7 &
  0.96\% \\ \cline{2-6} 
   &
  \textbf{Conditional instruction misclassified as restriction} &
  Conditional instructions labeled as restrictions. &
  Typical of instructions to admit unanswerability, e.g. “If you don’t know the answer, say None” &
  4 &
  0.55\% \\ \cline{2-6} 
   &
  \textbf{Task/question description mislabled as role specification} &
  This is typical of cases where task/question description is expressed as an assertion in the second person (rather than instruction or question). &
  E.g. "You are extracting data from a public financial document" &
  3 &
  0.41\% \\ \cline{2-6} 
   &
  \textbf{Mislabeled role specification} &
  Role specification islabeled as another type &
  E.g "Your name is \{ai\_name\}" labeled as otput content requirement. &
  3 &
  0.41\% \\ \cline{2-6} 
   &
  \textbf{Other} &
   &
   &
  21 &
  2.88\% \\ \cline{2-6}
   &
  \textbf{Total errors} &
   &
   &
  74 &
  10.16\% \\ \hline
\end{tabular}
\end{adjustbox}%
\caption{Error Analysis Summary: Instruction Sequences}
\label{tab:errors-instructions}
\end{table*}

\begin{table*}[p]   
\centering

\begin{adjustbox}{width=\textwidth}
\begin{tabular}{|B|C|D|E|N|N|}
\hline
  \textbf{Field} &
  \textbf{Error Type} &
  \textbf{Description} &
  \textbf{Example} &
  \textbf{Count} &
  \% \\ \hline
   &
  \textbf{Correct} &
   &
   &
  208 &
  91.63\% \\ \cline{2-6} 
   &
  \textbf{Missing context} &
  A context unit not detected by the model. &
   &
  8 &
  3.52\% \\ \cline{2-6} 
   &
  \textbf{Output format mislabeled as context} &
  Output format demonstrations or requirements labeled as context. &
   &
  7 &
  3.08\% \\ \cline{2-6} 
   &
  \textbf{Question mislabeled as context} &
  Input question labeled as context. &
   &
  1 &
  0.44\% \\ \cline{2-6} 
   &
  \textbf{Other} &
   &
   &
  3 &
  1.32\% \\ \cline{2-6}
  \MRText{-6}{\textbf{Context Evidence}} &
  \textbf{Total errors} &
   &
   &
  19 &
  8.37\% \\ \cline{1-6} 
   &
  \textbf{Correct} &
   &
   &
  206 &
  90.75\% \\ \cline{2-6} 
   &
  \textbf{Hallucinated type} &
  The model assigns a type that cannot be determined from the prompt text. &
   &
  8 &
  3.52\% \\ \cline{2-6} 
   &
  \textbf{Mislabeled type (inferable from prompt)} &
  An undefined or incorrect type where the correct type can be inferred from the prompt. &
  E.g. the type is mentioned elsewhere in the prompt or implied by input examples. &
  7 &
  3.08\% \\ \cline{2-6}  
   &
  \textbf{Mislabeled type (inferable from parametric knowledge)} &
  An undefined or incorrect type where the correct type can be inferred from general knowledge. &
  E.g. the model fails to classify the type of a Python function signature as "code". &
  4 &
  1.76\% \\ \cline{2-6}  
   &
  \textbf{Short text instead of text} &
  The model predicts “short text” where the prompt does not specify context length. &
   &
  2 &
  0.88\% \\ \cline{2-6} 
  \MRText{-6}{\textbf{Context Type}} &
  \textbf{Total errors} &
   &
   &
  21 &
  9.25\% \\ \cline{1-6} 
   &
  \textbf{Correct} &
   &
   &
  186 &
  81.94\% \\ \cline{2-6}  
   &
  \textbf{Hallucinated structure} &
  The model assigns a structure where it cannot be determined from the prompt. &
   &
  24 &
  10.57\% \\ \cline{2-6}  
   &
  \textbf{Mislabeled structure (inferable from prompt)} &
  An undefined or incorrect structure where the correct structure can be inferred from the prompt. &
  E.g., the structure is mentioned elsewhere in the prompt or demonstrated in input examples. &
  8 &
  3.52\% \\ \cline{2-6} 
   &
  \textbf{Mislabeled structure (inferable from parametric knowledge)} &
  An undefined or incorrect structure even though the correct structure can be inferred from general knowledge. &
  E.g., the structure label of a chat history should "list" because it contains multiple items- dialogue turns. &
  5 &
  2.20\% \\ \cline{2-6}  
   &
  \textbf{Output structure instead of context structure} &
  The model reports the output structure instead of the context structure. &
  E.g., “dictionary” where dictionary is the expected output structure. &
  2 &
  0.88\% \\ \cline{2-6} 
   &
  \textbf{Other} &
   &
   &
  2 &
  0.88\% \\ \cline{2-6} 
  \MRText{-7}{\textbf{Context Structure}} &
  \textbf{Total errors} &
   &
   &
  41 &
  18.06\% \\ \cline{1-6}
   &
  \textbf{Correct} &
   &
   &
  221 &
  97.36\% \\ \cline{2-6}
   &
  \textbf{Hallucinated language} &
  The model assigns a language where none applies (e.g., numeric input labeled as a specific language). &
   &
  2 &
  0.88\% \\ \cline{2-6} 
   &
  \textbf{Undefined language (inferable from examples)} &
  The model fails to deternine a language where it can be inferred from input examples. &
   &
  2 &
  0.88\% \\ \cline{2-6}  
   &
  \textbf{Undefined language (inferable from common sense)} &
  Undefined language though it can be inferred from general knowledge and common sense. &
  E.g., a placeholder value in a Japanese prompt is lmost ikely also in Japanese. &
  2 &
  0.88\% \\ \cline{2-6} 
  \MRText{-5}{\textbf{Context Language}} &
  \textbf{Total errors} &
   &
   &
  6 &
  2.64\% \\ \cline{1-6} 
   &
  \textbf{Correct} &
   &
   &
  221 &
  97.36\% \\ \cline{2-6} 
   &
  \textbf{Hallucinated modality (misreading the prompt)} &
  Wrong modality label based on an incorrect interpretation of the prompt. &
  E.g.., “context about a video” labeled as video modality. &
  3 &
  1.32\% \\ \cline{2-6}  
   &
  \textbf{Modality undefined (inferable from common sense)} &
  Undefined modality where it can be inferred from general knowledge. &
  E.g., a placeholder inside a formatted string is clearly text. &
  1 &
  0.44\% \\ \cline{2-6} 
   &
  \textbf{Modality undefined (explicitly stated in prompt)} &
  The model returns “undefined” where the modality is directly stated in the prompt. &
  E.g., instructions such as “Analyze the text…” clearly indicate textual context &
  1 &
  0.44\% \\ \cline{2-6}
   &
  \textbf{Hallucinated modality} &
  The model assigns a modality that cannot be determined from the prompt text. &
  \cellcolor[HTML]{FFFFFF} &
  1 &
  0.44\% \\ \cline{2-6} 
  \MRText{-6}{\textbf{Context Modality}} &
  \textbf{Total errors} &
   &
   &
  6 &
  2.64\% \\ \cline{1-6} 
   &
  \textbf{Correct} &
   &
   &
  95 &
  95.00\% \\ \cline{2-6} 
   &
  \textbf{Hallucinated variability} &
  The model assigns a variability label where it cannot be determined from the prompt. &
   &
  3 &
  3.00\% \\ \cline{2-6} 
   &
  \textbf{Incorrect variability due to misidentified context} &
  Wrong variability type as a result of incorrectly identified context. &
  E. g., predicting “none” (meaning that context is missing) when context is actually present but was not identified. &
  1 &
  1.00\% \\ \cline{2-6} 
   &
  \textbf{Incorrect variability due to ignored placeholder in the context} &
  Predicting "fixed" where context includes a placeholder (i.e. the correct label is "varying"). &
   &
  1 &
  1.00\% \\ \cline{2-6} 
  \MRText{-5}{\textbf{Context Variability}} &
  \textbf{Total errors} &
   &
   &
  5 &
  5.00\% \\ \hline
\end{tabular}
\end{adjustbox}%
\caption{Error Analysis Summary: Input Context}
\label{tab:errors-context}
\end{table*}

 \begin{table*}[p]  
\centering

\begin{adjustbox}{width=\textwidth}
\begin{tabular}{|B|C|D|E|N|N|}
\hline
  \textbf{Field} &
  \textbf{Error Type} &
  \textbf{Description} &
  \textbf{Example} &
  \textbf{Count} &
  \% \\ \hline
\MRText[2cm]{5}{\textbf{Directions Text}} &
  \textbf{Correct} &
   &
   &
  100 &
  69.44\%
   \\ \cline{2-6} 
 &
  \textbf{Meta-instructions labeled as directions} &
  Some of the meta-instructions are incorrectly labeled as directions. &
  E.g., instructions to admit unanswerability, reasoning instructions etc. &
  41 &
  28.47\%
   \\ \cline{2-6} 
 &
  \textbf{Input question mislabeled as directions} &
  The input question and directions sometimes overlap. This only counts as an error when clear directions, distinct from the question, are also present. &
   &
  2 &
  1.39\%
   \\ \cline{2-6} 
 &
  \textbf{Missing directions} &
  Directions clearly present in the text are not detected by the model. &
   &
  1 &
  0.69\%
   \\ \cline{2-6} 
 &
  \textbf{Total errors} &
   &
   &
  44 &
  30.56\%
   \\ \hline
\MRText[2cm]{2}{\textbf{Directions Language}} &
  \textbf{Correct} &
   &
   &
  144 &
  100.00\%
   \\ \cline{2-6} 
 &
  \textbf{Total errors} &
   &
   &
  0 &
  0.00\%
   \\ \hline
\MRText[2cm]{6}{\textbf{Question Evidence}} &
  \textbf{Correct} &
   &
   &
  137 &
  91.95\%
   \\ \cline{2-6} 
 &
  \textbf{Missing question unit} &
  A question unit was not detected by the model. &
   &
  6 &
  4.03\%
   \\ \cline{2-6} 
 &
  \textbf{Directions mislabeled as question} &
  Directions labeled as question units. Only counts as an error when distinct from the question. &
   &
  2 &
  1.34\%
   \\ \cline{2-6} 
 &
  \textbf{Role instructions mislabeled as question} &
  Role instructions labeled as question units. Only counts as an error when they are distinct. &
   &
  1 &
  0.67\%
   \\ \cline{2-6} 
 &
  \textbf{Other} &
   &
   &
  3 &
  2.01\%
   \\ \cline{2-6} 
 &
  \textbf{Total errors} &
   &
   &
  12 &
  8.05\%
   \\ \hline
\MRText[2cm]{3}{\textbf{Question Type}} &
  \textbf{Correct} &
   &
   &
  142 &
  95.30\%
   \\ \cline{2-6} 
 &
  \textbf{Hallucinated type} &
  The model assigns a type when it cannot be determined based on the prompt text &
   &
  7 &
  4.70\%
   \\ \cline{2-6} 
 &
  \textbf{Total errors} &
   &
   &
  7 &
  4.70\%
   \\ \hline
\MRText[2cm]{5}{\textbf{Question Structure}} &
  \textbf{Correct} &
   &
   &
  138 &
  92.62\%
   \\ \cline{2-6} 
 &
  \textbf{Hallucinated structure} &
  The model assigns a structure where it cannot be determined from the prompt. &
   &
  6 &
  4.03\%
   \\ \cline{2-6} 
 &
  \textbf{Output structure mislabeled as question structure} &
  The model reports the output structure instead of the question structure. &
  E.g., “dictionary” where dictionary is the expected output structure. &
  4 &
  2.68\%
   \\ \cline{2-6} 
 &
  \textbf{Undefined or wrong structure (inferable from prompt)} &
  The model predicts an undefined or incorrect structure even though the correct structure can be inferred from the prompt. &
  E.g., the structure is mentioned elsewhere in the prompt or demonstrated in input examples. &
  1 &
  0.67\%
   \\ \cline{2-6} 
 &
  \textbf{Total errors} &
   &
   &
  11 &
  7.38\%
   \\ \hline
\MRText[2cm]{3}{\textbf{Question Language}} &
  \textbf{Correct} &
   &
   &
  147 &
  98.66\%
   \\ \cline{2-6} 
 &
  \textbf{Undefined language (inferable from common sense)} &
  The model fails to identify a language where it can be inferred from general knowledge or common sense. &
   &
  2 &
  1.34\%
   \\ \cline{2-6} 
 &
  \textbf{Total errors} &
   &
   &
  2 &
  1.34\%
   \\ \hline
\MRText[2cm]{4}{\textbf{Question Variability}} &
  \textbf{Correct} &
   &
   &
  94 &
  94.00\%
   \\ \cline{2-6} 
 &
  \textbf{Incorrect variability due to misidentified question} &
  Errors caused by undetected or mislabeled question units. &
  E.g. the model failed to identify a question unit with a placeholder; as a result "fixed" was predicted instead of "varying". &
  4 &
  4.00\%
   \\ \cline{2-6} 
 &
  \textbf{Other} &
   &
   &
  2 &
  2.00\%
   \\ \cline{2-6} 
 &
  \textbf{Total errors} &
   &
   &
  6 &
  6.00\%
   \\ \hline
\end{tabular}%
\end{adjustbox}      
\caption{Error Analysis Summary: Input Directions\&Question}
\label{tab:errors-question}
\end{table*}

\begin{table*}[p]
\centering

\begin{adjustbox}{width=\textwidth}
\begin{adjustbox}{width=\textwidth}
\begin{tabular}{|B|C|D|E|N|N|}
\hline
  \textbf{Field} &
  \textbf{Error Type} &
  \textbf{Description} &
  \textbf{Example} &
  \textbf{Count} &
  \% \\ \hline
\MRText[2cm]{9}{\textbf{Output Type}} &
  \textbf{Correct} &
   &
   &
  67 &
  60.36\% \\ \cline{2-6} 
 &
  \textbf{Short text instead of text} &
  The model returns “short text” when the prompt does not specify length. &
   &
  17 &
  15.32\% \\ \cline{2-6} 
 &
  \textbf{Hallucinated type} &
  The model returns an output type that cannot be inferred from the prompt. &
   &
  10 &
  9.01\% \\ \cline{2-6} 
 &
  \textbf{Erroneous complex types} &
  Multiple types are grouped while should be split. &  E.g., “complex: code, explanation” though these are distinct output units. 
   &
  7 &
  6.31\% \\ \cline{2-6} 
 &
  \textbf{Ignored types} &
  The model omits one or more output types present in the prompt. &
  E.g., the model returns only “text” while output includes text and JSON. &
  4 &
  3.60\% \\ \cline{2-6} 
 &
  \textbf{Too specific} &
  The model hallucinates a more specific type than indicated in the prompt &
  E.g. the model returns “article” when the prompt only suggests that the output is text, without specifying the type. &
  1 &
  0.90\% \\ \cline{2-6} 
 &
  \textbf{Type mismatch} &
  The prompt explicitly specifies a type, but the model predicts a different one. &
  E.g., prompt says “sentence”, model returns “paragraph”. &
  1 &
  0.90\% \\ \cline{2-6} 
 &
  \textbf{Other} &
   &
   &
  4 &
  3.60\% \\ \cline{2-6} 
 &
  \textbf{Total errors} &
   &
   &
  44 &
  39.64\% \\ \hline
\MRText[2cm]{3}{\textbf{Output Structure}} &
  \textbf{Correct} &
   &
   &
  106 &
  95.50\% \\ \cline{2-6} 
 &
  \textbf{Hallucinated structure} &
  The model hallucinates a structure where it cannot be determined. &
   &
  5 &
  4.50\% \\ \cline{2-6} 
 &
  \textbf{Total errors} &
   &
   &
  5 &
  4.50\% \\ \hline
\MRText[2cm]{3}{\textbf{Output Language}} &
  \textbf{Correct} &
   &
   &
  104 &
  93.69\% \\ \cline{2-6} 
 &
  \textbf{Hallucinated output language} &
  The model hallucinates a language where it cannot be determined. &
   &
  7 &
  6.31\% \\ \cline{2-6} 
 &
  \textbf{Total errors} &
   &
   &
  7 &
  6.31\% \\ \hline
\MRText[2cm]{4}{\textbf{Output Modality}} &
  \textbf{Correct} &
   &
   &
  107 &
  96.40\% \\ \cline{2-6} 
 &
  \textbf{Hallucinated modality} &
  The model hallucinates a modality that cannot be determined. &
   &
  3 &
  2.70\% \\ \cline{2-6} 
 &
  \textbf{Undefined modality} &
  The modality is "undefined" while it can be inferred from the prompt text. &
   &
  1 &
  0.90\% \\ \cline{2-6} 
 &
  \textbf{Total errors} &
   &
   &
  4 &
  3.60\% \\ \hline
\MRText[2cm]{8}{\textbf{Answer Paradigm}} &
  \textbf{Correct} &
   &
   &
  89 &
  80.18\% \\ \cline{2-6} 
 &
  \textbf{Free generation instead of language or style transfer} &
  The model returns “free generation” when the prompt explicitly requests translation or style transfer. &
   &
  7 &
  6.31\% \\ \cline{2-6} 
 &
  \textbf{Free generation instead of summary/paraphrase} &
  The model returns “free generation” when the prompt explicitly requests summarization or paraphrasing. &
   &
  3 &
  2.70\% \\ \cline{2-6} 
 &
  \textbf{Binary answer instead of free generation} &
  The model outputs a binary answer with a “don’t know” option when the task requires free generation. &
  This error is typical of prompts with instructions to admit ananswerability, e.g. “If you don’t know, say None” &
  3 &
  2.70\% \\ \cline{2-6} 
 &
  \textbf{Undefined answer paradigm} &
  The the answer paradigm is “undefined” when it is inferrable from the prompt. &
   &
  3 &
  2.70\% \\ \cline{2-6} 
 &
  \textbf{Hallucinated answer paradigm} &
  The model invents an answer paradigm where it cannot be determined. &
   &
  1 &
  0.90\% \\ \cline{2-6} 
 &
  \textbf{Other} &
   &
   &
  5 &
  4.50\% \\ \cline{2-6} 
 &
  \textbf{Total errors} &
   &
   &
  22 &
  19.82\% \\ \hline
\end{tabular}%

\end{adjustbox}
\end{adjustbox}
\caption{Error Analysis Summary: Output}
\label{tab:errors-output}

\vspace{0.2cm}

\begin{adjustbox}{width=\textwidth}
\begin{adjustbox}{width=\textwidth}
\begin{tabular}{|B|C|D|E|N|N|}
\hline
  \textbf{Field} &
  \textbf{Error Type} &
  \textbf{Description} &
  \textbf{Example} &
  \textbf{Count} &
  \% \\ \hline
\MRText[2cm]{10}{\textbf{Prompting Techniques}} &
  \textbf{Correct} &
   &
   &
  1182 &
  98.50\% \\ \cline{2-6} 
 &
  \textbf{Ignored role assignment} &
  The model ignores explicit role assignment. &
   &
  4 &
  0.33\% \\ \cline{2-6} 
 &
  \textbf{Ignored output demonstrations} &
  The model ignores examples or demonstrations. &
   &
  4 &
  0.33\% \\ \cline{2-6} 
 &
  \textbf{Ignored decomposition} &
  The model ignores decomposition instructions in the prompt. &
   &
  3 &
  0.25\% \\ \cline{2-6} 
 &
  \textbf{Ignored sections} &
  The model overlooks explicit segmentation into sections. &
   &
  3 &
  0.25\% \\ \cline{2-6} 
 &
  \textbf{Hallucinated audience specification} &
  The model invents an audience specification not present in the prompt . &
  E.g., “use simple words that a 3-year-old understands”: "a 3-year-old" is misinterpreted as the audience. &
  1 &
  0.08\% \\ \cline{2-6} 
 &
  \textbf{Hallucinated prompt chaining} &
  The model assumes a previous output was generated by another prompt, while no evidence supports this. &
   &
  1 &
  0.08\% \\ \cline{2-6} 
 &
  \textbf{Hallucinated structured output} &
  The model hallucinates structured output instructions. &
   &
  1 &
  0.08\% \\ \cline{2-6} 
 &
  \textbf{Few-shot vs. output demonstrations confusion} &
  The model confuses few-shot (input-output) examples with output demonstrations. &
   &
  1 &
  0.08\% \\ \cline{2-6} 
 &
  \textbf{Total errors} &
   &
   &
  18 &
  1.50\% \\ \hline
\end{tabular}%

\end{adjustbox}
\end{adjustbox}
\caption{Error Analysis Summary: Prompting Techniques}
\label{tab:errors-techniques}

\end{table*}

\begin{table*}[p]
\centering

\begin{adjustbox}{width=\textwidth}

\begin{adjustbox}{width=\textwidth}
\begin{tabular}{|l|l|c|c|}
\hline
\textbf{Category} &
  \textbf{Field} &
  \multicolumn{1}{l|}{\textbf{Accuracy}} &
  \multicolumn{1}{l|}{\textbf{Number of Evaluated Units}} \\ \hline
\multirow{2}{*}{\textbf{Language}}         & \textbf{Prompt Language}            & 93.04\% & 158  \\ \cline{2-4} 
                                           & \textbf{Explicit Language Mentions} & 88.11\% & 143  \\ \hline
\multirow{2}{*}{\textbf{Task\&Domain}}     & \textbf{Task}                       & 89.93\% & 139  \\ \cline{2-4} 
                                           & \textbf{Domain}                     & 90.55\% & 127  \\ \hline
\multirow{3}{*}{\textbf{Instruction Sequences}} &
  \textbf{Instruction Kinds} &
  89.84\% &
  728 \\ \cline{2-4} 
                                           & \textbf{Central vs. Meta}           & 96.43\% & 728  \\ \cline{2-4} 
                                           & \textbf{Negative Instructions}      & 98.63\% & 728  \\ \hline
\multirow{6}{*}{\textbf{Input Context}}    & \textbf{Context Evidence}           & 91.63\% & 227  \\ \cline{2-4} 
                                           & \textbf{Context Type}               & 90.75\% & 227  \\ \cline{2-4} 
                                           & \textbf{Context Structure}          & 81.94\% & 227  \\ \cline{2-4} 
                                           & \textbf{Context Language}           & 97.36\% & 227  \\ \cline{2-4} 
                                           & \textbf{Context Modality}           & 97.36\% & 227  \\ \cline{2-4} 
                                           & \textbf{Context Variability}        & 95.00\% & 100  \\ \hline
\multirow{2}{*}{\textbf{Input Directions}} & \textbf{Directions Text}            & 69.44\% & 144  \\ \cline{2-4} 
                                           & \textbf{Directions Language}        & 100\%   & 144  \\ \hline
\multirow{5}{*}{\textbf{Input Question}}   & \textbf{Question Evidence}          & 91.95\% & 149  \\ \cline{2-4} 
                                           & \textbf{Question Type}              & 95.30\% & 149  \\ \cline{2-4} 
                                           & \textbf{Question Structure}         & 92.62\% & 149  \\ \cline{2-4} 
                                           & \textbf{Question Language}          & 98.66\% & 149  \\ \cline{2-4} 
                                           & \textbf{Question Variability}       & 94.00\% & 100  \\ \hline
\multirow{5}{*}{\textbf{Output}}           & \textbf{Output Type}                & 60.36\% & 111  \\ \cline{2-4} 
                                           & \textbf{Output Structure}           & 95.50\% & 111  \\ \cline{2-4} 
                                           & \textbf{Output Language}            & 93.69\% & 111  \\ \cline{2-4} 
                                           & \textbf{Output Modality}            & 96.40\% & 111  \\ \cline{2-4} 
                                           & \textbf{Answer Paradigm}            & 80.18\% & 111  \\ \hline
\textbf{Prompting Techniques}              & \textbf{Prompting Techniques}       & 98.50\% & 1200 \\ \hline
\end{tabular}%
\end{adjustbox}

\end{adjustbox}
\captionof{table}{Annotation Accuracy Per Field (based on error analysis results)}
\label{tab:errors-accuracy}

\vspace{0.5cm}

\begin{adjustbox}{width=\textwidth}
  \includegraphics[width=\textwidth]{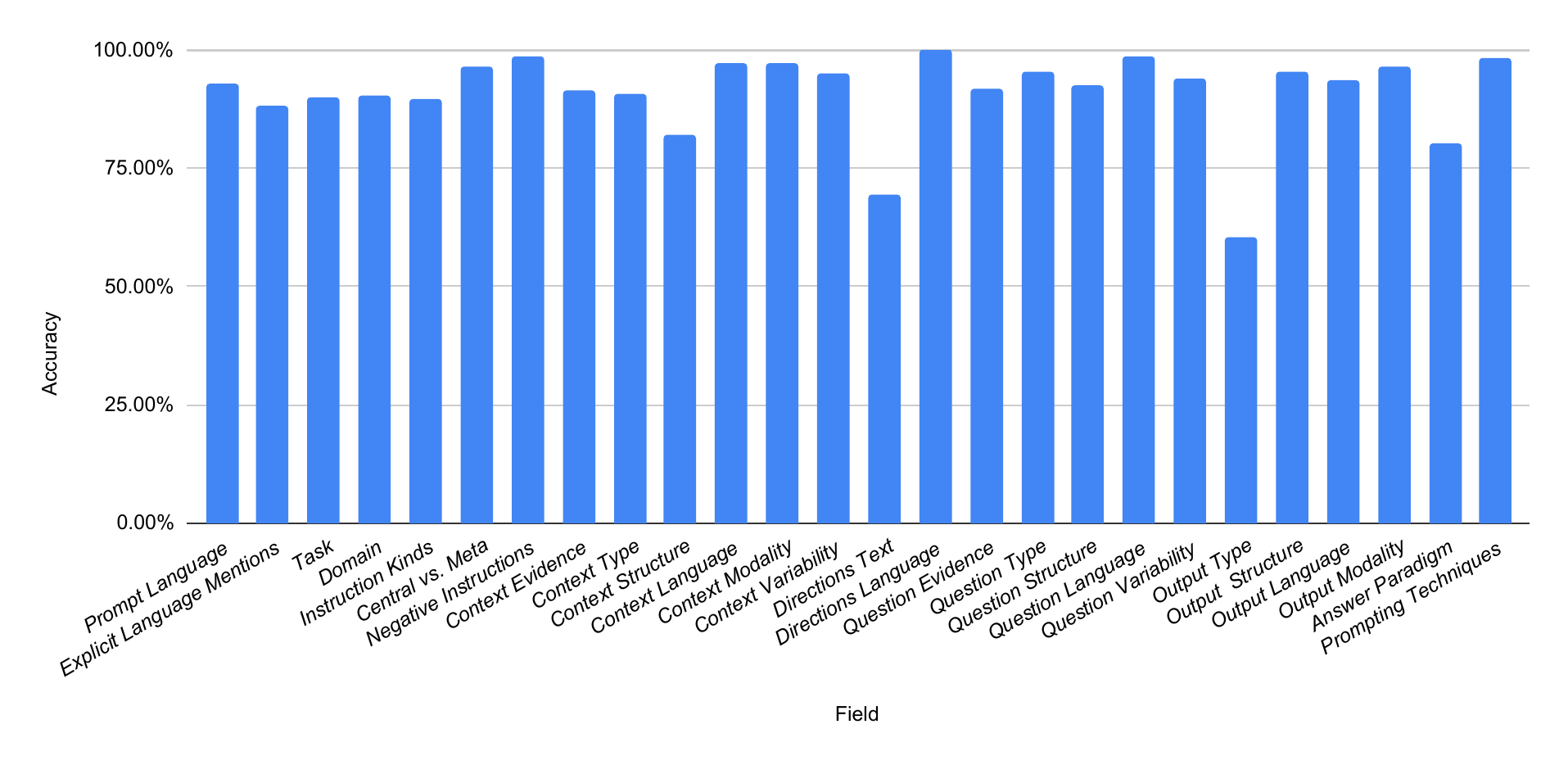}
\end{adjustbox}
\captionof{figure}{Annotation Accuracy Per Field (based on error analysis results)}
\label{fig:errors-accuracy}
\end{table*}
\FloatBarrier    
\clearpage       

\section{Analysis: Charts and Tables}
\label{sec:charts}
Tables \ref{tab:task-frequency}-\ref{fig:block-sets} and Figures \ref{fig:common-languages}-\ref{fig:prompting-technique-counts} below illustrate the results of the analysis presented in Section \ref{sec:analysis}.

\begin{table*}[!b]
\centering

\begin{adjustbox}{width=\textwidth} 

\begin{tabular}{|lrlrlr|}
\hline
\multicolumn{6}{|c|}{\textbf{Task Cases}} \\ \hline
\multicolumn{2}{|l|}{\cellcolor[HTML]{D9D9D9}\textbf{Top 10 tasks}} &
  \multicolumn{2}{l|}{\cellcolor[HTML]{D9D9D9}\textbf{Mid-frequency Tasks}} &
  \multicolumn{2}{l|}{\cellcolor[HTML]{D9D9D9}\textbf{Long-tail Tasks}} \\ \hline
\multicolumn{1}{|l|}{\textbf{Task}} &
  \multicolumn{1}{l|}{\textbf{Count}} &
  \multicolumn{1}{l|}{\textbf{Task}} &
  \multicolumn{1}{l|}{\textbf{Count}} &
  \multicolumn{1}{l|}{\textbf{Task}} &
  \multicolumn{1}{l|}{\textbf{Count}} \\ \hline
\multicolumn{1}{|l|}{question\_answering} &
  \multicolumn{1}{r|}{14176} &
  \multicolumn{1}{l|}{ranking} &
  \multicolumn{1}{r|}{428} &
  \multicolumn{1}{l|}{state\_tracking} &
  16 \\ \hline
\multicolumn{1}{|l|}{general\_text\_generation} &
  \multicolumn{1}{r|}{11359} &
  \multicolumn{1}{l|}{code\_transformation} &
  \multicolumn{1}{r|}{393} &
  \multicolumn{1}{l|}{system\_integration} &
  12 \\ \hline
\multicolumn{1}{|l|}{information\_extraction} &
  \multicolumn{1}{r|}{6764} &
  \multicolumn{1}{l|}{text\_analysis} &
  \multicolumn{1}{r|}{387} &
  \multicolumn{1}{l|}{style\_analysis} &
  11 \\ \hline
\multicolumn{1}{|l|}{summarization} &
  \multicolumn{1}{r|}{6496} &
  \multicolumn{1}{l|}{image\_processing} &
  \multicolumn{1}{r|}{253} &
  \multicolumn{1}{l|}{game\_strategy} &
  10 \\ \hline
\multicolumn{1}{|l|}{classification} &
  \multicolumn{1}{r|}{4969} &
  \multicolumn{1}{l|}{diagnosis} &
  \multicolumn{1}{r|}{228} &
  \multicolumn{1}{l|}{knowledge\_management} &
  9 \\ \hline
\multicolumn{1}{|l|}{code\_generation} &
  \multicolumn{1}{r|}{2908} &
  \multicolumn{1}{l|}{dialogue\_management} &
  \multicolumn{1}{r|}{217} &
  \multicolumn{1}{l|}{task\_formulation} &
  9 \\ \hline
\multicolumn{1}{|l|}{explanatory\_and\_instructional\_generation} &
  \multicolumn{1}{r|}{2157} &
  \multicolumn{1}{l|}{creative\_and\_narrative\_generation} &
  \multicolumn{1}{r|}{184} &
  \multicolumn{1}{l|}{data\_management} &
  6 \\ \hline
\multicolumn{1}{|l|}{planning} &
  \multicolumn{1}{r|}{2134} &
  \multicolumn{1}{l|}{data\_cleaning} &
  \multicolumn{1}{r|}{181} &
  \multicolumn{1}{l|}{natural\_language\_understanding} &
  4 \\ \hline
\multicolumn{1}{|l|}{dialogue\_and\_response\_generation} &
  \multicolumn{1}{r|}{1864} &
  \multicolumn{1}{l|}{parsing} &
  \multicolumn{1}{r|}{174} &
  \multicolumn{1}{l|}{policy\_generation} &
  3 \\ \hline
\multicolumn{1}{|l|}{recommendation} &
  \multicolumn{1}{r|}{1710} &
  \multicolumn{1}{l|}{speech\_processing} &
  \multicolumn{1}{r|}{158} &
  \multicolumn{1}{l|}{information\_fusion} &
  3 \\ \hline
\end{tabular}%

\end{adjustbox}
\captionof{table}{Task Distribution in the Dataset: Top, Mid-Frequency, and Long-Tail}
\label{tab:task-frequency}

\end{table*}

\begin{figure*}[!b]
\includegraphics[width=\textwidth]{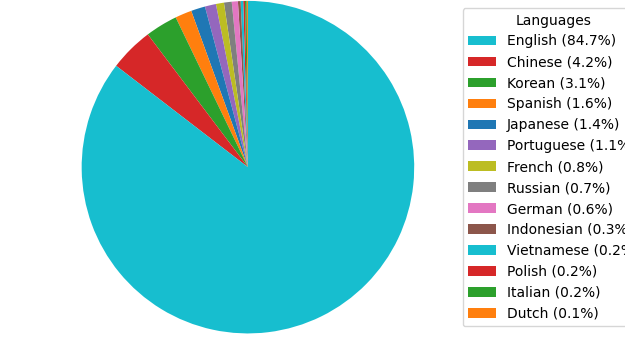}
\caption{Most frequent prompt languages in the dataset (top 14, $\geq$1\% each)}
\label{fig:common-languages}
\end{figure*}

\begin{figure*}[t!]
\includegraphics[width=\textwidth]{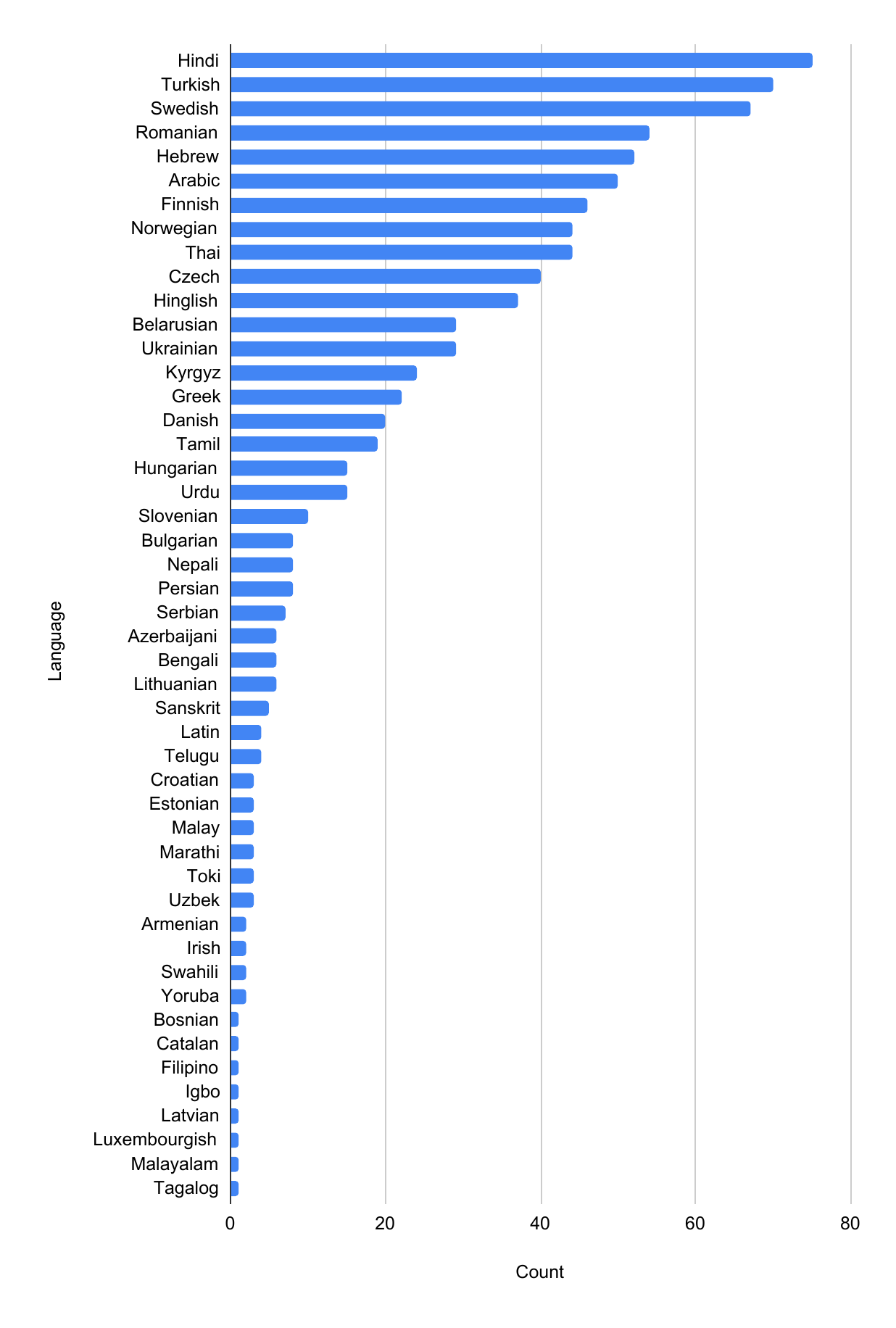}
\caption{Long-tail languages occurring below 100 times in the data}
\label{fig:long-tail-languages}
\end{figure*}

\begin{figure*}[t!]
\centering
\includegraphics[width=\textwidth]{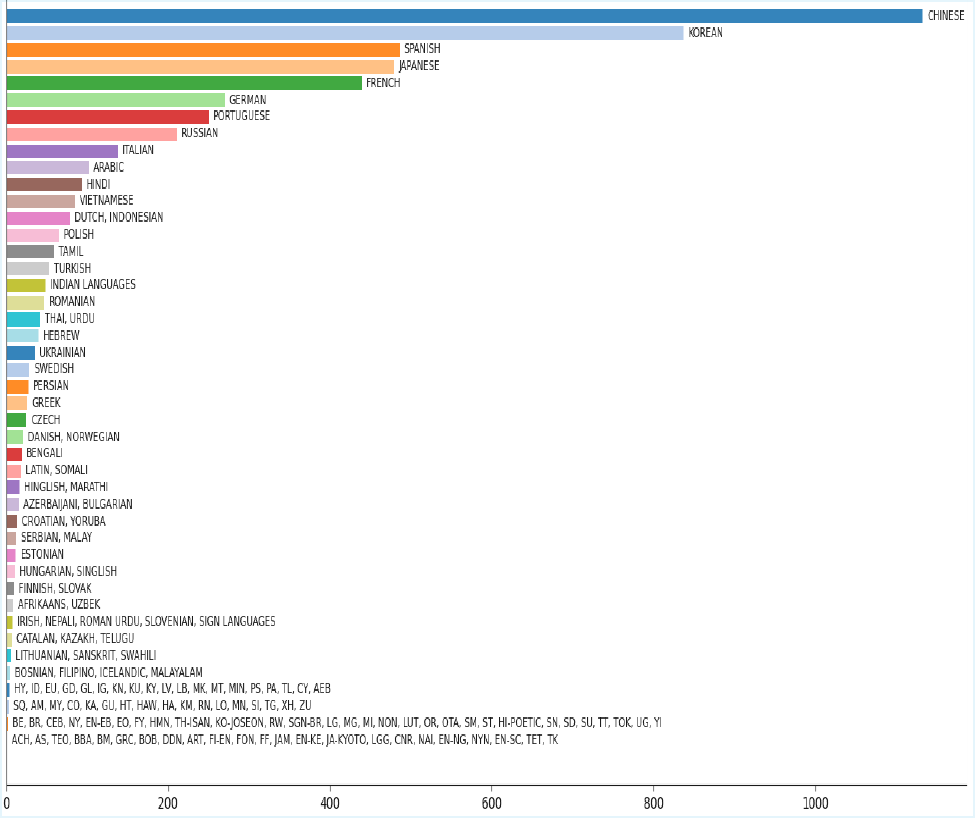}

\caption{Explicit Language Mentions (excluding English). Abbreviation key: Armenian = HY; Bahasa Indonesia = ID; Basque = EU; Galician = GL; Igbo = IG; Kannada = KN; Kurdish = KU; Kyrgyz = KY; Latvian = LV; Luxembourgish = LB; Macedonian = MK; Maltese = MT; Minang = MIN; Pashto = PS; Punjabi = PA; Tagalog = TL; Welsh = CY; Albanian = SQ; Amharic = AM; Burmese = MY; Corsican = CO; Georgian = KA; Gujarati = GU; Haitian Creole = HT; Hausa = HA; Khmer = KM; Kirundi = RN; Lao = LO; Mongolian = MN; Sinhala = SI; Tajik = TG; Xhosa = XH; Zulu = ZU; Belarusian = BE; Breton = BR; Cebuano = CEB; Chichewa = NY; Ebonics = EN-EB; Esperanto = EO; Frisian = FY; Hawaiian = HAW; Hmong = HMN; Isan = TH-ISAN; Joseon = KO-JOSEON; Kinyarwanda = RW; Libras = SGN-BR; Luganda = LG; Malagasy = MG; Maori = MI; Norse = NON; Odia = OR; Ottoman Turkish = OTA; Samoan = SM; Sesotho = ST; Shayari = HI-POETIC; Shona = SN; Sindhi = SD; Sundanese = SU; Tatar = TT; Toki Pona = TOK; Uyghur = UG; Yiddish = YI; Acholi = ACH; Assamese = AS; Ateso = TEO; Baatonum = BBA; Bambara = BM; Biblical Greek = GRC; Bobo = BOB; Dendi = DDN; Elfish = ART; Finglish = FI-EN; Fongbe = FON; Fula = FF; Gaelic = GD; Jamaican Patois = JAM; Kenyan = EN-KE; Kyoto dialect (Japanese) = JA-KYOTO; Lugbara = LGG; Luhshootseed = LUT; Montenegrin = CNR; Native American = NAI; Nigerian = EN-NG; Runyankole = NYN; Scottish = EN-SC; Tetun = TET; Tunisian Darija = AEB; Turkmen = TK}
\label{fig:mentions}
\end{figure*}

\begin{figure*}[t!]
\centering

\includegraphics[width=\textwidth]{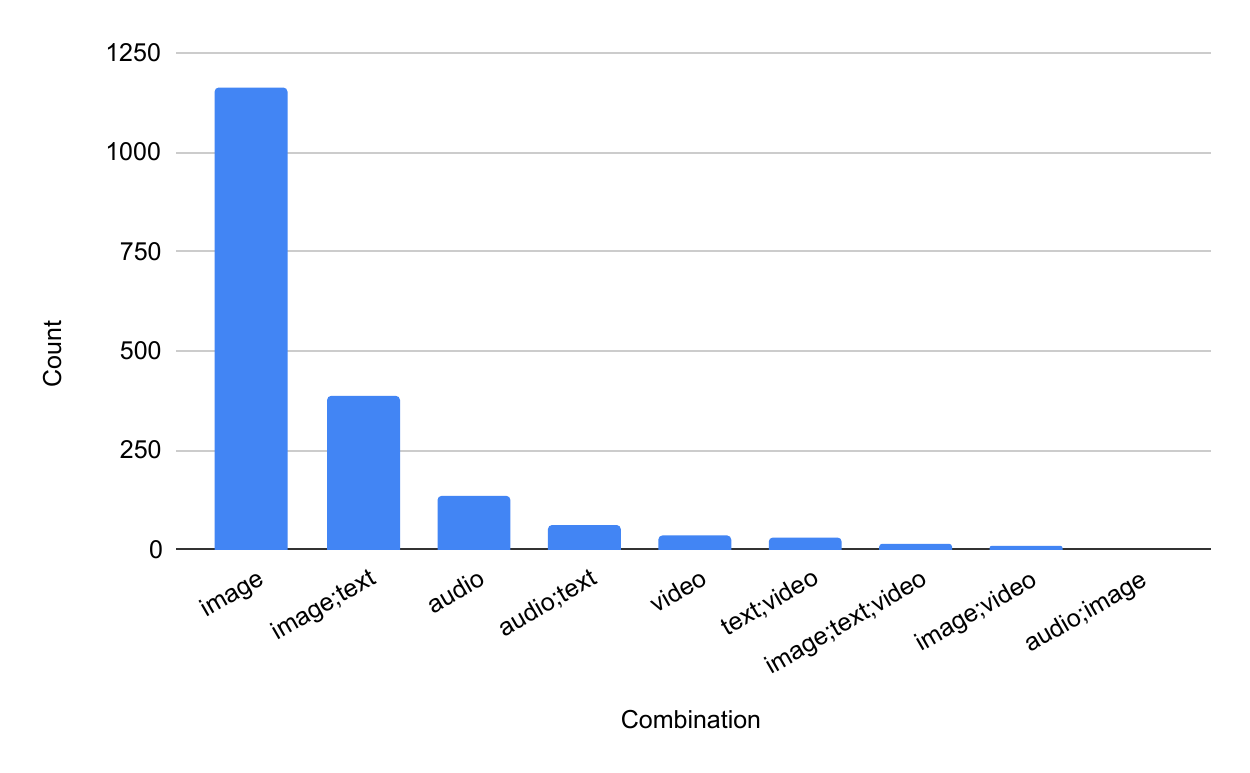}
\caption{Input non-text modality combinations}
\label{fig:input_non_text-modalities}

\vspace{0.5cm}

\includegraphics[width=\textwidth]{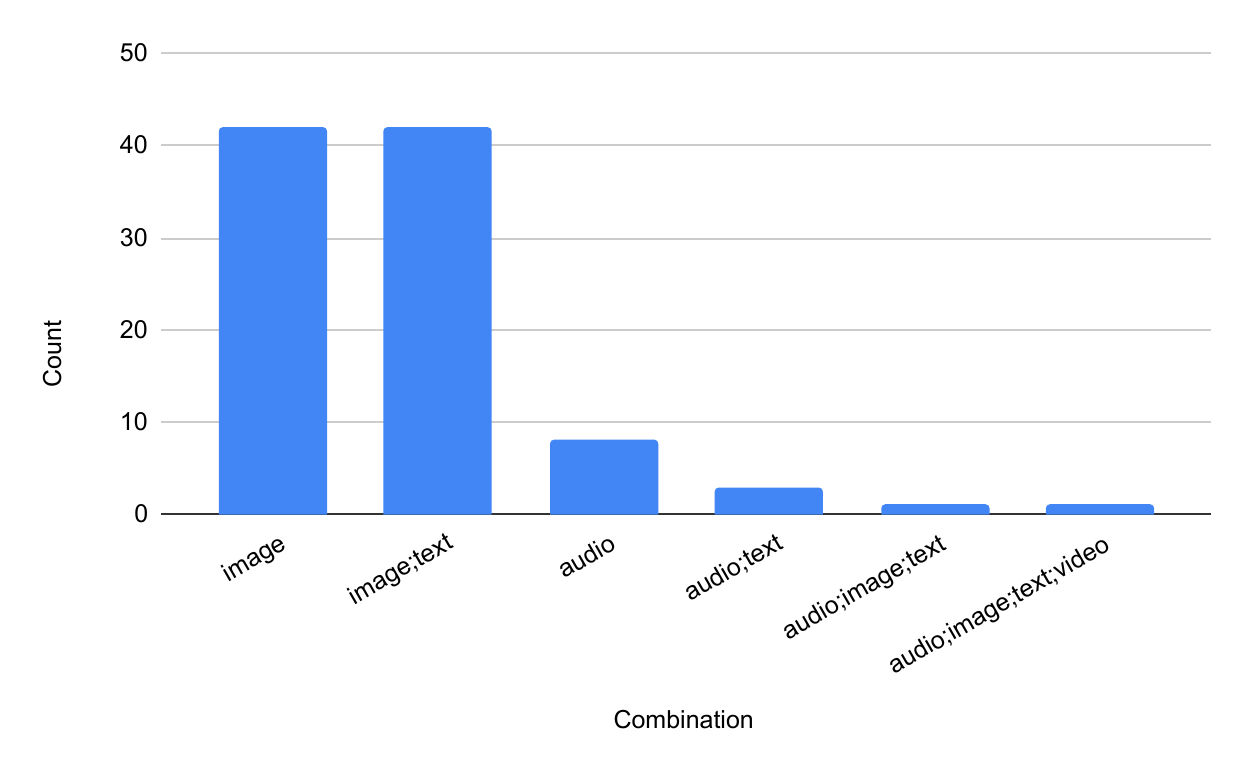}
\caption{Output non-text modality combinations}
\label{fig:output_non_text_modalities}

\end{figure*}

\begin{figure*}[t]
\includegraphics[width=\textwidth]{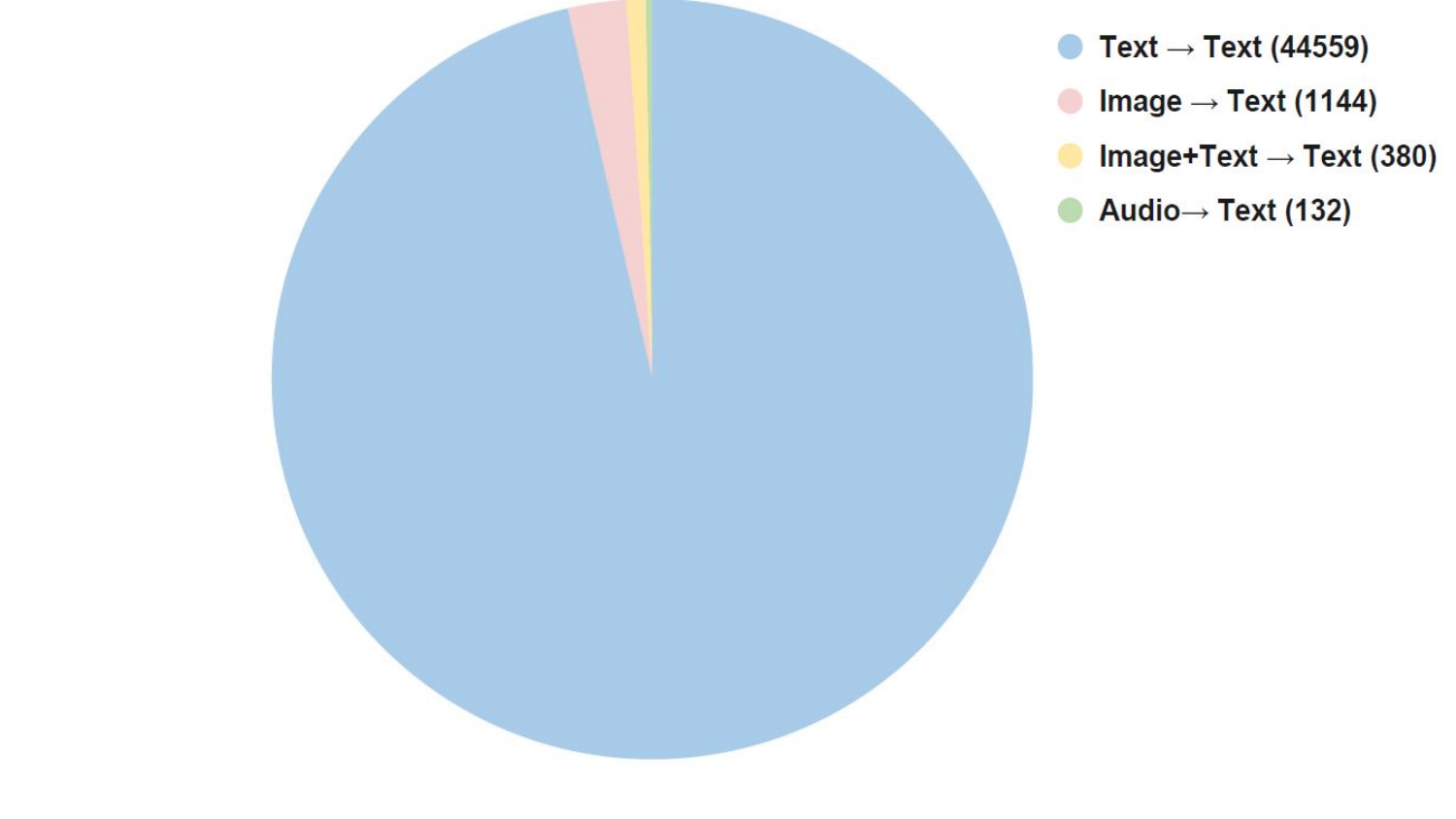}
\caption{Main input-output modality combinations}
\label{fig:main_modality_combinations}
\end{figure*}

\begin{figure*}[t!]
\includegraphics[width=\textwidth]{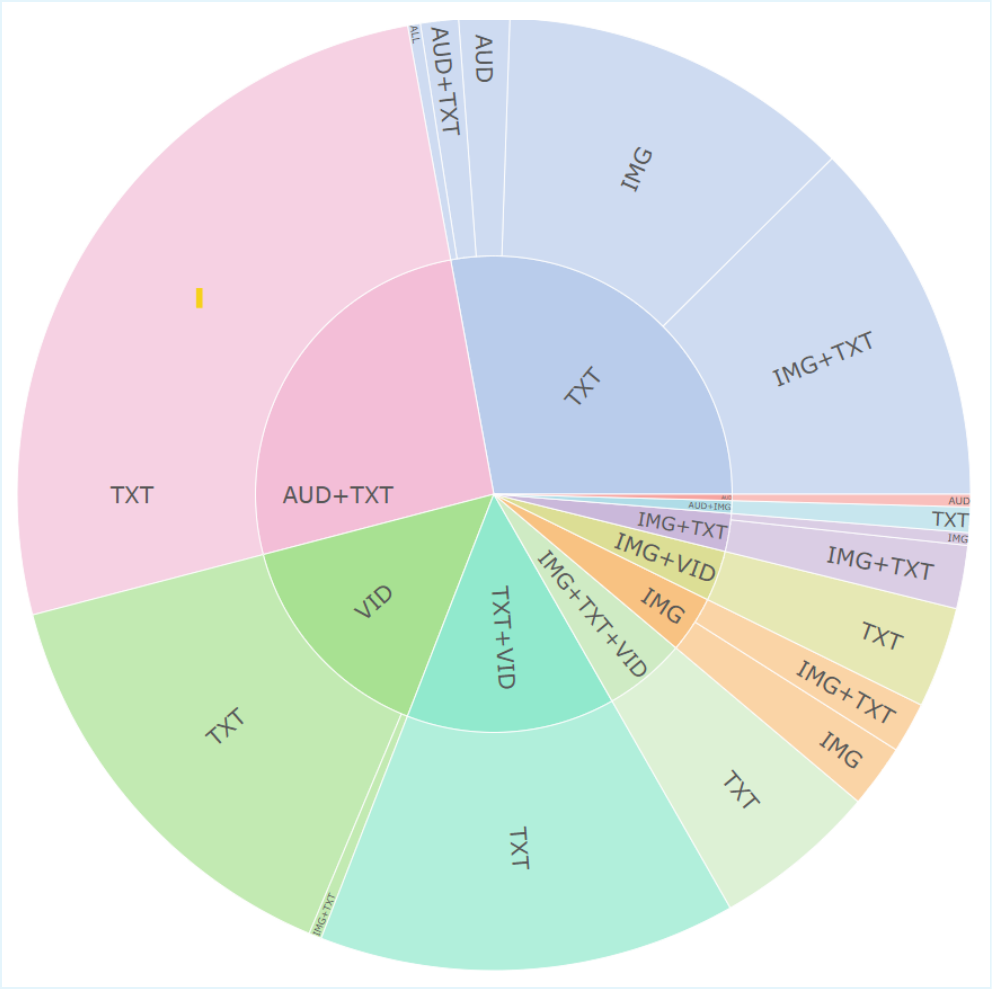}
\caption{Long-tail input–output modality combinations (less than 0.1\% each). The inner circle indicates the input; the outer circle shows the output. Text=TXT; image=IMG, audio=AUD, video=AUD, all 4 modalities=ALL.}
\label{fig:longtail_modality_combinations}
\end{figure*}

\begin{figure*}[t!]
\includegraphics[width=\textwidth]{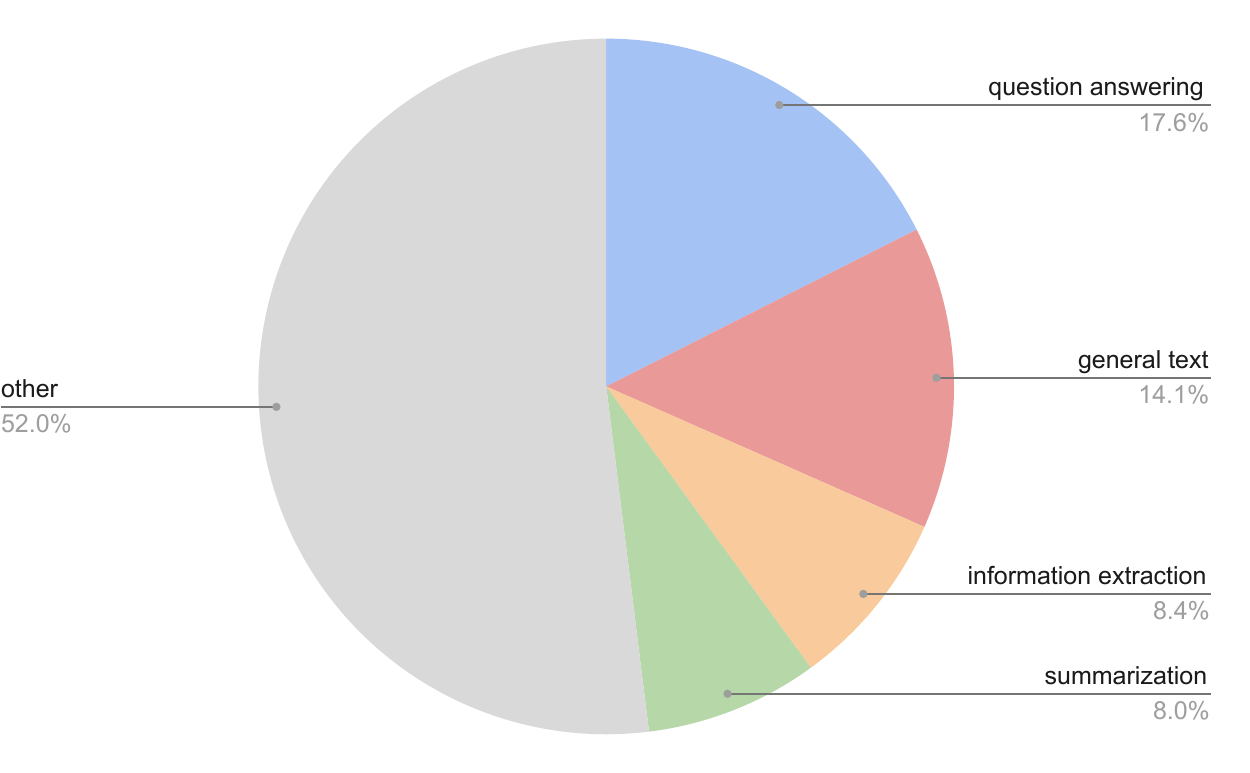}
\caption{Top four tasks covering over 48\% of the data}
\label{tab:top-4-tasks}
\end{figure*}

\begin{figure*}[t]
\centering
\includegraphics[width=\textwidth]{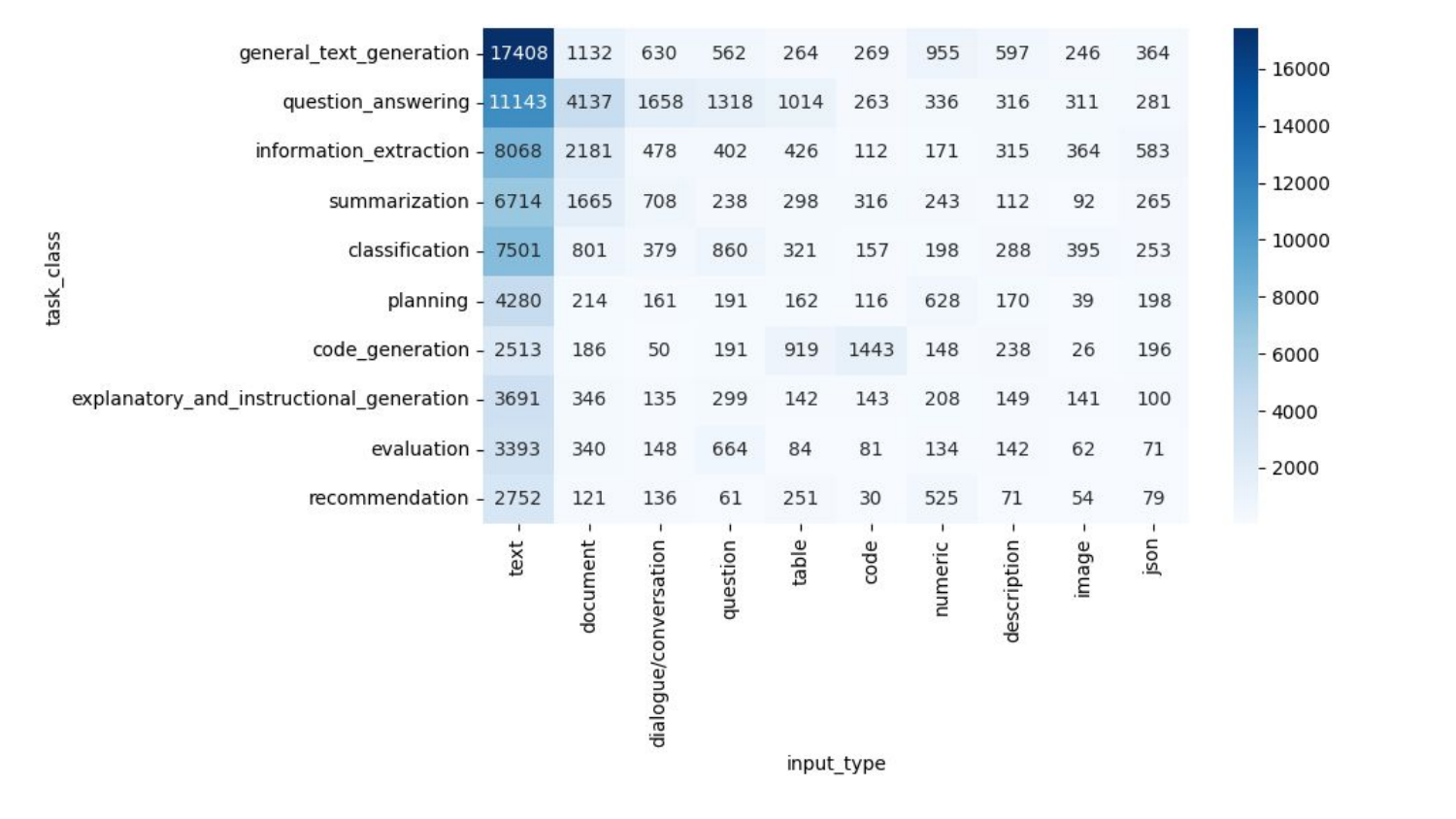}
\caption{Distribution of the top 10 input types across the top 10 tasks in the collection.}
\label{fig:task-input}
\end{figure*}

\begin{figure*}[t]

  \includegraphics[width=\linewidth]{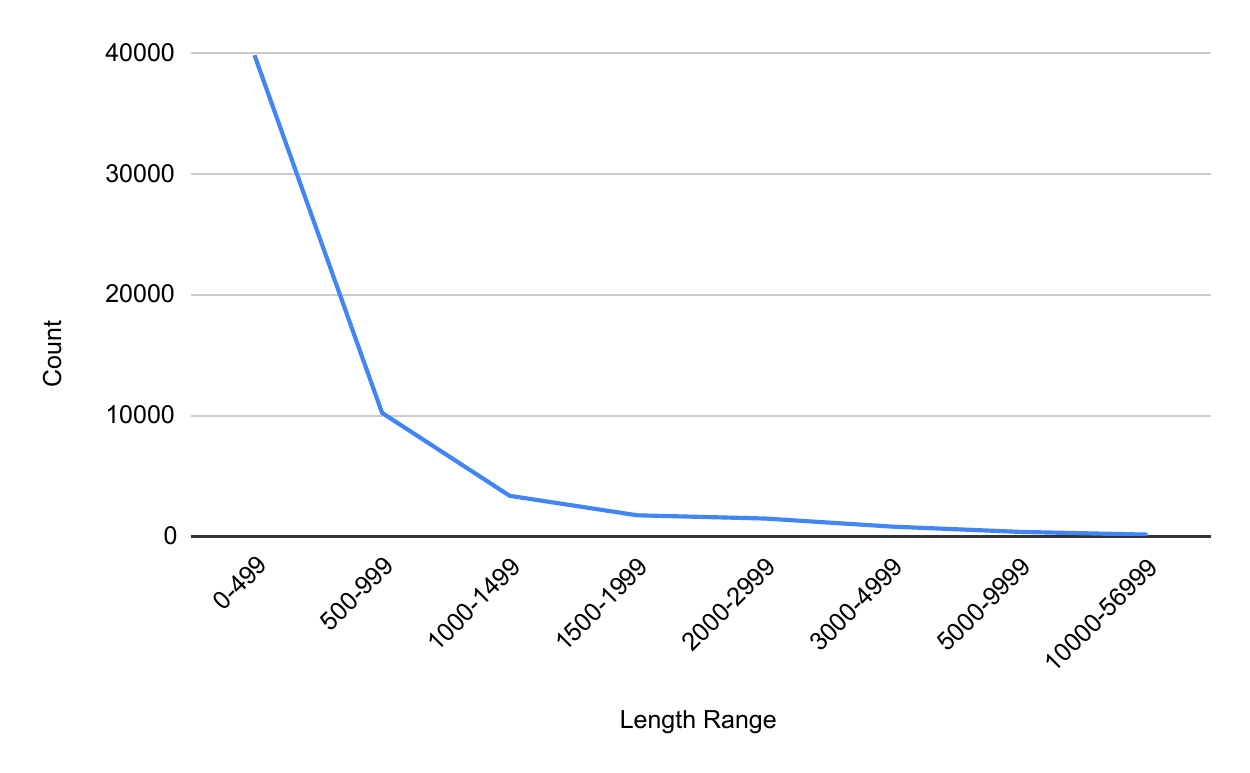}
  \caption{Distribution of prompt text lengths in the dataset}
  \label{fig:prompt-text-lengths}

  \centering
  \includegraphics[width=\linewidth]{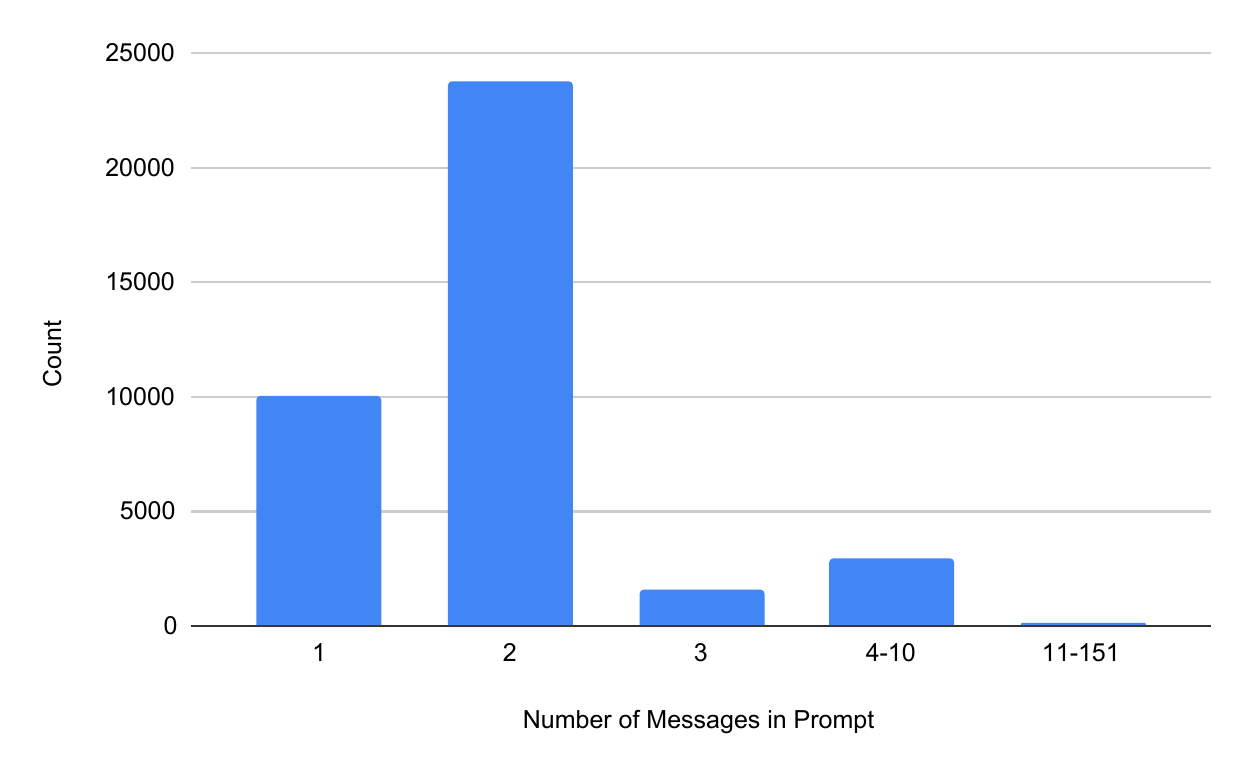}
  \caption{Number of messages per prompt (for \textit{chat.completions.create} data only).}
  \label{fig:messages-per-prompt}

\end{figure*}

\begin{figure*}[t!]
\centering
\includegraphics[width=\textwidth]{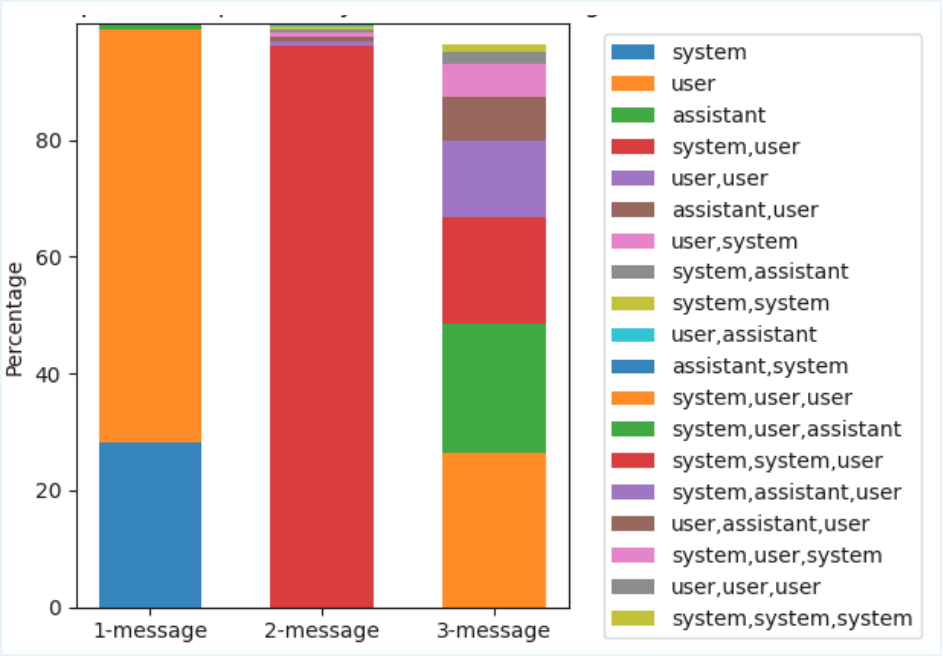}
\caption{Prompt role sequences by number of messages.}
\label{fig:role-sequences}
\end{figure*}

\begin{figure*}[p]
\centering
\includegraphics[width=\textwidth]{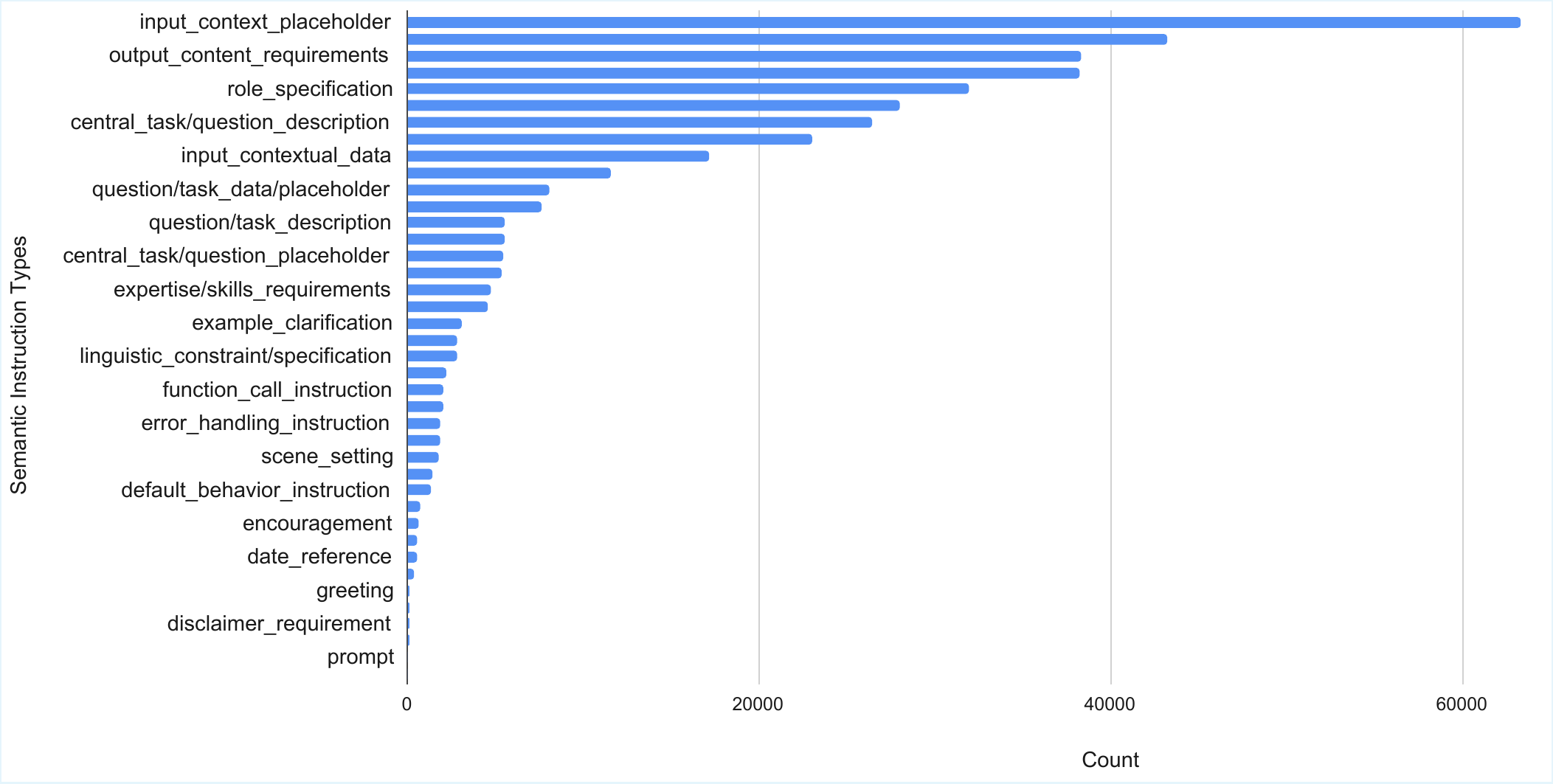}
\caption{Semantic instruction type frequencies.}
\label{fig:block-frequencies}
\end{figure*}

\begin{table*}[t]
\centering

\begin{adjustbox}{width=\textwidth}

\begin{tabular}{|p{14cm}|r|r|}
\hline
\textbf{Sequence} &
  \multicolumn{1}{l|}{\textbf{Count}} &
  \multicolumn{1}{l|}{\textbf{Num Blocks}} \\ \hline
output\_content\_requirements $\rightarrow$ constraint/restriction &
  10452 &
  2 \\ \hline
input\_context\_placeholder $\rightarrow$ output\_content\_requirements &
  9703 &
  2 \\ \hline
central\_task/question\_description $\rightarrow$ role\_specification &
  9230 &
  2 \\ \hline
input\_context\_placeholder$\rightarrow${}output\_content\_requirements$\rightarrow${}constraint/restriction &
  5038 &
  3 \\ \hline
central\_task/question\_description $\rightarrow$ output\_format\_requirement $\rightarrow$ role\_specification &
  3739 &
  3 \\ \hline
role\_specification $\rightarrow$ output\_format\_requirement $\rightarrow$ input\_context\_description &
  3433 &
  3 \\ \hline
central\_task/question\_description $\rightarrow$ role\_specification $\rightarrow$ output\_format\_requirement $\rightarrow$ input\_context\_description &
  1983 &
  4 \\ \hline
central\_task/question\_description $\rightarrow$ output\_format\_requirement $\rightarrow$ role\_specification $\rightarrow$ input\_context\_description &
  1442 &
  4 \\ \hline
input\_context\_placeholder $\rightarrow$ output\_content\_requirements $\rightarrow$ constraint/restriction $\rightarrow$ conditional\_instruction &
  1311 &
  4 \\ \hline
central\_task/question\_description $\rightarrow$ output\_format\_requirement $\rightarrow$ role\_specification $\rightarrow$ input\_context\_description $\rightarrow$ input\_context\_placeholder &
  512 &
  5 \\ \hline
central\_task/question\_description $\rightarrow$ role\_specification $\rightarrow$ output\_format\_requirement $\rightarrow$ input\_context\_description $\rightarrow$ input\_context\_placeholder &
  494 &
  5 \\ \hline
output\_format\_requirement $\rightarrow$ input\_context\_description $\rightarrow$ input\_context\_placeholder $\rightarrow$ output\_content\_requirements $\rightarrow$ constraint/restriction &
  426 &
  5 \\ \hline
\end{tabular}%

\end{adjustbox}
\captionof{table}{Most frequent sequences of semantic types for instruction block chains of length 2–5.}
\label{fig:block-sequences}

\vspace{0.5cm}

\begin{adjustbox}{width=\textwidth}

\begin{tabular}{|p{14cm}|r|r|}
\hline
\textbf{Combination}                                                             & \multicolumn{1}{l|}{\textbf{Count}} & \multicolumn{1}{l|}{\textbf{Num Blocks}} \\ \hline
input\_context\_placeholder; role\_specification                                 & 20687                               & 2                                        \\ \hline
input\_context\_placeholder; output\_format\_requirement                         & 19111                               & 2                                        \\ \hline
constraint/restriction; input\_context\_placeholder                              & 17657                               & 2                                        \\ \hline
constraint/restriction; input\_context\_placeholder; output\_format\_requirement & 10344                               & 3                                        \\ \hline
constraint/restriction; input\_context\_placeholder; role\_specification         & 10016                               & 3                                        \\ \hline
input\_context\_placeholder; output\_format\_requirement; role\_specification    & 9969                                & 3                                        \\ \hline
constraint/restriction; input\_context\_placeholder; output\_format\_requirement; role\_specification                                                & 5884 & 4 \\ \hline
central\_task/question\_description; constraint/restriction; input\_context\_placeholder; output\_format\_requirement                                & 5825 & 4 \\ \hline
constraint/restriction; input\_context\_placeholder; output\_content\_requirements; output\_format\_requirement                                      & 5591 & 4 \\ \hline
central\_task/question\_description; constraint/restriction; input\_context\_placeholder; output\_format\_requirement; role\_specification           & 3462 & 5 \\ \hline
constraint/restriction; input\_context\_placeholder; output\_content\_requirements; output\_format\_requirement; role\_specification                 & 3290 & 5 \\ \hline
central\_task/question\_description; constraint/restriction; input\_context\_placeholder; output\_content\_requirements; output\_format\_requirement & 3208 & 5 \\ \hline
\end{tabular}%

\end{adjustbox}

\captionof{table}{Most frequent sets of semantic types for instruction block chains of length 2–5 (regardless of order).}
\label{fig:block-sets}

\end{table*}

\begin{figure*}[!t]
\centering
\includegraphics[width=\textwidth]{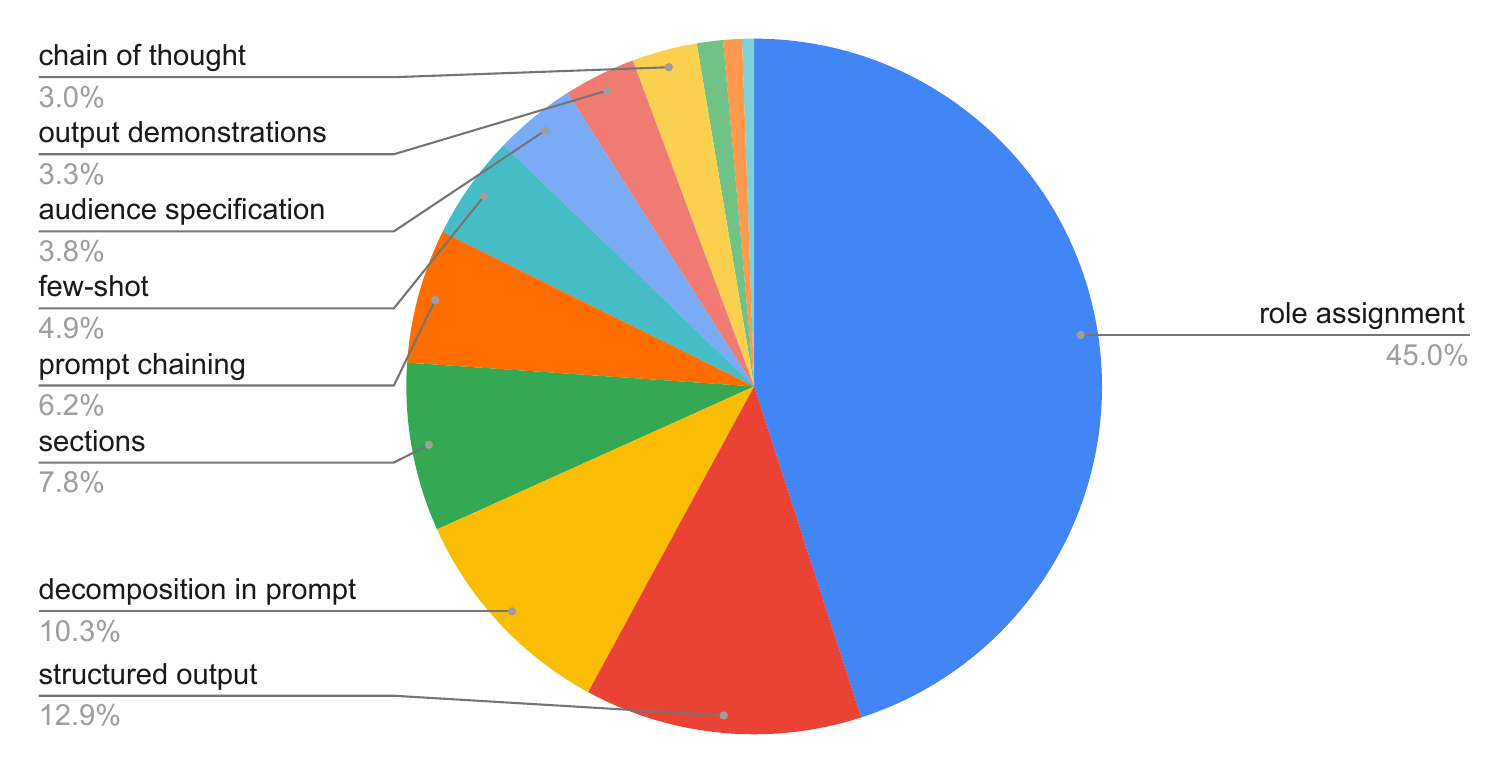}
\caption{Distribution of prompting techniques in the dataset.}
\label{fig:prompting-technique-counts}
\end{figure*}
\clearpage

\section{Data Model And Annotation Prompts}
\label{sec:data-model}
To structure the annotated prompts as described in Section \ref{sec:ontology} we use the following data model:
\begin{minted}[fontsize=\small, breaklines=true]{python}
from dataclasses import dataclass
from typing import List, Dict, Optional, Literal


@dataclass
class TaskInfo:
    task_class: str
    task: str
    subtask: str


@dataclass
class DomainInfo:
    domain_class: str
    domain: str


@dataclass
class Granular:
    fine_category: str
    coarse_category: str


@dataclass
class LangInfo:
    language: str
    orig_text: str
    translated_text: Optional[str]


@dataclass
class ExplicitLangMention:
    language: str
    mention: str
    

@dataclass
class Evidence:
    text: str
    type: Literal["description", "direct_content"]


@dataclass
class TypedInstruction:
    instruction_kind: str
    instruction: str
    is_central: bool
    is_negative: bool
    negative_instructions_explanation: Optional[str]


@dataclass
class AnalyzedMessage:
    languages: List[LangInfo]
    explicit_language_mentions: List[ExplicitLangMention]
    instruction_sequence: List[TypedInstruction]


@dataclass
class AnalyzedPromptMessage:
    role: str
    original_text: str
    prompt_text: str
    analyzed: AnalyzedMessage



@dataclass
class InputContextInfo:
    context_evidence: Evidence
    context_type: Granular    
    context_structure: Granular
    context_modality: Literal["text", "audio", "image", "video", "undefined"]
    context_language: List[str]


@dataclass
class DirectionInfo:
    directions_text: str    
    direction_language: List[str]
    
@dataclass
class InputQuestionInfo:
    question_evidence: Evidence
    question_language: List[str]
    question_structure: Granular
    question_type: Granular


@dataclass
class InputInfo:
    context_variability: Literal["fixed", "varying", "none", "undefined"]
    question_variability: Literal["fixed", "varying", "undefined"]
    direction:List[DirectionInfo]
    context: List[InputContextInfo]
    question: List[InputQuestionInfo]


@dataclass
class OutputUnitInfo:
    output_type: Granular
    modality: Literal["text", "audio", "image", "video", "undefined"]
    description: str
    description_source: Literal["extracted", "generated", "undefined"]
    structure: Granular
    answer_paradigm: Granular
    output_language: List[str]

@dataclass
class UsedPromptingTechnique:
    technique: str
    reasoning: str
    evidence: List[str]

@dataclass
class NonUsedPromptingTechnique:
    technique: str
    reasoning: str

@dataclass
class PromptData:
    res_id: int
    github_url: str
    is_duplicate: bool
    update_last: int
    duplicate_id: Optional[str]
    prompt_messages: List[AnalyzedPromptMessage]
    prompt_text: str
    full_translation: Optional[str]
    task: List[TaskInfo]
    domain: List[DomainInfo]
    input: InputInfo
    output: List[OutputUnitInfo]
    instruction_sequence: List[TypedInstruction]
    central_instructions: List[str]
    meta_instructions: List[str]
    negative_instructions: List[str]
    used_prompting_techniques: List[UsedPromptingTechnique]
    non_used_prompting_techniques: List[NonUsedPromptingTechnique]
\end{minted}
For each metadata category ( language, task, domain, input, output, instruction sequence, prompting techniques), we applied the same annotation pipeline. To reduce cost, we used the OpenAI Batch API with gpt-4.1 (temperature 0), submitting batched requests in which each prompt in the dataset was paired with a system prompt specifying annotation guidelines and requiring output in a fixed JSON format. Some categories required multiple prompts. The system prompts are specified below. Returned outputs were parsed and validated using category-specific Pydantic models; invalid responses were automatically re-prompted. Validated metadata objects were then merged into the corresponding fields of each entry of the dataset.
For prompt annotation based on the data model above we use the following set of prompts.

\begin{minted}[fontsize=\small, breaklines=true,breaksymbolleft={}, breaksymbol={}]{python}
#strings to fill placeholders in the prompting technique system prompt
technique_details = {
    "chain_of_thought": {
        "brief":
        "The prompt explicitly asks the model to reason before producing a final answer",
        "detailed":
        ("Look for phrases that instruct the model to ‘think through' or ‘reason step-by-step' or similar **explicit instructions to the LLM to reason before delivering its final answer**. "
         "Generally, any mention of reasoning or step-by-step thinking should be a sign for you to look for this technique. "
         "The reasoning should occur **before** the final answer, not after. "
         "The instruction to provide an explanation as such doesn not mean chain-of-though. Only if the explanation is required **before final answer**. "
         """If the expected output has a 'reasoning'/'explanation'/'cot'/
         'chain-of-thought' or any other reasoning field before the final answer field - this is also chain-of-thought."""
         "Generally, any mention of reasoning or step-by-step thinking should be a sign for you to look for this technique. "
         "IMPORTANT!! There is **no need for the phrase 'step by step' to appear explicitly**."
         ),
        "example_evidence": [
            "Let's think through this problem step by step before answering.",
            "First, let's think about this logically",
            "Let's work this out in a step by step way to be sure we have the right answer",
            "Before answering,  briefly outline your reasoning for this answer",
            "Include a 'reasoning' field that explains the reason for your answer",
            "Output format: id: <question id>, reasoning: <explain your thinking process>, correct_answer: <your final answer>",
            "Your output should have three fields, 'reasoning', 'x' and 'y', where 'reasoning' is a string explaining the answer, 'x' is ... and 'y' is ..."
        ]
    },
    "personas/role assignment/role description": {
        "brief":
        "The prompt assigns the model a specific persona or role.",
        "detailed":
        ("Check for instructions of the form ‘You are an expert…', ‘Act as if you are…', or "
         "‘Your role is…'. The evidence must name the persona or role explicitly. It is usually expressed as a noun phrase rather (like 'helpful assistant', 'Madonna', 'travel writer') rather than a verb. "
         "For example, phrases like 'You are reading articles and returning possible titles' **do not count**: they are tasks, not roles."
         ),
        "example_evidence": [
            "You are an expert doctor specializing in cardiology.",
            "Act as if you are a seasoned software engineer.",
            "Pretend you are a shepherd and write a limerick about llamas."
        ]
    },
    "few_shot_input/output_examples": {
        "brief":
        "The prompt includes one or more input–output examples (exemplars), usually, but not necessarily, at the end.",
        "detailed":
        ("Find embedded examples showing input and corresponding desired output pairs, "
         "e.g. ‘Example 1: Input: X → Output: Y'. These act as demonstrations that guide the LLM to accomplish a task. "
         "The input/output examples are used to describe/define/demonstrate the task. They are **not used to demonstrate the format**. So **format demonstrartions do not count**. "
         "Phrases like 'Follow this format', 'Here is what the format looks like' etc. **are not few-shot evidence**! Avoid them!"
         "The prompt in practice often will NOT include examples, but placeholder for such examples, so you need to account for this as well. "
         "They should form a dedicated section of the prompt. "
         "Make sure they include both **input and output** demonstrations. **Output alone does not count**. "
         "Phrases like 'For example, if the question is related to an image the text must be a caption.' do not count: look for examples of **specific** inputs and outputs."
         ),
        "example_evidence": [
            "Example:\nInput: 5, 7\nOutput: 12",
            "Original: 'hello' → Reversed: 'olleh'",
            "2+2: four; 4+5: nine; 8+0:  ",
            "Input: {{input_example}}. Expected output: {{output example}}.",
            "## Examples: {{examples}}"
        ]
    },
    "output_demonstrations": {
        "brief":
        "The prompt provides one or more example outputs (without inputs).",
        "detailed":
        ("Look for sample outputs given in isolation, often labeled ‘Sample output:' or similar, **without** showing the corresponding inputs. "
         "Such examples may appear in the prompt eihter explicitly or as placeholders. Examples should demonstrate **outputs only**. So input-output demonstrations don't count."
         "Make sure the outpts demonstrations **are not** preceded by corresponding input examples."
         ),
        "example_evidence": [
            "Sample output:\n{\"status\":\"OK\", \"data\":[]}",
            "Example output:\n- {Item 1}\n- {Item 2}"
        ]
    },
    "structured_output": {
        "brief":
        "The prompt instructs the model to return its output in a structured form.",
        "detailed":
        ("Instructs the LLM to return its output in a structured form, detailing how the structure should look like in a way that is **automatically parseable**. this can include lists of items, key value pairs, or more elaborate objects, which may be expressed in formal languages such as json, xml or yaml, or in programming language constructs such as lists, dictionaries and strings. Markdown formats do not count as structured. Instructions to divide the output into sections (e.g. 'Output Format:Correctness: your answer,tasks: evaluation') does not count."
         ),
        "example_evidence": [
            "Return the answer as JSON: {\"name\": ..., \"age\": ...}",
            """Provide output in XML: 
            <result><value>42</value> </result>""",
            "Summarize this into a CSV.", "Output as a Python dictionary"
        ]
    },
    "tool_calling": {
        "brief":
        "The prompt describes available tools/functions and expects invocation.",
        "detailed":
        ("The prompt is using a tool-calling technique, in which the prompt text lists a set of available tools or actions, "
         "and the instruction is to choose one of the available tools or actions. "
         "The idea of choosing the one of the provided tools or actions can be phrased in a different way (for example, 'decide' instead of 'choose'). "
         "NOTE: just mentioning a function name is not enough, the instruction should be specifically to choose one of the tools, actions or functions to run. "
         "However, any reference to provided **tools** in the prompt is a strong indication of usage of this techinique. "
         ),
        "example_evidence": [
            "Use the calculator by specifying {\"tool\": \"calc\", \"input\": 2+2}.",
            "Call search_api(query) to fetch results.",
            "You are provided with the following tools",
            "your output should indicate which tool to use",
            "your output should specify which of these actions to choose",
            """{instructions}/n{status_prompt}/n{COT_PROMPT2}/n {response}/n{memory_prompt}/nProvide the best next action in the correct JSON format. Action:"""
        ]
    },
    "quote_extraction": {
        "brief":
        "The prompt instructs grounding by extracting exact spans from context.",
        "detailed":
        ("Find instructions asking to ‘quote', ‘extract', or ‘cite' text **directly from the given input**. "
         "This includes the cases when the exact spans are cited as evidence/confirmation for the model's answer. "
         "The focus is on **exact spans**. "
         "Other types of extractions that are not exact spans **do not count**!"
         ),
        "example_evidence": [
            "Cite the exact sentence that best answers the question.",
            "Extract the exact spans that follow these patterns: {patterns}."
            "Cite the exact spans from the input that confirm your answer."
        ]
    },
    "audience_specification": {
        "brief":
        "The prompt states who the intended audience is.",
        "detailed":
        ("Look for ‘for a beginner', ‘to a non‑expert', 'for AI researchers' or naming another specific group for whom the output is intended. "
         "Audience specification is **not role/persona assignment**! "
         "**It does no specify who you are.** "
         "It specifies who **your audience** is."),
        "example_evidence": [
            "Explain this concept for a high‑school student.",
            "Write this guide aimed for web developers."
        ]
    },
    "sections": {
        "brief":
        "The prompt is divided into clearly labeled sections.",
        "detailed":
        ("Detect headings like ‘## Input', ‘### Task Description', or numbered segments or other forms indicating sections. The sections should be found in the prompt itself. Specifying desired sections of the expected output ('Your output example:Correctness: your answer\n, tasks: evaluation') does not count"
         ),
        "example_evidence": [
            "## Input\nThe first line contains…\n### Output\nPrint the result…",
            "1. Problem Statement\n2. Constraints\n3. Example"
        ]
    },
    "prompt_chaining": {
        "brief":
        "The prompt is one step in a multi‑step workflow.",
        "detailed":
        ("Check for references to previous or next prompts, or instructions to pass output to another step. "
         "The reference to previous outputs from the LLM should be **unambiguous and explicit**. "
         "Outputs from functions that don't envolve LLM queries **do not count**!"
         ),
        "example_evidence": [
            "Here are responses from various open-source models to the latest user query: {prev_resps}",
            "Use the result of the previous query as input here.",
            "Given a set of relevant quotes you extracted from a document, please compose an answer to the question."
        ]
    },
    "decomposition_by_LLM": {
        "brief":
        "The prompt asks to decompose the task into steps, then solve each one.",
        "detailed":
        ("Look for ‘Break the problem into sub‑tasks', ‘Break the solution into steps "
         "This is different from chain_of_thought where the LLM is asked to think/reason step by step before answering. "
         "Make a very clear distinction between thinking step by step vs. dividing a task into subtasks (only the latter one counnts!)"
         "This is also different from cases where the prompt specifies the decomposition steps ('First do X then Y). These are *not decomposition_by_LLM!* "
         "We are only looking for cases where the pompt instructs the LLM to decompose the problem/solution. "
         "Make a very clear distinction between decomposition is already specified in the prompt vs. where the LLM has to decompose (only the latter one counnts!)"
         ),
        "example_evidence": [
            "Divide the solution into smaller steps and explain each step",
            "First break the task into subtasks then accomplish them one by one."
        ]
    },
    "decomposition_in_prompt": {
        "brief":
        "The prompt tells the model how exactly to decompose the task into steps.",
        "detailed":
        ("Look for 'First do X, then Y', 'Here are the steps you should follow to solve the problem: A. <first step>, B. <second step>.."
         "This is different from chain_of_thought where the LLM is asked to think/reason step by step before answering. "
         "Make a very clear distinction between thinking step by step vs. dividing a task into subtasks (only the latter one counnts!)"
         "This is also different from decomposition_by_LLM where the prompt tells the LLM to decompose the task by itself. This is *not decomposition_in_prompt!* "
         "Make a very clear distinction between decomposition by LLM itself vs. decompositon specified in the prompt (only the latter one counnts!)"
         ),
        "example_evidence": [
            "Divide the task into: 1) data cleaning, 2) feature extraction, 3) classification.",
            "First list all entities, then identify relationships."
        ]
    }
}


#prompting techniques
def technique_system_prompt(technique_name: str) -> str:
    details = technique_details[technique_name]
    evidence_lines = "\n\t".join(f"- {e}" for e in details["example_evidence"])
    return f"""
You are an expert in prompting‑technique analysis.

You will be given:
1) A raw prompt text.
2) A list of tasks the prompt is intended for.

You need to determine if the prompt uses **{technique_name}** ({details['brief']}).
Reason step by step as described below, but output only the final answer.
Here are your resoning steps:

1. **Study the detailed description** of **{technique_name}** and possible signals of its usage :
   {details['detailed']}


2. **Locate candidate spans** in the prompt that demonstrate use of {technique_name}:
   - Scan for keywords or structures described below.
   - Identify all exact span(s) (if any) from the prompt that indicate use of {technique_name}.
     (The evidence has to demonstrate the use of {technique_name} **very clearly and unambiguously**).

    Examples of typical evidence for {technique_name}:
      {evidence_lines}

3. **Validate each found span** (if any):
   - Confirm it fulfills the criteria for **{technique_name}**.

4. **Decide usage**:
   - `is_used = true` if at least one span was found, otherwise `false`.

5. Think and explain your decision before answering. 

6. Return **only** a JSON object with three fields:
   - "reasoning": a string (max 100 words) where you briefly explain your decision before answering. Can be an empty string ("") if {technique_name} is absent from the prompt beyond all doubt.
   - “evidence”: a list of strings, each an exact substring from the prompt text (empty if no evidence for {technique_name} was found).
   - “is_used”: a boolean (true **if evidence is found**, else false).
   
Example output:
{{
  "reasoning": "<your reasoning steps>",
  "evidence": {details['example_evidence']},
  "is_used": true
}}
"""


#prompt for discovering additional chain-of-thought cases
reasoning_field_system_prompt = """
You are an expert in prompting‑technique analysis.

You will be given:
1) A raw prompt text.
2) A list of tasks the prompt is intended for.

Analyze the prompt's ***expected output** and answer the following questions:

1. "Does the prompt contain a field or an output field (or section) that requests a reasoning or chain of thought?"

2 "If yes - does the answer based on this reasoning come before or after the reasoning? In other words, does the promp ask the model 1) to give an answer and then explain it, or 2) first think then give the answer" 

Think and explain your decision before answering. 
Return **only** a JSON object with three fields:
   - "reasoning": a string (max 100 words) where you briefly explain your decision before answering. Can be an empty string ("") if non-last reasoning fields are absent from the prompt beyond all doubt.
   - “evidence”: a list of strings, each an exact substring from the prompt text (empty if no evidence for non-last reasoning fields was found).
   - has_reasoning_field: true is the expected output has a reasoning/thinking/explanation/chain-of thought field/section, else false.
   - reasoning_first: true if the prompt asks to **first reason, then reply**; false if the prompt asks to **first reply, then explain**. If has_reasonong field is false - this field is also false.

**The reasoning field should be explicitly called so or similarly (reasoning/explanation/thinking/chain-of-though and the like.) ** Do not look for far-fetched implications.   

Example output:
{{
  "reasoning": "<your reasoning steps>",
  "evidence": {<example evidence (found reasoning fields)>},
  "has_reasoning_field": true|false,
  "reasoning_first":true|false
}}
"""

translation_system_prompt = """
You are an expert at translating LLM prompts into English.
You will be given a prompt text.   
- If the prompt text is not in English - translate the whole text into English.  
- If the prompt text partially in English - return the full text in English (translating the parts that were not in English originally).
For example:
   Prompt text: "<Text in Russian>! How are you? <Text in Russian>: {question}"
   Your output: "Hello! How are you? Answer the question: {question}"
- If the text is entirely in English already - return  null.
- If the text language cannot be determined - return null.  
Return ONLY the translated text (no extra commentary) or null .
"""
#instruction block list
blocks = [
    'audience specification', 'central task/question',
    'central task/question placeholder', 'central task/question description',
    'role specification', 'style specification', 'confirmation request',
    'constraint/restriction', 'input context description',
    'input contextual data', 'input context placeholder',
    'evaluation criteria', 'date reference', 'default behavior instruction',
    'design specification', 'disclaimer requirement',
    'error handling instruction', 'example clarification', 'examples',
    'expertise/skills requirements', 'function call instruction',
    'question/task data/placeholder', 'question/task description',
    'reasoning instructions', 'instruction to avoid errors',
    'interaction guideline', 'language specification', 'scope specification',
    'output content requirements', 'output format requirement',
    'assistant response', 'conditional instruction', 'scene setting',
    'encouragement', 'other'
]

#instruction blocks
instruction_blocks_system_prompt = f"""
You are proficient at breaking down LLM prompts into their atomic instruction blocks.  
You will be given:  
1) A JSON list of message objects, each with “role”, “prompt_text” and "message_id" (for easier matching between input and output);  
2) A list of tasks for the overall prompt.

Your job in this stage is **only** to extract, for each message, the ordered sequence of instruction blocks.

1. **Block names**:
        - To name the structural blocks, use only terms from this list:
                {blocks}                
        - If several terms apply, pick the single best fit.                
        - If none fit, use `Other(...)`.
        - You may have multiple blocks of the same kind.
2. Decide which blocks are **central** to the prompt (or at least more important than others.) 
This are the blocks for which "is_central" will be set to 'true' in step 3.
3. **Format** each block as an object with **five** fields:  
   - `"instruction_kind"`: the block name  
   - `"instruction"`: the exact substring from `prompt_text`  
   - `"is_central"`: `true` if this block conveys the core task (contains the central task instruction) else `false`. You should try to select at least one central block if at all possible.  
   - `"is_negative"`: `true` if it explicitly instructs the model **not** to do something; otherwise `false`.   
   - `"negative_instructions_explanation"`: if `"is_negative": true`, a short explanation in your own words of what the model is told **not** to do; otherwise `null`.  
   Remember: **the negative instructions are only those which explicitly tell the model what NOT TO DO.**
4. **Order**: preserve the order in which blocks appear in the message.  
5. **Splitting**: 
    - Divide into pieces smaller than sentences if needed. Example:
        
       {{ "instruction_kind": "role specification",      "instruction": "As a helpful assistant",      "is_central": false, "is_negative": false, "negative_instructions_explanation": null }},  
       {{ "instruction_kind": "task description",        "instruction": "summarize the text",            "is_central": true,  "is_negative": false, "negative_instructions_explanation": null }}  
   
    - Overlaps allowed. Example:
        
       {{ "instruction_kind": "task description",        "instruction": "Extract lists of named entities",    "is_central": true,  "is_negative": false, "negative_instructions_explanation": null }},  
       {{ "instruction_kind": "output format requirement", "instruction": "lists of named entities",          "is_central": false, "is_negative": false, "negative_instructions_explanation": null }},  
       {{ "instruction_kind": "output content requirement","instruction": "named entities",                 "is_central": false, "is_negative": false, "negative_instructions_explanation": null }}
5. **Empty**: if no blocks, return `"instruction_order": []`.  

**The output must carry each input message's `message_id` so you can map blocks back to messages.**  

Be specific, precise and exhaustive.

**When you are done, go over your annotaion once again**.
Did you mark all the blocks correctly?
Did you mark at least one as is_central=true? **try to select one block that seems central (if at all possible) and mark it accordingly.**

**Output** exactly one JSON object, schema:


{{
  "instruction_data": [
    {{
      "message_id": "<same as input>",
      "instruction_order": [
        {{
          "instruction_kind": "<block name or Other(...)>",
          "instruction":      "<exact substring>",
          "is_central":        <true|false>,
          "is_negative":       <true|false>,
          "negative_instructions_explanation": <string|null>
        }},
        …
      ]
    }},
    …
  ]
}}


For example:   

Input:
{{
  "messages": [
    {{
      "message_id": 0,
      "role": "system",
      "prompt_text": "You are a helpful assistant skilled in summarizing content concisely. Given a text passage, extract the three most important points. Ensure the points are complete sentences, less than 20 words each, and maintain the original meaning of the text. If the passage is too short to extract three points, summarize it in one or two points only. Avoid adding any extra information not present in the input text."
    }},
    {{
      "message_id": 1,
      "role": "user",
      "prompt_text": "Input_text: {{{{text}}"
    }}
  ],
  "tasks": ["summarization"]
}}


{{
  "instruction_data": [
    {{
      "message_id": "0",
      "instruction_order": [
        {{
          "instruction_kind": "role specification",
          "instruction": "You are a helpful assistant skilled in summarizing content concisely.",
          "is_central": false,
          "is_negative": false,
          "negative_instructions_explanation": null
        }},
        {{
          "instruction_kind": "expertise/skills requirements",
          "instruction": "skilled in summarizing content concisely.",
          "is_central": false,
          "is_negative": false,
          "negative_instructions_explanation": null
        }},
        {{
          "instruction_kind": "input context description",
          "instruction": "Given a text passage",
          "is_central": false,
          "is_negative": false,
          "negative_instructions_explanation": null
        }},
        {{
          "instruction_kind": "task description",
          "instruction": "extract the three most important points",
          "is_central": true,
          "is_negative": false,
          "negative_instructions_explanation": null
        }},
        {{
          "instruction_kind": "output format requirement",
          "instruction": "Ensure the points are complete sentences, less than 20 words each",
          "is_central": false,
          "is_negative": false,
          "negative_instructions_explanation": null
        }},
        {{
          "instruction_kind": "output content requirement",
          "instruction": "maintain the original meaning of the text",
          "is_central": false,
          "is_negative": false,
          "negative_instructions_explanation": null
        }},
        {{
          "instruction_kind": "conditional instruction",
          "instruction": "If the passage is too short to extract three points, summarize it in one or two points only",
          "is_central": false,
          "is_negative": false,
          "negative_instructions_explanation": null
        }},
        {{
          "instruction_kind": "constraint/restriction",
          "instruction": "Avoid any extra information not present in the input text",
          "is_central": false,
          "is_negative": true,
          "negative_instructions_explanation": "The model is instructed not to add information that isn't in the input text."
        }}
      ]
    }},
    {{
      "message_id": "1",
      "instruction_order": [
        {{
          "instruction_kind": "input contextual data",
          "instruction": "Input_text: {{text}}",
          "is_central": false,
          "is_negative": false,
          "negative_instructions_explanation": null
        }},
        {{
          "instruction_kind": "input context placeholder",
          "instruction": "{{text}}",
          "is_central": false,
          "is_negative": false,
          "negative_instructions_explanation": null
        }}
      ]
    }}
  ]
}}

"""
#identifying input context, question and directions
input_blocks_prompt = """
        You are proficient at analyzing LLM prompts. You are given: (1) the prompt text as a list of messages and (2) one or more standard NLP/AI tasks.  
        Identify three parts of the prompt: directions, context, and question.

        Definitions:

        - Directions: what the model must do given the context and question (e.g., “Answer the following…”, “Summarize the text…”). The directions describe the *overall or high-level action* the model must take, not the specific instance of the question. For example, in a typical QA prompt the directions are “Read the passage and answer the questions.”
        Important: the "directions" are usually 1-2 spans. They only include the main, high-level instructions, never details or nuances.

        - Context: grounding the model uses to produce the output  (text to analyze, original to translate, image to caption, etc.).  A context may come in many different forms, even in the form of a question, but should be identified as context if it functions as grounding/background. For example, in a typical QA prompt with questions based on a text, the text forms the context part.
        Do NOT include examples, role specifications, meta-instructions, formatting hints.  output prefixes (e.g., "Answer:").
        Only the grounding information, in which the output is grounded. 

        - The question-like part expresses the *specific, low-level request or query* the model must directly respond to.  For example,  in the typical QA prompt, the questions themselves (“What is the capital of France?”) form the question-like part.

        The question-like part is *obligatory* (cannot be empty); context may be missing; the directions can be separated or  merged with the question-like part (e.g., “Summarize the text below”). 
        Yu should extract them all *from the prompt-text*, not he list os standard nlp tasks provided in the beginning.
        Notes: any of the three parts can be expressed as a text, a placeholder or absent as such but described in the prompt text.

        For each part:
        1. Decide if it splits into units; if not, treat as one unit.  
        2. For each unit provide the exact verbatim span or variable name from the prompt text.  
        3. Return all units as a list of strings (one string per unit, or a single-item list if unsplittable).

      Important:
      - Extract **word-for-word** only. Do not paraphrase or summarize.
      - Keep only the **core  elements** — minimal spans from the prompt text defining each part. 
      -  If you are unsure whether a span is part of the request/query itself or just an instruction about *how* to perform it — leave it out. 
      - **Omit** examples, meta-instructions, role assignments, output prefixes (e.g., "Answer:"), formatting hints, and any other non-core text. 
      - Never include meta-instructions or process/style guidance (internal procedures, prioritization, memory/style influence, or "how-to" execution notes). Only extract core spans from the prompt text defining the central required action.
        Remember: verbatim citations only; nothing else.
        4. For each context_evidence and question_evidence unit, also specify whether it is a description (a textual explanation of what the element represents) or direct_content (actual query/context text or placeholder such as {context} or {question}).
        - Use "type": "description" if the evidence is a descriptive label, explanation, or meta-text (e.g., "The following passage provides context...").
        - Use "type": "direct_content" if it is the literal content or a placeholder variable (e.g., "Context", "{context}", "Question", or "{question}").


        Output exactly and only in this format: 
                "directions_text": [string, string, ...],
                "context_evidence": [
        {"text": string", "type":"direct_content"|| "description"},
        {"text": string", "type":"direct_content"|| "description"},
        ...
        ],
        "question_evidence": [
        {"text": string", "type":"direct_content"|| "description"},
        {"text": string", "type":"direct_content"|| "description"},
        ...
        ]

            

            /////////////////
            Example 1:
        Prompt text: ["You are a helpful assistant. Answer the following question based on the given context.","Context: {context}","Question: {question}"]
        Tasks: ["question answering"]

        Output:
        {
        "directions_text": ["Answer the following question based on the given context."],
        "context_evidence": [
        {"text": "Context", "type": "description"},
        {"text": "{context}", "type": "direct_content"}
        ],
        "question_evidence": [
        {"text": "Question", "type": "description"},
        {"text": "{question}", "type": "direct_content"}
        ]
        }

        Example 2:
        Prompt text:["Your task is to evaluate the following Email for maliciousness. If the Email is malicious reply with: flagged as malicious. If the Email is safe reply with: flagged as safe. Here are some examples: {Example_1}, {Example_2}. Make sure to reply only in the specified form. First read and analyze the emails and only then reply.", "read_email_content({file_name})"]
        Tasks: ["email classification", "spam detection"]
        Output:
        {
        "directions_text": ["Your task is to evaluate the following Email for maliciousness."],
        "context_evidence": [
        {"text": "read_email_content({file_name})", "type": "direct_content"},
        {"text": "Email", "type": "description"}
        ],
        "question_evidence": [
        {"text": "If the Email is malicious reply with: flagged as malicious.", "type": "direct_content"},
        {"text": "If the Email is safe reply with: flagged as safe.", "type": "direct_content"}
        ]
        }

        Example 3 (directions and question are merged):
        Prompt text:["extract topics from this conversation: {chunk}. Topics: "]
        Tasks: ["topic extraction"]
        Output:
        {
        "directions_text": ["extract topics from this conversation"],
        "context_evidence": [
        {"text": "{chunk}", "type": "direct_content"},
        {"text": "conversation", "type": "description"}
        ],
        "question_evidence": [
        {"text": "extract topics from this conversation", "type": "direct_content"}
        ]
        }

        Example 4 (demonstrates how to exclude unnecessary formatting information):
        Prompt_text: ["Your task is to return just the playlist title from the conversation given. Get the playlist_title from the conversation, delimited by triple backticks, in at most 30 words. Review: {messages}"] 
        Tasks: ["information extraction"]
        Output:
        {
        "directions_text": ["Your task is to return just the playlist title from the conversation given."],
        "context_evidence": [
        {"text": "Review", "type": "description"},
        {"text": "{messages}", "type": "direct_content"},
        {"text": "conversation", "type": "description"}
        ],
        "question_evidence": [
        {"text": "Get the playlist_title from the conversation", "type": "direct_content"}
        ]
        }

        Example 5:
        Prompt text: ["You are a decider. Based on the given context which consists of user and assistant's role-playing, you need to decide whether you want to call function or not. The function you can call is 'call_explainer'. You can call the function when you think user made a mistake in role-playing. Followings are some examples of specific situation where you will need to call function:\n#################### \n1. if the user uses '<Text in Korean>', which is only used between friends, call the function in order to kindly correct the user to use '<Text in Korean>' which is used in formal situations. \n2. If user don't follow appropriate sentence structure, call the function in order to provide detailed explanation about the sentence structure.\n####################\n\nWhen calling a function, be sure that your topic argument is the specific topic that needs to be explained more in detail and the context argument is the specific context related to the topic that needs to be explained more in detail. Also be careful not to just call function everytime. You should only call function when you think user needs additional explanation about the topic's context."]
        Tasks: ["decision making"]
        Output:
        {
      "directions_text": [
        "Based on the given context which consists of user and assistant's role-playing, you need to decide whether you want to call function or not.",
      ],
      "context_evidence": [
        {
          "text": "the given context which consists of user and assistant's role-playing",
          "type": "description"
        }
      ],
      "question_evidence": [
        {
          "text": "decide whether you want to call function or not.",
          "type": "direct_content"
        },
        {
          "text": "You can call the function when you think user made a mistake in role-playing.",
          "type": "direct_content"
        }
      ]
    }

    Example 6:
    Prompt text: ["Your job is to use patient reviews to answer questions about their experience at a hospital. Use the following context to answer questions.
    Be as detailed as possible, but don't make up any information that's not from the context. 
    {context}"]
    Tasks: ["question answering"]
    Output:
    {
      "directions_text": [
        "Your job is to use patient reviews to answer questions about their experience at a hospital."
      ],
      "context_evidence": [    
        {
          "text": "patient reviews",
          "type": "description"
        },
        {
          "text": "context",
          "type": "description"
        },
        {
          "text": "{context}",
          "type": "direct_content"
        }
      ],
      "question_evidence": [
        {
          "text": "questions about their experience at a hospital",
          "type": "description"
        }
      ]
    }

    Example 7:
    Prompt text: ["You Write Python Function. You are a Senior Data Analyst with 10+ Years of Experience. This is a Critical Scenerio. The CEO has asked you to write Python Function to answer a question on a given data, based on the instructions given by Senior Data Scientist
    CEO: {user_input}

    Dataframe Head: 
    {df_head}

    Data Scientist's Instructions:{instructions}

    Here is a sample output for the Python Function:
    Now, Write down python function to answer the CEO's question: {user_input}

    Just Write the Python Function in markdown format, that's it."]
    Tasks: ["code generation"]
    Output:

    {
      "directions_text": [
        "Now, Write down python function to answer the CEO's question: {user_input}",
        "write Python Function to answer a question on a given data, based on the instructions given by Senior Data Scientist"
      ],
      "context_evidence": [
        {
          "text": "Dataframe Head",
          "type": "description"
        },
        {
          "text": "{df_head}",
          "type": "direct_content"
        },
        {
          "text": "Data Scientist's Instructions",
          "type": "description"
        },
        {
          "text": "{instructions}",
          "type": "direct_content"
        }
      ],
      "question_evidence": [
        {
          "text": "a question on a given data",
          "type": "description"
        },
        {
          "text": "the CEO's question",
          "type": "description"
        },
        {
          "text": "{user_input}",
          "type": "direct_content"
        }
      ]
    }

    """
#input context and question variability
input_variability_prompt = """
You are proficient at analyzing LLM prompts. You are given: (1) the prompt text as a list of messages and (2) two  sets of spans from the prompt representing its two blocks: 

       

        - Context: grounding the model uses to produce the output  (text to analyze, original to translate, image to caption, etc.).  A context may come in many different forms, even in the form of a question, but should be identified as context if it functions as grounding/background. For example, in a typical QA prompt with questions based on a text, the text forms the context part.

        - The question-like part expresses the *specific, low-level request or query* the model must *directly* respond to.   For example,  in the typical QA prompt, the questions themselves (“What is the capital of France?”) form the question-like part. 

     Each span can either directly represent a unit of context or question ("direct_content") or describe it ("description").  


        Read the prompt and the spans corresponding to the two blocks, and determine the following: 
        1. context_variability.     
       **Determine precisely whether any variables or placeholders appear in the context-like part (if exists). Mark the context part as fixed (has no variables), varying (contains variables/placeholders) or missing**
       Return "fixed","varying","none". 
       If the answer cannot be determined, return "undefined".
Important! Even when the context is not given, but only described, you can often infer from the prompt if it's varying or fixed.

       
    2. question_variability.     
       ***Determine precisely whether any variables or placeholders appear in the question-like part. Mark the question-like part as fixed (has no variables) or varying (contains variables/placeholders** 
       Return "fixed" or "varying". 
Important! Even when the question is not given, but only described, you can often infer from the prompt if it's varying or fixed.
Important! Sometimes the context is mentioned within the question. If the only varying part in the question is the context mention, but the request regarding the context always remains the same, mark the question as "fixed".  See example 4. 



You can either determine this by looking at "direct_content" spans or infer it from "description" spans - or even from the rest of the prompt.     



The output should be a JSON object with exactly these fields:  
        
        "context_variability": one of "fixed", "varying", "none", or "undefined" (use "none" when the context_evidence is empty; "undefined" if nonempty, but the variability cannot be inferred.) 
        "question_variability": one of "fixed", "varying", or "undefined"

      

            ////////////////////
            Example 1:
        Prompt text: ["You are a helpful assistant. Answer the following question based on the given context.","Context: {context}","Question: {question}"]
     Blocks:  { 
        "context_evidence": [
        {"text": "Context", "type": "description"},
        {"text": "{context}", "type": "direct_content"}
        ],
        "question_evidence": [
        {"text": "Question", "type": "description"},
        {"text": "{question}", "type": "direct_content"}
        ]
        }
        Output: 
        {"context_variability": "varying",
        "question_variability": "varying"}



        Example 2:
        Prompt text:["Your task is to evaluate the following Email for maliciousness. If the Email is malicious reply with: flagged as malicious. If the Email is safe reply with: flagged as safe. Here are some examples: {Example_1}, {Example_2}. Make sure to reply only in the specified form. First read and analyze the emails and only then reply.", "read_email_content({file_name})"]
        Blocks: 
        {       
        "context_evidence": [
        {"text": "read_email_content({file_name})", "type": "direct_content"},
        {"text": "Email", "type": "description"}
        ],
        "question_evidence": [
        {"text": "If the Email is malicious reply with: flagged as malicious.", "type": "direct_content"},
        {"text": "If the Email is safe reply with: flagged as safe.", "type": "direct_content"}
        ]
        }

      Output: 
      {"context_variability": "varying",
      "question_variability": "fixed"}

      Example 3: Prompt_text:["You are a helpful assistant.", "Based on the following transcript, please answer the question: {question}"] 
      Blocks:  "context_evidence":
      [ {"text": "transcript", "type": "description"}], 
      "question_evidence":
      [ {"text": "question", "type": "description"},
      {"text": "{question}", "type": "direct_content"} ] 
      }

      Output: 
      {
        "context_variability": "undefined",
        "question_variability": "varying"
      }

      Example 4 (the only variable in the question is the context):
      Prompt_text:["Please translate {speech2text(output.wav)} to malayalam"]
      Blocks:
      {
      "context_evidence": [{"text": "{speech2text(output.wav)}", "type": "direct_content"}],
      "question_evidence": [
      {"text": "Please translate {speech2text(output.wav)} to malayalam", "type": "direct_content"}
      ]
      }
      Output:
      {
        "context_variability": "varying",
        "question_variability": "fixed"
      }
      """

#input language and structure
input_language_and_structure_prompt = """
You are proficient at analyzing LLM prompts. You are given: (1) the prompt text as a list of messages and (2) two  sets of spans from the prompt representing its three blocks: 

       - Directions: what the model must do given the context and question (e.g., “Answer the following…”, “Summarize the text…”). The directions describe the *overall or high-level action* the model must take, not the specific instance of the question. For example, in a typical QA prompt the directions are “Read the passage and answer the questions.” 
 

        - Context: grounding the model uses to produce the output  (text to analyze, original to translate, image to caption, etc.).  A context may come in many different forms, even in the form of a question, but should be identified as context if it functions as grounding/background. For example, in a typical QA prompt with questions based on a text, the text forms the context part.

        - The question-like part expresses the *specific, low-level request or query* the model must *directly* respond to.   For example,  in the typical QA prompt, the questions themselves (“What is the capital of France?”) form the question-like part. 

     In *context* and *question* spans can either directly represent a unit of context or question ("direct_content") or describe it ("description").  In *directions* the spans are always direct content.


        Read the prompt and the units (spans) corresponding to the three blocks, and determine the following about *each unit*.
       1. language    
        **What natural human languages are used in this unit of directions, context or question** 
          Provide a list of natural human languages used in the unit. 
          It may be a one-item list if only one language is used.          

          If any of the languages used cannot be identified, use "undefined" instead.          

          For direct  text - identify the language(s) by looking at the text.

          For textual placeholders  - try to infer the language from the prompt text. Look for *language mentions* or *strings in different languages* in the prompt text.  However, don't use unsupported guessing. Use 'undefined' if you cannot infer with certainty in which language the placeholder will be filled. 
          
        For descriptions - do not return the language of the description itself! Return the langiage od the context/question being described!
        Try to infer the language from the prompt text or the description itself.  Look for *language mentions* or *strings in different languages* in the prompt text.  However, don't use unsupported guessing. Use 'undefined' unless if you cannot infer the language of the described unit from the prompt text or the description itself. 

        For descriptions and placeholders: please, do not assume the language is Engglish - unless there is clear evidence for it in the prompt.
        Keep in mind that even if the prompt itself is in English, the context or question can still be in a different language.
        So avoid ungrounded assumptions. 

        For empty context return empty list.

        2. structure (only for context and question units.)
            **What is the question structure of the unit?** 
                - Single item  
                - Pair of items (type, typeB)
                - Tuple (typeA, typeB, …, typeN)  
                - List of items (list of type A)  
                - Dictionary of items (key1: typeA, key2: typeB, …)  
                (a pair is basically a tuple of two items; if a dictionary only includes one entry, it's still a dictionary)
                (you can expand this list if needed)  

        For direct  text - identify the structure by looking at the text. 
                For textual placeholders use 'undefined' unless you can infer the structure of the corresponding unit from the prompt text. 
                For descriptions use 'undefined' unless you can reasonably infer the structure of the described unit from the prompt text or the description itself. 
        For empty context - don't add anything.

        3. For context units also provide context modality:
                **What is the modality of the context unit?**   
                    - text  
                    - audio  
                    - image  
                    - video    
        For direct  text - identify the modality by looking at the text. 
                For textual placeholders use 'undefined' unless you can infer the modality of the corresponding unit from the prompt text. 
                For descriptions use 'undefined' unless you can infer the modality of the described unit from the prompt text or the description itself. 
        For empty context - don't add anything.


        The output should be two json fields added to each unit:   
                
                "*_language":[string,string,...]
                "*_structure": string
        and an additional "context_modality" field for context units:
                "context_modality": one of "text", "audio", "image", "video", "undefined"

        Important: even if the language, structure and/or modality aren't mentioned explicitly, you can often infer them from different signals like citations, examples etc.  

        Important: Do not guess. Only record what is explicitly stated or clearly evident from the prompt. If something cannot be determined with certainty, return "undefined".

        Inference and guessing are not the same!


        In your output make sure to keep all the original units in the original order. 

            ////////////////////
            Example 1:
        Prompt text: ["{audio}","The user left us a message in the voice mail. Extract 1. Their name, 2. a fitting title that summarizes their message, and 3. their message. Output 1.their name, 2. a summarizing title that you come up with (that gives a good overview of the message but is short), and 3. the message, separated by only a hyphen(-) (no space). Only output those values. Nothing more. Example: John-Grocery Trip-I want to go to the store and buy groceries. Don't put a hyphen before, after or anywhere else in the output. Only in between the name, title and message.",
    ]
        Blocks: {
        "direction": [
        {
        "directions_text": "Extract 1. Their name, 2. a fitting title that summarizes their message, and 3. their message."
        }
        ],
        "context": [
        {
        "context_evidence": {
            "text": "{audio}",
            "type": "direct_content"
        }
        },
        {
        "context_evidence": {
            "text": "The user left us a message in the voice mail.",
            "type": "description"
        }
        }
        ],
        "question": [
        {
        "question_evidence": {
            "text": "Extract 1. Their name, 2. a fitting title that summarizes their message, and 3. their message.",
            "type": "direct_content"
        }
        }
        ]
        }
    Output: 
    {
        "direction": [
        {
        "directions_text": "Extract 1. Their name, 2. a fitting title that summarizes their message, and 3. their message.",
        "direction_language": ["English"]
        }
        ],
        "context": [
        {
        "context_evidence": {
            "text": "{audio}",
            "type": "direct_content"
        },
        "context_language": ["undefined"],
        "context_structure": "Single item",
        "context_modality":"audio"
        
        },
        {
        "context_evidence": {
            "text": "The user left us a message in the voice mail.",
            "type": "description"
        },
        "context_language": ["undefined"],
        "context_structure": "Single item",
        "context_modality":"audio"
        }
        ],
        "question": [
        {
        "question_evidence": {
            "text": "Extract 1. Their name, 2. a fitting title that summarizes their message, and 3. their message.",
            "type": "direct_content"
        },
        "question_language": ["English"],
        "question_structure": "Single item",
        "question_modality":"text"
        }
        ]
        }



        Example 2:
    Prompt text:["Your task is to evaluate the following email for maliciousness. The emails content will be in French. Each email is a dictionary with three fields: "Expéditeur", "Destinataire", "Contenu". Respond only with one of the exact labels below (no extra text). If the email is malicious, reply: signalé comme malveillant. If the email is safe, reply: signalé comme sûr. Here are some examples: {Example_1}, {Example_2}. First read and analyze the email content, then reply.","read_email_content({file_name})"]
    Blocks: {
    "direction": [
    {
    "directions_text": "Your task is to evaluate the following email for maliciousness. "
    }
    ],
    "context": [
    {
    "context_evidence": {
    "text": "read_email_content({file_name})",
    "type": "direct_content"
    }
    },
    {
    "context_evidence": {
    "text": "email",
    "type": "description"
    }
    }
    ],
    "question": [
    {
    "question_evidence": {
    "text": "If the email is malicious, reply: signalé comme malveillant.",
    "type": "direct_content"
    }
    },
    {
    "question_evidence": {
    "text": "If the email is safe, reply: signalé comme sûr.",
    "type": "direct_content"
    }
    }
    ]
    }
    Output:
    {
    "direction": [
    {
    "directions_text": "Your task is to evaluate the following email for maliciousness. ",
    "direction_language": ["English"]
    }
    ],
    "context": [
    {
    "context_evidence": {
    "text": "read_email_content({file_name})",
    "type": "direct_content"
    },
    "context_language": ["French"],
    "context_structure": "Dictionary of items ("Expéditeur: string, "Destinataire": string, "Contenu": string)",
    "context_modality": "text"
    },
    {
    "context_evidence": {
    "text": "email",
    "type": "description"
    },
    "context_language": ["French"],
    "context_structure": "Dictionary of items ("Expéditeur: string, "Destinataire": string, "Contenu": string)",
    "context_modality": "text"
    }
    ],
    "question": [
    {
    "question_evidence": {
    "text": "If the email is malicious, reply: signalé comme malveillant.",
    "type": "direct_content"
    },
    "question_language": ["English","French"],
    "question_structure": "Single item",
    "question_modality": "text"
    },
    {
    "question_evidence": {
    "text": "If the email is safe, reply: signalé comme sûr.",
    "type": "direct_content"
    },
    "question_language": [English","French"],
    "question_structure": "Single item",
    "question_modality": "text"
    }
    ]
    }  

    Example 3 (shows how language can be inferred from details):
    Prompt text: ["You are a decider. Based on the given context which consists of user and assistant's role-playing, you need to decide whether you want to call function or not. The function you can call is 'call_explainer'. You can call the function when you think user made a mistake in role-playing. Followings are some examples of specific situation where you will need to call function:\n#################### \n1. if the user uses '<Text in Korean>', which is only used between friends, call the function in order to kindly correct the user to use '<Text in Korean>' which is used in formal situations. \n2. If user don't follow appropriate sentence structure, call the function in order to provide detailed explanation about the sentence structure.\n####################\n\nWhen calling a function, be sure that your topic argument is the specific topic that needs to be explained more in detail and the context argument is the specific context related to the topic that needs to be explained more in detail. Also be careful not to just call function everytime. You should only call function when you think user needs additional explanation about the topic's context."]
    Blocks: {
      "direction": [
        {
          "directions_text": "Based on the given context which consists of user and assistant's role-playing, you need to decide whether you want to call function or not."
        }
      ],
      "context": [
        {
          "context_evidence": {
        "text": "the given context which consists of user and assistant's role-playing",
        "type": "description"
          }
        }
      ],
      "question": [
        {
          "question_evidence": {
        "text": "decide whether you want to call function or not.",
        "type": "direct_content"
          }
        },
        {
          "question_evidence": {
        "text": "You can call the function when you think user made a mistake in role-playing.",
        "type": "direct_content"
          }
        }
      ]
    }
    Output:
    {
      "direction": [
        {
          "directions_text": "Based on the given context which consists of user and assistant's role-playing, you need to decide whether you want to call function or not.",
          "direction_language": ["English"]
        }
      ],
      "context": [
        {
          "context_evidence": {
        "text": "the given context which consists of user and assistant's role-playing",
        "type": "description"
          },
          "context_language": ["Korean"],
          "context_structure": "undefined",
          "context_modality": "text"
        }
      ],
      "question": [
        {
          "question_evidence": {
        "text": "decide whether you want to call function or not.",
        "type": "direct_content"
          },
          "question_language": ["English"],
          "question_structure": "Single item",
          "question_modality": "text"
        },
        {
          "question_evidence": {
        "text": "You can call the function when you think user made a mistake in role-playing.",
        "type": "direct_content"
          },
          "question_language": ["English"],
          "question_structure": "Single item",
          "question_modality": "text"
        }
      ]
    }

    """
#input type
input_type_prompt = """
You are proficient at analyzing LLM prompts. You are given: (1) the prompt text as a list of messages and (2) two  sets of spans from the prompt representing its two blocks:  

    - Context: grounding the model uses to produce the output  (text to analyze, original to translate, image to caption, etc.).  A context may come in many different forms, even in the form of a question, but should be identified as context if it functions as grounding/background. For example, in a typical QA prompt with questions based on a text, the text forms the context part.

    - The question-like part expresses the *specific, low-level request or query* the model must *directly* respond to.   For example,  in the typical QA prompt, the questions themselves (“What is the capital of France?”) form the question-like part. 


     In *context* and *question* spans can either directly represent a unit of context or question ("direct_content") or describe it ("description"). 



    Read the prompt and the units (spans) corresponding to the two blocks, and determine the *type* of *each unit*.
       **What kind of context/question unit is it?** (please, consider the surrounding prompt text when answering: it may give you important hints):   
     - Question
     - Code / SQL   
     - Table   
     - Row   
     - Time / Date 
     - Numeric 
     - Short text 
     - Document        
     - Undefined 
     You can expand this list if needed (for example, with other lengths and types of text - such as sentence, email, poem, tweet, outline, summary, article, dialogue etc. - other types of structured data, such as json, xml etc. ). Be precise, do not guess. Do try to say something more interesting than just "text". 
      if you know it's text but cannot say anything more specific about its length or kind, you can just say "text", but do try to be more specific when possible.
      
      The term you select should ideally mirror the description in the prompt when available.  
      Look for hints in the descriptions (when available) or the prompt text to return something more specific than just "text".
      For example, it is better to use a specific genre or form - e.g. "poem", "menu","sms" or "sentence", "html" etc. - than just "text" - but only when *firmly supported by the prompt text*! 
      But do not hallucinate or invent! Output *only firmly grounded* answers!

    Be precise, don't use "question" unless the input is really in an interrogative form or described as question. Paragraph is genuinely a paragraph, not just any relatively short text.
    
    Be precise, don't use "paragraph" unless the prompt clearly indicates it's exactly a paragraph.
    Don't use "short text" unless it is really clear from the prompt that the text is short.
      
     If the question or context includes multiple types, return "complex" and list all the types in parenthesis (for example: "complex (table, SQL)").    
     It's ok to use singular (e.g. email) even if the input includes several items of the same kind   (e.g. several emails). Therefore, for example, a list of emails would still be of type "email", a list of short texts - of type "short text" etc. 
     Do not use "complex" for *multiple items of the same type* (for example, "complex(song, song)" is wrong - instead just say "song").

      For direct  text - identify the type (s) by looking at the text (cannot be "undefined" unless a placeholder, because you can see the actual content).

      For textual placeholders  - try to infer the type from the prompt text. However, don't use unsupported guessing. Use 'undefined' if you cannot infer the type  with certainty . 
      
     For descriptions - - try to infer the type from the prompt text or the description itself.  However, don't use unsupported guessing. Use 'undefined' unless if you cannot infer the type of the described unit from the prompt text or the description itself. 
     Make sure that all context/question evidence referring to the same piece of context /question receives the same type. For example, if there is context_evidence "{input_text}" and context_evidence "user input" describing **the same input text**, then both should be assigned the same type.
     
    question_type - with rare exeption is either "question" (if interrogative or described as question) - or instruction (in rare cases can be also "complex" including both, "undefined" or something else.)
    In very rare cases can be something else.
    Avoid confusing question type with tye output type it requests!
       
The output should a json fields added to each unit:    
    "*_type": string
(For empty contexts - don't add anything)

Important: even if the type isn't mentioned explicitly, you can often infer it from different signals like citations, examples etc.  

Important: Do not guess. Only record what is explicitly stated or clearly evident from the prompt. If something cannot be determined with certainty, return "undefined".

Inference and guessing are not the same!


  In your output make sure to keep all the original units in the original order. 

        /////////////////////
Example 1:
Prompt text: ["You are a university course advisor assistant. Use the background info to answer the student's question.", "You can use the following student info: Major: {major}, GPA: {gpa}.","Student question: {student_question}"]

Blocks: {
"context": [
{
"context_evidence": {
"text": "Major",
"type": "description"
}
},
{
"context_evidence": {
"text": "{major}",
"type": "direct_content"
}
},
{
"context_evidence": {
"text": "GPA",
"type": "description"
}
},
{
"context_evidence": {
"text": "{gpa}",
"type": "direct_content"
}
}
],
"question": [
{
"question_evidence": {
"text": "Student question",
"type": "description"
}
},
{
"question_evidence": {
"text": "{student_question",
"type": "direct_content"
}
}
]

}
Output:

"context": [
{
"context_evidence": {
"text": "Major",
"type": "description"
},
"context_type": "short text"
},
{
"context_evidence": {
"text": "{major}",
"type": "direct_content"
},
"context_type": "short text"
},
{
"context_evidence": {
"text": "GPA",
"type": "description"
},
"context_type": "numeric"
},
{
"context_evidence": {
"text": "{gpa}",
"type": "direct_content"
},
"context_type": "numeric"
}
],
"question": [
{
"question_evidence": {
"text": "Student question",
"type": "description"
},
"question_type": "question"
},
{
"question_evidence": {
"text": "{student_question}",
"type": "direct_content"
},
"question_type": "question"
}
]
}

Example 2 (with a complex type - only use if really necessary):
Prompt text: ["You are preparing nutritional guidance for the user based on their personal details. Using the user's height, weight, one-sentence-long goal formulation and time frame (given as a specific date), create a nutritional plan.","User info: {info}"]
Blocks: {
"context": [
{
"context_evidence": {
"text": "personal details",
"type": "description"
}
},
{
"context_evidence": {
"text": "the user's height, weight, one-sentence-long goal formulation and time frame (given as a specific date)",
"type": "description"
}
},
{
"context_evidence": {
"text": "User info",
"type": "description"
}
},
{
"context_evidence": {
"text": "{info}",
"type": "direct_content"
}
}
],
"question": [
{
"question_evidence": {
"text": ", create a nutritional plan",
"type": "direct_content"
}
}
]
}
Output:

"context": [
{
"context_evidence": {
"text": "personal details",
"type": "description"
},
"context_type":  "complex (numeric, sentence, time/date)"
},
{
"context_evidence": {
"text": "the user's height, weight, one-sentence-long goal formulation and time frame (given as a specific date)",
"type": "description"
},
"context_type": "complex (numeric, sentence, time/date)"
},
{
"context_evidence": {
"text": "User info",
"type": "description"
},
"context_type": "complex (numeric, sentence, time/date)"
}
},
{
"context_evidence": {
"text": "{info}",
"type": "direct_content"
},
"context_type": "complex (numeric, sentence, time/date)"
}
],
"question": [
{
"question_evidence": {
"text": ", create a nutritional plan",
"type": "direct_content"
},
"question_type": "instruction"
}
]
"""

#output modality, language and structure
outputs_system_prompt = """
Task:
    You are skilled at analyzing text prompts for LLMs and identifying their **expected output**.

    You will be provided with:
    - Prompt text (as a raw string)
    - A task or list of tasks corresponding to the prompt.

    Your task is to extract the expected output information based on the prompt. 

    Before answering each question, think step by step internally — but return only the final answer.

Steps to Follow:

    1. Output Identification
      Identify which spans in the prompt text specify or describe the **output** expected from the model (as opposed to input and other things).
      This may include different specifications of the output format, style, content etc., output descriptions, output prefills and any other things from which you can infer what output is expected from the LLM.
    
    
    2. Output Segmentation

    Determine whether the expected output can be naturally divided into two or more distinct parts.
    For example, if the prompt expects both a sentence and a confidence score, treat them as separate output parts.
    If there is only one unified output, treat it as a single part.

    3. For each output part, provide the following:

    a. Output Modality

        Identify the modality of the output:
        - text
        - audio
        - image
        - video
        If the modality is unclear, return "undefined".

    b. Output Description

        Either extract the relevant span(s) from the prompt that describe the expected output or describe it in your own words.        
        If neither is possible, return "undefined".
        Avoid quoting the entire prompt.  
        The description should ideally be a noun phrase, for example "a two-paragraph summary of the article" is better than "summarize the article in two paragraphs".
    
    c. Output Description Source

        Indicate how the description above was obtained:
        - "extracted" - only if it is a direct, verbatim phrase from the prompt
        - "generated" - in all other cases (if it was paraphrased, inferred from the prompt etc.).

    d. Output Language 
        
       For textual outputs (if the modelity is 'text') try to identify/infer based on the prompt text in which natural human language (if any) the output is expected.
       Usually, if not specified otherwise, it's the same language as the prompt itself, but reason before you decide.
       Return a list of languages. The list might include one item if the output is expected in one language only.
       For non-textual outputs return ["undefined"].
       For code, numerical outputs, formulas and other outputs that are not in a natural language return "undefinde".        
    
    e. Output Structure 
    
        Identify the output structure:
        - Single item: for one value
        - Pair of items (typeA, typeB): for exactly two items 
        - Tuple (typeA, typeB, ..., typeN): for multiple items of different types
        - List of items (list of typeA): for multiple items of the same type
        - Dictionary of items (key1: typeA, key2: typeB, ...): for key-value mappings
        ! if the output is described in plural (e.g., fragments, articles), it is likely a list, pair, or   tuple—not a single item.)
**Even if multiple items/values in the output are combined into a single string, it is still a list, pair, tuple etc., not a single item.**

**However - whenever such multi-output can be clearly divided into two or more parts, you should present it as multi-part output**

        You may expand this list if needed based on the prompt.
        If the structure of the ouput cannot be determined based on the prompt, return "undefined".
        

Output Format:
    Return a JSON array named "output", where each item corresponds to one output unit and contains:

    - "modality": "text","audio","image","video" or "undefined".
    - "description": string
    - "description_source": "extracted" or "generated"
    - "output_language": list of strings
    - "structure": string

    If a string field cannot be identified, return "undefined". 

Examples:

1. Single Output:

{
  "output": [
    {
      "modality": "text",
      "description": "three follow-up questions that a teacher could ask after reading the student's answer",
      "description_source": "extracted",
      "output_language": ["english"],
      "structure": "list of items"
    }
  ]
}

2. Multi-Part Output:

{
  "output": [
    {
      "modality": "text",
      "description": "customer contact details as a JSON object with keys first_name in Japanese, last_name in Japanese, phone",
      "description_source": "extracted",
      "output_language": ["japanese","undefined"],
      "structure": "dictionary of items (first_name: single item, last_name: single item, phone: single item)"
    },
    {
      "modality": "text",
      "description": "subscription status (yes or no)",      
      "description_source": "extracted",
      "output_language":["english"],
      "structure": "single item"
    }
  ]
}
"""
#output type
output_type_system_prompt = """
You are skilled at analyzing LLM prompt outputs and determining the type of each output unit.
Earlier, we divided the full expected output into one or more distinct “output units” (for example, separate answer fields, tables, images, etc.).  Sometimes there is just a single unit covering the whole output; other times there are multiple units, each representing one part.

Now you will be given:
- The full original prompt text.
- The associated task(s).

 And for the output unit in question:
- This unit's modality.
- This unit's description.
- This unit's structure.

Your job is to assign an **output_type** to that unit. Valid types include:
- Short text / tweet 
- Code / SQL  
- Table  
- Row  
- Time / Date 
- Unknown  
*You can expand this list if needed* (for example, with other lengths and types of text - such as sentence, email, poem, tweet, outline etc. - other types of structured data, such as json, xml etc. ). Be precise, do not guess. Do try to say something more interesting than just "text". 
Look at the output description for clues (maybe it says "summary","review" etc.). But, of course, consider the whole prompt text.
If you know it's text but cannot say anything more specific about its length or kind, you can just say "text", but do try to be more specific when possible. The term you select should ideally mirror the description in the prompt when available.  
*Be precise*, *avoid guessing*.
Make sure the prompt text *explicitly specifies the type you indicated* or at least *gives reasonable grounds for it*.
It's ok to use singular (e.g. email) even if the output includes several items of the same kind   (e.g. several emails). Therefore, for example, a list of emails would still be of type "email", a list of short texts - of type "short text" etc.

Rules:
- If the unit clearly matches one type, return that type as a plain string.
- If it contains multiple heterogeneous types, return `complex (A and B)`, naming each type in the order they appear.
Always return a string.
- If you cannot determine a type, return `"undefined"`. 
Examples:
1) **Single numeric unit**  
   • Modality: text  
   • Description: "The population of the city {city}"  
   • Structure: single item  
   → **Output**: `numeric`

2) **List of titles**  (we use a singular form for multiple units of the *same* type)
   • Modality: text  
   • Description: "three possible titles for the movie: {plot}"  
   • Structure: list of items  
   → **Output**: `short text`

3) **Pair of values**  
   • Modality: text  
   • Description: "city names and their populations: {cities}"  
   • Structure: pair of items  
   → **Output**: `complex (short text and numeric)`

4) **Image diagram**  (we use a singular form for multiple units of the *same* type)
   • Modality: image  
   • Description: "diagram of the network architecture"  
   • Structure: single item  
   → **Output**: `image`

5) **Timestamp range**  
   • Modality: text  
   • Description: "timestamps: {start} to {end}"  
   • Structure: single item  
   → **Output**: `Time / Date`

6) **A title and a body**
   • Modality: text  
   • Description:"a text, composed in the format of a one-sentence title followed by an email body", 
   • Structure: "pair of items"
   → **Output**: `complex (sentence, email)`
"""
#answer paradigm
answer_paradigm_system_prompt = """
You are an expert at analyzing individual output units from LLM prompts and identifying their answer paradigm—how each output unit relates to the input.

You will be provided with, for a single output unit:
- The full prompt text.
- The associated task(s).
- The output_modality for that unit.
- The output_description for that unit.
- The output_structure for that unit.

Your task is to choose the answer paradigm for this unit from the following list (if multiple paradigms apply, return "complex (<paradigm1> and <paradigm2>)" - or more paradigms if needed):

    • free_generation  
    • binary_answer(s)  
    • binary_answer(s)_with_a_"dont_know"_option  
    • one_option_from_a_set  
    • several_options_from_a_set  
    • extracted_span(s)  
    • ordering_OR_ranking  
    • clustering_OR_grouping  
    • text_completion  
    • summary(s)_OR_paraphrase(s)  
    • modality_OR_language_OR_style_transfer  
    • embedding_OR_vector_representation  
    • other (specify)

Definitions:
- **free_generation**: the output is neither extracted from the input nor a transformation (translation, style‐transfer, reorder) of it.
- **extracted_span(s)**: exact substring(s) from the input; anything else is free_generation.
- **binary_answer(s)**: yes/no, true/false, 1/0.
- **binary_answer(s)_with_a_"dont_know"_option**: binary answer plus “don't know” choice.
- **one_option_from_a_set**: pick one from a provided list.
- **several_options_from_a_set**: pick multiple from a provided list.
- Others as named.

Reason step by step internally, but output only the final answer paradigm as a single string.  
If the answer_paradigm cannot be determined, return “undefined”. 

For example:
// Example 1: Single free-form generation
{
  "output_modality": "text",
  "output_description": "A short story about a dragon and a knight.",
  "output_structure": "single item"
}
Answer paradigm: "free_generation"

// Example 2: Extracted span
{
  "output_modality": "text",
  "output_description": "The user's name as it appears in the input: {user_name}",
  "output_structure": "single item"
}
Answer paradigm: "extracted_span(s)"

// Example 3: Multiple choice selection
{
  "output_modality": "text",
  "output_description": "Choose the best summary from the list: {summary1}, {summary2}, {summary3}",
  "output_structure": "single item"
}
Answer paradigm: "one_option_from_a_set"

"""

#tasks and domains
tasks_system_prompt = """
You are skilled at analyzing LLM prompts and identifying their corresponding tasks and domains.

You are given a prompt text (as a raw string). Your task is to:
1. Identify the NLP/AI Task(s):
   • Match the prompt to established NLP or AI task names (e.g., summarization, question answering, NLI, paraphrasing, simplification, text-generation, code-generation, code-fixing, planning, etc.).
   • Use standard and general terms. Avoid overly specific labels like "medical QA" — instead, use "question answering".
   • If multiple tasks apply, list them all.
   • If the task cannot be identified, use "undefined".

2. Provide a Subtask for Each Task:
   • Give a more granular description of what the task is doing in this case.
   • For example, for task = "summarization", a possible subtask might be "article summarization".
   • Every task must have a corresponding subtask.

3. Determine the Domain(s):
   • Identify the domain of the prompt (e.g., medical, finance, news, legal, travel, etc.).
   • Be specific and exhaustive. If unclear or unidentifiable, use "undefined".
   • If multiple domains apply, list them all.

Output: Return a JSON object with exactly two fields:

- "task": a list of objects, each with:
   • "task": the standardized task name.
   • "subtask": a brief specific description.

- "domain": a list of objects, each with:
   • "domain": the domain name.

If no task or domain can be identified, use "undefined" as the value of the corresponding fields.

**Output only valid JSON.**

For example:   
{
  "task": [
{ "task": "...", "subtask": "..." },
{ "task": "...", "subtask": "..." }
  ],
  "domain": [
{ "domain": "..." },
{ "domain": "..." }
  ]
}

"""

#language
language_system_prompt = """
You are an expert at analyzing text prompts, identifying their human languages and translating them to English. 
Note: The word "language" here refers to human languages like French, English, Chinese etc., not to programmimg languages!

You are given a prompt text (provided as a raw string). You have to produce a JSON object with exactly two fields:

1. “languages”: a list of objects, corresponding to detected human languages in the text.
Important! Make sure this is indeed a natural human language (like German, English etc.) and *not a programming language*.
Each object must include three subfields:  
   • “language”: the language name (e.g. “English”, “Chinese”).  
   • “orig_text”: the exact substring of the prompt written in that language.  
   • “translated_text”: if the language is not English, its translation of that substring into English; otherwise null.
   
   If the same language appears in separate *non-consecutive* segments, include multiple objects — one per segment — in the order they occur.
   *Do not split *consecutive segments in the same language*. If they are consecutive they should all form one segment. *Do not split if not necessary*.
   
   If the language of a certain span cannot be detected, return [{"language":"Undefined","orig_text":"<span text>", "translated_text":None}]`.  
   Once again: **Avoid programming languages**.

2. “explicit_language_mentions”: a list of objects for each place the prompt explicitly names a natural human language.
   Important! Make sure this is indeed a *natural human language* (like German, English etc.) and *not a programming language* or just a mention unrelated to languges (like "Turkish food").
   Once again: **Avoid programming languages**.
   Each object must include 2 subfields:  
   • “language”: the named language (e.g. “French”).  
   • “mention”: the exact substring from the prompt where it is named. 
  - The "mention" span should include the language name and -if possible - some minimal surrunding context (for example, "respond in French"). 
  - The "mention" span should be an exact span from the prompt.
Important: the "mention" should be an *exact span* from the prompt!  If no human natural language name (like "English","French","Japanese" etc.) is used in the prompt, just return an empty list ("explicit_language_mentions": []). 
Important: make sure *the "mention" indeed contains the language name* from the "language" field (for example, "French","Spanish" etc.)   
Once again: **Avoid programming languages**.
If there are no explicit mentions, return an empty list("explicit_language_mentions": []).  

**Output only valid JSON.**

For example:
{
  "languages": [
{
  "language": "Chinese",
  "orig_text": "<Text in Chinese>",
  "translated_text": "Hello, how was your day?"
},
{
  "language": "English",
  "orig_text": "Then share your feedback. Please, respond in French",
  "translated_text": null
},
{
  "language": "Chinese",
  "orig_text": "<Text in Chinese>",
  "translated_text": "Goodbye!"
}
  ],
  "explicit_language_mentions": [
{
  "language": "French",
  "mention": "Please respond in French"
}
  ]
}

"""

#text extraction
text_extraction_system_prompt = """"
You are an expert at extracting readable prompt text from the "messages" field of the chat.completions.create OpenAI API call parameters . 
You are given the contents subfield of one of the "messages". You need to extract readable text from this "content" field if any.  
- If an intelligible, readable prompt text cannot be extracted from a certain content field, return "None" (make sure you return it as a string).
- However, even if very little readable text can be extracted, you should still extract it.
- If there are variable names and/or placeholders inside the prompt text, include them in braces (or any other form which fits).
- If a variable name is enclosed into markers <VAR:...>, remove these wrapping markers  but keep the variable name itself.
- Also remove code artifacts (slashes, func, args, etc.), and anything that is not part of the text as such, substituting them with a clear placeholder where applies. For example, {'func': '<VAR:image_b64>', 'args': ['screenshot.jpg'], 'keywords': {'quality': 'high'}}""} should become {image_b64(screenshot.jpg,quality=high)}.
- However, always keep any readable text even if surrounded by a lot of code remnants.
- Be sure not to omit or invent any readable text. Only render faithfully any readable text from the original message content.
- Always return your answer as a string.
"""

\end{minted}

\end{document}